\documentclass[Afour,sageh,times]{sagej}

\usepackage[T1]{fontenc}
\usepackage[latin9]{inputenc}
\usepackage[active]{srcltx}
\usepackage{array}
\usepackage{textcomp}
\usepackage{multirow}
\usepackage{amsmath}
\usepackage{amssymb}
\usepackage{stmaryrd}
\usepackage{graphicx}
\usepackage{float}
\PassOptionsToPackage{normalem}{ulem}
\usepackage{ulem}

\providecommand{\selectlanguage}[1]{}
\renewcommand{\selectlanguage}[1]{}
\providecommand{\foreignlanguage}[1]{}
\renewcommand{\foreignlanguage}[1]{}

\usepackage[unicode=true,pdftex, colorlinks=true, linkcolor=red!80!black, citecolor = green!50!black, urlcolor = black,filecolor=black , pagebackref=false,hypertexnames=false, plainpages=false, pdfpagelabels]{hyperref}

\makeatletter

\let\cite\citep
\makeatother

\newcommand{\noun}[1]{\textsc{#1}}
\providecommand{\tabularnewline}{\\}

\newcommand{\added}[1]{\textcolor{black}{#1}}
\newenvironment{addedstuff}{\par\color{black}}{\par}

\newcommand{\addednew}[1]{\textcolor{black}{#1}}
\newenvironment{addedstuffnew}{\par\color{black}}{\par}

\usepackage[T1]{fontenc}
\usepackage{courier}
\usepackage{subcaption}
\usepackage{color}

\usepackage{upquote}
\usepackage[table]{xcolor}
\usepackage{listings}
\usepackage[most]{tcolorbox}

\usepackage{tikz}
\usetikzlibrary{shapes,snakes}
\definecolor{my_blue}{RGB}{203,245,255}
\definecolor{my_yellow}{RGB}{250,247,203}
\newcommand\myboxblue[2][]{\hspace{0.4mm}\tikz[overlay]\node[fill=my_blue,inner sep=2.0pt, anchor=text, rectangle, rounded corners=0.0mm,draw=black] {#2};\phantom{#2}\hspace{0.8mm}}
\newcommand\myboxyellow[2][]{\hspace{0.4mm}\tikz[overlay]\node[fill=my_yellow,inner sep=2.0pt, anchor=text, rectangle, rounded corners=0.0mm,draw=black] {#2};\phantom{#2}\hspace{0.8mm}}

\usepackage{caption}
\usepackage{graphics}
\usepackage{placeins}
\usepackage{graphicx, epstopdf}
\usepackage[framed, numbered]{matlab-prettifier}

\usepackage[linesnumbered,commentsnumbered,ruled,vlined]{algorithm2e}

\DeclareFixedFont{\ttb}{T1}{txtt}{bx}{n}{12} %
\DeclareFixedFont{\ttm}{T1}{txtt}{m}{n}{12}  %

\usepackage{color}
\definecolor{deepblue}{rgb}{0,0,1}
\definecolor{deepred}{rgb}{0.6,0,0}
\definecolor{deepgreen}{rgb}{0,0.5,0}

\makeatother

\definecolor{noticeframe}{RGB}{31,78,121}
\definecolor{noticeback}{RGB}{238,244,250}
\newcommand{\acceptednoticecitation}{%
J.~Tordesillas, V.~Klemm, J.~P.~How, and M.~Hutter. RAYEN: Imposition of Hard Convex Constraints on Neural Networks. \textit{International Journal of Robotics Research}, 2026.}
\AddToHookNext{shipout/foreground}{%
  \put(0,0){\begin{tikzpicture}[overlay]
    \node[anchor=north west, draw=noticeframe, fill=noticeback,
          line width=0.5pt, rounded corners=1mm,
          inner xsep=2.5mm, inner ysep=1.1mm,
          text width={\textwidth-5mm-0.5pt}, align=left,
          font=\sffamily\fontsize{7.5}{9}\selectfont]
      at ({1in+\hoffset+\oddsidemargin},-2.6mm)
      {\textbf{\color{noticeframe}Accepted for publication in the \textit{International Journal of Robotics Research} (IJRR).} Please cite as follows:
       \acceptednoticecitation};
  \end{tikzpicture}}}

\begin{document}

\title{RAYEN: Imposition of Hard Convex Constraints on Neural Networks}

\author{Jesus Tordesillas\affilnum{1}, Victor Klemm\affilnum{2}, Jonathan P.\ How\affilnum{3}, Marco Hutter\affilnum{2}}

\affiliation{
\affilnum{1}Trustworthy Robotics and AI Lab (TRAIL), IIT-ICAI, Comillas Pontifical University, Madrid, Spain \texttt{jtordesillas@comillas.edu}\\
\affilnum{2}Robotic Systems Lab, ETH Z{\"u}rich, Z{\"u}rich, Switzerland \{\texttt{{vklemm}, mahutter\}@ethz.ch}\\
\affilnum{3}Aerospace Controls Laboratory, Massachusetts Institute of Technology, Cambridge, MA, USA \texttt{jhow@mit.edu}
}

\renewcommand{\lstlistingname}{Script}
\begin{abstract}

\added{Despite the numerous applications of convex constraints in Robotics, enforcing them within learning-based frameworks remains an open challenge. Existing techniques either fail to guarantee satisfaction at all times, or incur prohibitive computational costs.} This paper presents RAYEN, a framework for imposing hard convex constraints
on the output or latent variables of a neural network. \added{RAYEN guarantees constraint satisfaction during both training and testing, for any input and any network weights. Unlike prior approaches, RAYEN avoids computationally-expensive orthogonal projections, soft constraints, conservative approximations of the feasible set, and slow iterative corrections.} RAYEN supports any combination of linear,
convex quadratic, second-order cone~(SOC), and linear matrix inequality
(LMI) constraints, with negligible overhead compared
to unconstrained networks. 
For instance, it imposes 1K quadratic
constraints on a 1K-dimensional variable \added{with only 8 ms of overhead compared to a network that does not enforce these constraints.} 
An LMI constraint with ${300\times300}$ dense
matrices on a 10K-dimensional variable \added{can be guaranteed with only 12~ms additional overhead.} 
When used in neural networks that approximate the solution of constrained \added{trajectory} optimization
problems, RAYEN runs 20 to 7468 times
faster than state-of-the-art algorithms, while guaranteeing constraint satisfaction at all times and achieving a near-optimal cost.
\added{Finally, we demonstrate RAYEN's ability to enforce actuator constraints on a learned locomotion policy by validating constraint satisfaction in both simulation and real-world experiments on a quadruped robot.}
\end{abstract}

\keywords{Constraints, convex, neural network, optimization, differentiable.\\\textbf{Code}\\ \url{https://github.com/leggedrobotics/rayen}}

\maketitle
\renewcommand{\thefootnote}{\arabic{footnote}}
\setcounter{footnote}{0}

\section{Introduction and Related Work \label{sec:Introduction-and-Related}}

\begin{figure}[t]
\begin{centering}
\includegraphics[width=1\columnwidth]{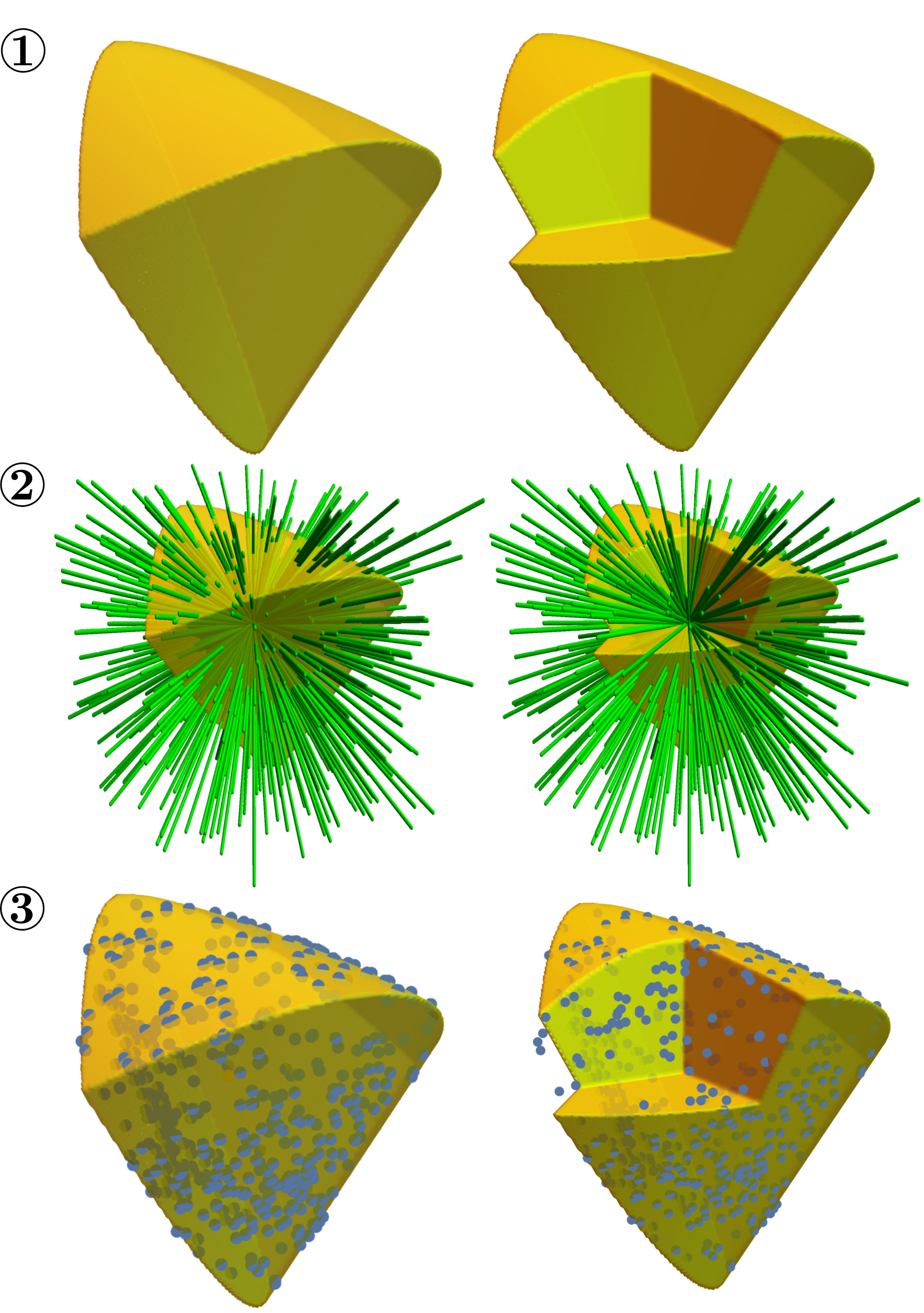}
\par\end{centering}
\caption{
RAYEN applied to a batch of 500 samples with a feasible set (\raisebox{.5pt}{\textcircled{\raisebox{-.9pt} {1}}})
defined by linear, convex quadratic, SOC, and LMI constraints. For
each sample in the batch, RAYEN lets the corresponding latent variable
of the network be the vector that defines the step to take from an
interior point of the feasible set (\raisebox{.5pt}{\textcircled{\raisebox{-.9pt} {2}}}).
The length of this vector is then adjusted to ensure that the endpoint
lies within the set (\raisebox{.5pt}{\textcircled{\raisebox{-.9pt} {3}}}).
For visualization purposes, a section of the set has been removed
in the right plots.  \label{fig:firstimage}}
\vspace{-0.6cm}
\end{figure}

Convex \added{constraints play a crucial role in many areas of Robotics such as control \cite{boyd1994linear, ames2019control}, trajectory planning \cite{tordesillas2020mader, schulman2014motion, malyuta2022convex}, state estimation~\cite{carlone2018convex, tian2021distributed}, signal processing \cite{mattingley2010real}, and computer vision \cite{monga2017handbook}. 
For example, linear constraints are extensively used to guarantee safety in obstacle avoidance~\cite{van2011reciprocal, tordesillas2020faster, liu2017planning}, convex quadratic constraints are leveraged in barrier functions~\cite{wang2021chance}, SOC constraints are used in grasping, contacts, and motion planning~\cite{lobo1998applications, grandia2019feedback, marcucci2023motion}, and LMI constraints are important in pose-graph optimization and Lyapunov's stability theory~\cite{boyd1994linear, majumdar2013control, tian2021distributed, ames2019control}.   }
In recent years, there has been an extensive use of neural networks
in all these applications due to their expressive power. However,
their lack of constraint satisfaction guarantees greatly limits their
applicability, especially for safety-critical applications. 

There are some \added{convex} constraints that can be easily imposed on the output
(or latent variable) of a neural network. For instance, box constraints
can be imposed using \noun{sigmoid} functions, nonnegative constraints
can be enforced using \noun{relu} functions, simplex constraints can
be guaranteed using \noun{softmax} functions, and some spherical or
ellipsoidal constraints can be imposed using normalization. However,
these methods are not applicable to more general types of constraints.

One common way to bias the network towards the satisfaction of the
constraints is via soft constraints, which consist of the addition
of terms in the training loss to penalize the violation of the constraints
\cite{marquez2017imposing}. Similar penalties are also used in physics-informed
neural networks \cite{raissi2019physics}. The main disadvantage of
this approach is that there are no constraint satisfaction guarantees
at test time. \added{Works such as \citet{chzhen2026improving} use data-driven approaches to approximately enforce the constraints, but also lack hard constraint guarantees}.

When the constraints are homogeneous linear inequality constraints
(i.e., $\boldsymbol{A}\boldsymbol{y}\le\boldsymbol{0}$), \citet{frerix2020homogeneous} leverage the Minkowski-Weyl theorem
to guarantee the satisfaction of these linear constraints at all times.\footnote{The Minkowski-Weyl theorem can be leveraged for any (convex)
polyhedron defined by $\boldsymbol{A}\boldsymbol{y}\le\boldsymbol{b}$,
but \cite{frerix2020homogeneous} focuses on the case $\boldsymbol{A}\boldsymbol{y}\le\boldsymbol{0}$} The main disadvantage of this approach is that it is only applicable
to linear inequality constraints. Moreover, this method requires running
offline the double description method \cite{motzkin1953double,fukuda2005double}
to obtain the V-representation (vertices and rays) of the polyhedron
defined by its H-representation (intersection of half-spaces). Even
if done offline, the double description method becomes intractable
for high-dimensional problems. Other related methods that focus on linear constraints
include \cite{hendriks2020linearly,balestriero2023police,huang2020linearly, li2023learning, cristian2023end, zeng2024glinsat, bouvier2024policed, ataei2025mpolice, zhu2026t}.
In contrast to these works, which focus only on linear constraints,
RAYEN supports any combination of linear, convex quadratic, SOC, and
LMI constraints. \added{
To avoid the orthogonal projection step, \citet{maruyama2018guaranteeing} proposes to first use a sigmoid to constrain the output to a hypercube that encloses the feasible set, and then use the distances to the border of the set to scale it down to force it to stay in the set. However, the question of how to compute these distances, which is crucial for its computational tractability, is left unanswered. Moreover, the results shown in \citet{maruyama2018guaranteeing} are only for linear constraints, and linear equality constraints are not taken into account.
}

There have also been many recent advances in implicit layers that
solve different types of optimization problems~\cite{djolonga2017differentiable,amos2018differentiable,agrawal2019differentiating,gould2019deep},
with some works focusing on quadratic programming~\cite{amos2017optnet}
or, more generally, convex optimization problems~\cite{agrawal2019differentiable}.
These implicit layers can be leveraged to enforce constraints on the
network during training and/or testing by obtaining the orthogonal
projection onto the feasible set. This orthogonal projection could
also be obtained using the Dykstra\textquoteright s Projection
Algorithm \cite{gaffke1989cyclic,boyle1986method,chen2018approximating}.
While the use of these orthogonal projections guarantees the satisfaction
of the constraints, it is typically at the expense of very high computation
times. By leveraging analytic expressions of the distance to the
boundaries of the convex set, RAYEN is able to avoid this slow orthogonal
projection step and guarantee the constraints in a much more computationally-efficient
manner. 

With the goal of reducing the computation time, \citet{donti2021dc3} proposed
DC3, an algorithm that first uses completion
to guarantee the equality constraints, and then enforces the inequality
constraints by using an inner gradient \added{descent} procedure that takes
steps along the manifold defined by the equality constraints. This
inner gradient descent procedure is performed both at training and
testing time. While DC3 is typically less computationally expensive
\added{than} projection-based methods, this inequality correction may still
require many steps, as we will show in Section \ref{subsec:results_optimization}.
Moreover, this inner gradient correction may also suffer from convergence
issues for general convex constraints. Compared to DC3, RAYEN does
not need to rely on an inner gradient descent correction, avoiding therefore
any convergence issues and substantially reducing the computation
time. \added{Other approaches that also rely on (potentially computationally expensive) iterative algorithms to ensure the feasibility include \citet{grontas2025pinet}, \citet{nguyen2025fsnet}, and \citet{liang2025efficient}. Similarly, works such as \citet{schneider2026soft} need to use Newton-Raphson or bisection methods for general convex sets, while~\citet{liang2023low, liang2024homeomorphic} propose to learn homeomorphic projections and use bisection to enforce feasibility. Another option to enforce convex constraints is to perform a projection relying on computationally-expensive differentiable convex optimization solvers used as an implicit layer of the network \cite{min2024hardnet, tang2026lmi}. By eschewing iterative algorithms and learned projections in favor of leveraging analytically-found distances to the set boundaries, RAYEN significantly reduces computational overhead.
}

\added{There have also been extensive sets of works \cite{mezard2009constraint, johnston1989learning, amizadeh2019pdp} that use neural networks for Constraint Satisfaction Problems, where the goal is to generate a set of variables defined over finite and discrete domains (Boolean domains for example) that satisfy some constraints~\cite{kumar1992algorithms}. Our focus is instead on convex constraints, where the decision variables are defined over continuous (and potentially unbounded) domains.}

\added{Since our initial preprint, subsequent works have expanded on these themes. For instance, \citet{konstantinov2023new} proposes a similar framework for linear and convex quadratic constraints, while \citet{liu2025fast} builds upon our approach by employing a homeomorphism between the convex constraint set and a unit ball.}

The contributions of this work are therefore summarized as follows:
\begin{itemize}
\item Framework to impose by construction hard convex constraints on the
output or latent variable of a neural network. The constraints are
guaranteed to be satisfied at all times, for any input and/or weights
of the network. No conservative approximations of the set are used. 
\item Any combination of linear, convex quadratic, SOC, and LMI constraints
is supported. For example, RAYEN can impose 1K quadratic constraints on a
1K-dimensional variable with a computation overhead of only 8 ms.
Similarly, a $300\times300$ dense LMI constraint can be imposed on a 10K-dimensional
variable with an overhead of less than 11~ms. 
\item When used in neural networks that approximate the solution of optimization
problems, RAYEN showcases computation times between 20 and 7468 times
faster than other state-of-the-art algorithms, while generating feasible
solutions whose costs are very close to the optimal value.
\end{itemize}
The notation used throughout the paper is available in Table~\ref{tab:Notation}.

\section{Problem Setup\label{sec:Problem-Setup}}

\begin{table}
\begin{centering}
\caption{Notation used in this paper \label{tab:Notation}}
\par\end{centering}
\noindent\resizebox{\columnwidth}{!}{%
\begin{centering}
\begin{tabular}{|>{\centering}m{0.25\columnwidth}|>{\raggedright}m{0.85\columnwidth}|}
\hline 
\textbf{Symbol} & \textbf{\qquad \qquad \qquad \qquad Meaning}\tabularnewline
\hline 
\hline 
$c,\boldsymbol{c},\boldsymbol{C},\mathcal{C}$ & Scalar, column vector, matrix, and set\tabularnewline
\hline 
$\left\Vert \boldsymbol{c}\right\Vert $ & Euclidean norm of the vector $\boldsymbol{c}$\tabularnewline
\hline 
$\boldsymbol{C}\succeq\boldsymbol{0}$ & $\boldsymbol{C}$ is a (symmetric) positive semidefinite matrix\tabularnewline
\hline 
$\boldsymbol{C}\succ\boldsymbol{0}$ & $\boldsymbol{C}$ is a (symmetric) positive definite matrix\tabularnewline
\hline 
$\boldsymbol{I}$ & Identity matrix\tabularnewline
\hline 
$\bar{\boldsymbol{c}}$ & $\bar{\boldsymbol{c}}:=\frac{\boldsymbol{c}}{\left\Vert \boldsymbol{c}\right\Vert }$\tabularnewline
\hline 
$\boldsymbol{C}^{\dagger}$ & Pseudoinverse of a matrix $\boldsymbol{C}$\tabularnewline
\hline 
$\text{max}\left(\boldsymbol{c}\right)$ & Maximum of the elements of the column vector $\boldsymbol{c}$\tabularnewline
\hline 
$\text{max}\left(\mathcal{C}\right)$ & Maximum of the elements of the set $\mathcal{C}$\tabularnewline
\hline 
$\text{inf}\left(\mathcal{C}\right)$ & Infimum of the set $\mathcal{C}$\tabularnewline
\hline 
$\added{\text{relu}\left(a\right)}$ &\added{$ \text{max}  \left( \{ 0,a  \} \right) $}\tabularnewline
\hline 
$\added{\text{relu}\left(\boldsymbol{a}\right)}$ &\added{$\text{relu}\left(\cdot\right)$ applied elementwise to the elements of the vector $\boldsymbol{a}$}\tabularnewline
\hline 
$\partial\mathcal{C}$ & Frontier of the set $\mathcal{C}$\tabularnewline
\hline 
$\boldsymbol{a}\le\boldsymbol{b}$, $\boldsymbol{a}=\boldsymbol{b}$ & Element-wise inequality, element-wise equality\tabularnewline
\hline 
$\boldsymbol{A}\oslash\boldsymbol{B}$ & Element-wise division\tabularnewline
\hline 
$\boldsymbol{0},\boldsymbol{1}$ & Matrix or vector of zeros/ones (dimensions given by the context).
If the dimensions need to be specified, the subscript $a\times b$ (rows $\times$ columns)
will be used\tabularnewline
\hline 
$\emptyset$ & Empty set\tabularnewline
\hline 
$\text{aff}\left(\mathcal{S}\right)$ & Affine hull of a set $\mathcal{S}$ (i.e., smallest affine set containing
$\mathcal{S}$)\tabularnewline
\hline 
$\boldsymbol{c}_{[i]}$ & $i$-th element of the column vector $\boldsymbol{c}\in\mathbb{R}^{a}$.
$i\in\{0,\ldots,a-1\}$\tabularnewline
\hline 
$\boldsymbol{C}_{[i,:]}$ & $i$-th row of the matrix $\boldsymbol{C}$, $i\in\{0,\ldots,a-1\}$.
If $\boldsymbol{C}\in\mathbb{R}^{a\times b}$, then $\boldsymbol{C}_{[i,:]}\in\mathbb{R}^{1\times b}$
(i.e., a row vector).\tabularnewline
\hline 
$\boldsymbol{C}_{[\mathcal{S},:]}$ , $\boldsymbol{c}_{[\mathcal{S}]}$ & $\boldsymbol{C}_{[\mathcal{S},:]}$ is a matrix whose rows are the
rows of $\boldsymbol{C}$ whose indexes are in the set $\mathcal{\mathcal{S}}\subseteq\mathbb{N}$.
Analogous definition for $\boldsymbol{c}_{[\mathcal{S}]}$\tabularnewline
\hline 
$\text{abs}\left(\boldsymbol{a}\right)$ & Element-wise absolute value\tabularnewline
\hline 
$\text{softmax}\left(\boldsymbol{a}\right)$ & Softmax function. I.e., if $\boldsymbol{b}=\text{softmax}\left(\boldsymbol{a}\right)$,
then $\boldsymbol{b}_{[i]}=e^{\boldsymbol{a}_{[i]}}/\sum_{j}e^{\boldsymbol{a}_{[j]}}$\tabularnewline
\hline 
$\text{eig}\left(\boldsymbol{C}\right)$ & Column vector containing all the eigenvalues of the matrix $\boldsymbol{C}$\tabularnewline
\hline 
SOC & Second-order Cone\tabularnewline
\hline 
LMI & Linear Matrix Inequality\tabularnewline
\hline 
\addednew{LP} & \addednew{Linear Program}\tabularnewline
\hline 
\addednew{QCQP} & \addednew{Quadratically constrained Quadratic Program}\tabularnewline
\hline 
\end{tabular}
\par\end{centering}
}
\end{table}

This work addresses the problem of how to ensure that the output (or
a latent variable) $\arraycolsep=1.2pt\boldsymbol{y}:=\left[\begin{array}{ccc}
\boldsymbol{y}_{[0]} & \cdots & \boldsymbol{y}_{[k-1]}\end{array}\right]^{T}\in\mathbb{R}^{k}$ of a neural network lies in a convex set $\mathcal{Y}$ defined by
the following constraints:
{\small{}
\begin{align}
 & \boldsymbol{A}_{1}\boldsymbol{y}\le\boldsymbol{b}_{1}\label{eq:linear_ineq}\\
 & \boldsymbol{A}_{2}\boldsymbol{y}=\boldsymbol{b}_{2}\label{eq:linear_eq}\\
 & g_{i}\left(\boldsymbol{y}\right):=\frac{1}{2}\boldsymbol{y}^{T}\boldsymbol{P}_{i}\boldsymbol{y}+\boldsymbol{q}_{i}^{T}\boldsymbol{y}+r_{i}\le0\;\;i=0,\ldots,\eta-1\label{eq:convex_quad}\\
 & h_{j}\left(\boldsymbol{y}\right):=\left\Vert \boldsymbol{M}_{j}\boldsymbol{y}+\boldsymbol{s}_{j}\right\Vert -\boldsymbol{c}_{j}^{T}\boldsymbol{y}-d_{j}\le0\;\;j=0,\ldots,\mu-1\label{eq:soc}\\
 & \boldsymbol{W}(\boldsymbol{y}):=\boldsymbol{y}_{[0]}\boldsymbol{F}_{0}+\ldots+\boldsymbol{y}_{[k-1]}\boldsymbol{F}_{k-1}+\boldsymbol{F}_{k}\succeq\boldsymbol{0}\label{eq:lmi}
\end{align}
}{\small\par}
\begin{figure*}
\begin{centering}
\includegraphics[width=1\textwidth]{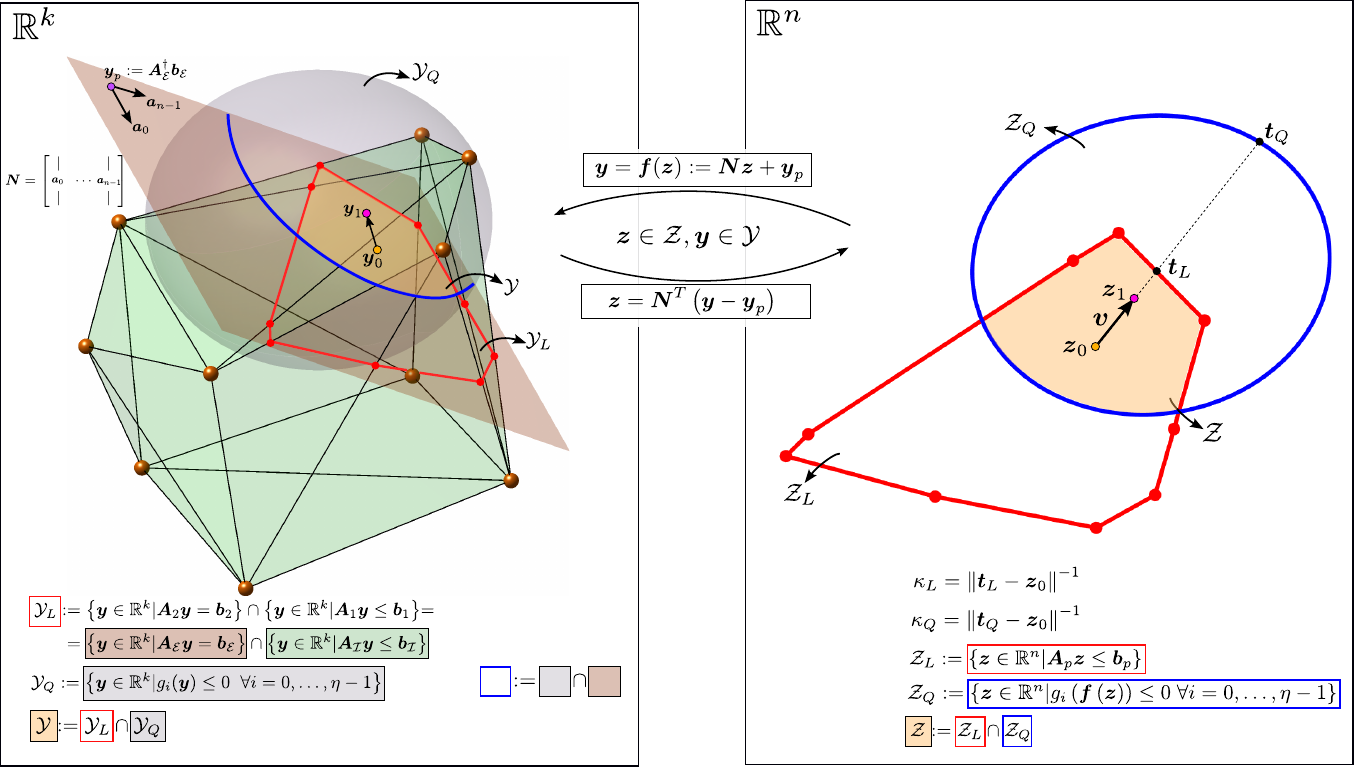}
\par\end{centering}
\caption{Sets $\mathcal{Y}\subseteq\mathbb{R}^{k}$ and $\mathcal{Z}\subseteq\mathbb{R}^{n}$.
$\boldsymbol{z}_{0}$ is a point in the interior of $\mathcal{Z}$,
and any point $\boldsymbol{z}\in\mathcal{Z}$ is mapped to its corresponding
point $\boldsymbol{y}\in\mathcal{Y}$ using $\boldsymbol{y}=\boldsymbol{f}(\boldsymbol{z})$.
In this example, the dimension of the ambient space is $k=3$, while
the dimension of $\operatorname{aff}\left(\mathcal{Y}\right)$ is
$n=2$. For visualization purposes, here $\mathcal{Y}$ is defined
by only linear and quadratic constraints. The values of $\kappa_{L}$
and $\kappa_{Q}$ (inverse distances to, respectively, $\partial\mathcal{Z}_{L}$
and $\partial\mathcal{Z}_{Q}$ along the direction $\bar{\boldsymbol{v}}:=\frac{\boldsymbol{v}}{\left\Vert \boldsymbol{v}\right\Vert }$, see Section \ref{sec:Online}), are also shown. In this figure,
$\boldsymbol{y}_{0}:=\boldsymbol{f}(\boldsymbol{z}_{0})$ and $\boldsymbol{y}_{1}:=\boldsymbol{f}(\boldsymbol{z}_{1})$. \label{fig:polyhedron3d_and_projection}}

\vskip-2ex
\end{figure*}
\noindent where $\boldsymbol{P}_{i}\succeq\boldsymbol{0}\;\forall i=0,...,\eta-1$,
and where $\boldsymbol{F}_{0},...,\boldsymbol{F}_{k}$ are symmetric
matrices. This set $\mathcal{Y}$, which can be bounded or unbounded,
is defined by linear constraints (Eqs.~\ref{eq:linear_ineq} and~\ref{eq:linear_eq}),
convex quadratic constraints (Eq.~\ref{eq:convex_quad}), SOC constraints
(Eq.~\ref{eq:soc}), and LMI constraints (Eq. \ref{eq:lmi}, also
known as semidefinite constraints). Constraints~\ref{eq:linear_ineq},
\ref{eq:linear_eq}, \ref{eq:convex_quad}, and~\ref{eq:soc} could
also be written as an LMI constraint, but for notational convenience
throughout the paper (and without any loss of generality), we explicitly
distinguish them. Note also that several LMIs can be converted into
one LMI by simply stacking the matrices appropriately (see, e.g.,
\citet[Section 4.6.2]{boyd2004convex}). 

Let us also introduce the following definitions:
\begin{align*}
\mathcal{Y}_{L} & :=\left\{ \boldsymbol{y}\in\mathbb{R}^{k}|\boldsymbol{A}_{1}\boldsymbol{y}\le\boldsymbol{b}_{1}\right\} \cap\left\{ \boldsymbol{y}\in\mathbb{R}^{k}|\boldsymbol{A}_{2}\boldsymbol{y}=\boldsymbol{b}_{2}\right\} \\
\mathcal{Y}_{Q} & :=\left\{ \boldsymbol{y}\in\mathbb{R}^{k}|g_{i}\left(\boldsymbol{y}\right)\le0\;\forall i=0,\ldots,\eta-1\right\} \\
\mathcal{Y}_{S} & :=\left\{ \boldsymbol{y}\in\mathbb{R}^{k}|h_{j}\left(\boldsymbol{y}\right)\le0\;\forall j=0,\ldots,\mu-1\right\} \\
\mathcal{Y}_{M} & :=\left\{ \boldsymbol{y}\in\mathbb{R}^{k}|\boldsymbol{W}(\boldsymbol{y})\succeq\boldsymbol{0}\right\} 
\end{align*}
Hence, we have that $\mathcal{Y}:=\mathcal{Y}_{L}\cap\mathcal{Y}_{Q}\cap\mathcal{Y}_{S}\cap\mathcal{Y}_{M}$.
For simplicity, we will assume throughout the paper that $\text{aff}\left(\mathcal{Y}_{L}\right)=\text{aff}\left(\mathcal{Y}\right)$,
where $\text{aff}\left(\cdot\right)$ denotes the affine hull. Some
degenerate cases do not satisfy this assumption, but in those cases
the same set $\mathcal{Y}$ can be parametrized with a different set
of constraints for which this assumption holds.\footnote{An example of a degenerate case in 3D (i.e., $k=3$) with only linear
and quadratic constraints would be when $\mathcal{Y}_{L}$ is the
unit cube, and $\mathcal{Y}_{Q}$ is a cylinder tangent to one of
its faces such that $\mathcal{Y}=\mathcal{Y}_{L}\cap\mathcal{Y}_{Q}$
is the segment they have in common. By simply reparametrizing $\mathcal{Y}$
with only linear constraints (the segment itself), then the assumption
$\text{aff}\left(\mathcal{Y}_{L}\right)=\text{aff}\left(\mathcal{Y}\right)$
is satisfied. }

Overall, RAYEN works as follows (see also Fig.~\ref{fig:polyhedron3d_and_projection}):
In an \uline{offline phase} (Section~\ref{sec:Offline:-Computation-of}),
the affine hull of $\mathcal{Y}$ is obtained, and the constraints
are expressed in that linear subspace. Letting $\mathcal{Z}$ denote
the feasible set in this subspace, an interior point $\boldsymbol{z}_{0}$
of this set $\mathcal{Z}$ is also found. In the \uline{online
phase} (Section~\ref{sec:Online}), a linear map is applied to an
upstream latent variable of the network to obtain a vector $\boldsymbol{v}$
of the same dimension as the subspace, that will determine the step
direction from $\boldsymbol{z}_{0}$. The length $\lambda\ge0$ of
this step is adjusted to ensure that $\boldsymbol{z}_{0}+\lambda\frac{\boldsymbol{v}}{\left\Vert \boldsymbol{v}\right\Vert }$
lies in $\mathcal{Z}$. After lifting to the original dimension, the
output is therefore guaranteed to lie in $\mathcal{Y}$. \added{These offline and online steps are also detailed in Alg.~\ref{alg:rayen-algorithm}.}

\let\oldnl\nl%
\newcommand{\nonl}{\renewcommand{\nl}{\let\nl\oldnl}}%
\begin{algorithm}[h]
	\SetAlgoLined
	\nonl \added{\textbf{Offline:}} \\
	\hspace{0.5cm}\added{Compute affine hull of $\mathcal{Y}$ (Section~\ref{subsec:affine-hull-of})}\\
	\hspace{0.5cm}\added{Compute $\mathcal{Z}$ (Section~\ref{subsec:set_z})}\\
	\hspace{0.5cm}\added{Compute $\boldsymbol{z}_0$, an interior point of $\mathcal{Z}$ (Section~\ref{subsec:Interior-point-of})}\\
	\nonl \added{\textbf{Online} (both training and testing):} \\
	\hspace{0.5cm}\added{Pass the input through the NN to obtain $\boldsymbol{x}\in\mathbb{R}^m$}\\
	\hspace{0.5cm}\added{Pass $\boldsymbol{x}\in\mathbb{R}^m$ through $\text{L}(m,n)$ to obtain $\boldsymbol{v}\in\mathbb{R}^n$}\\
	\hspace{0.5cm}\added{Compute $\kappa$ (Table~\ref{tab:definition_kappas})}\\
	\hspace{0.5cm}\added{Compute $\boldsymbol{z}_1$ and  $\boldsymbol{y}:=\boldsymbol{f}\left(\boldsymbol{z}_1\right)$ (Eqs.~\ref{eq:f_definition} and  \ref{eq:z1_definition}) }\\
	\nonl \hspace{0.5cm}\added{{\footnotesize\texttt{(The output $\boldsymbol{y}$ is guaranteed to be in $\mathcal{Y}$)}}} \\
	\hspace{0.5cm}\added{\textbf{if} \textit{training} \textbf{then}}\\
	\hspace{1.0cm}\added{Compute loss and backpropagate}
	\caption{\added{Pseudocode of the steps involved in RAYEN. Here, L$(a,b)$ denotes a linear layer with input size $a$ and output size $b$. The online phase is also depicted in Fig.~\ref{fig:pipeline}.} \label{alg:rayen-algorithm}}
\end{algorithm}

\section{RAYEN: Offline \label{sec:Offline:-Computation-of}}

\subsection{Affine Hull of $\mathcal{Y}$ \label{subsec:affine-hull-of}}

To find the affine hull of the set $\mathcal{Y}$, we first express
all the linear constraints (Eqs. \ref{eq:linear_ineq} and \ref{eq:linear_eq})
as linear inequality constraints by simply stacking the matrices:
\begin{align*}
\tilde{\boldsymbol{A}}:= & \arraycolsep=1.2pt\left[\begin{array}{ccc}
\boldsymbol{A}_{1}^{T} & \boldsymbol{A}_{2}^{T} & -\boldsymbol{A}_{2}^{T}\end{array}\right]^{T}\\
\tilde{\boldsymbol{b}}:= & \arraycolsep=1.2pt\left[\begin{array}{ccc}
\boldsymbol{b}_{1}^{T} & \boldsymbol{b}_{2}^{T} & -\boldsymbol{b}_{2}^{T}\end{array}\right]^{T}
\end{align*}
and therefore $\mathcal{Y}_{L}$ is now $\left\{ \boldsymbol{y}\in\mathbb{R}^{k}|\tilde{\boldsymbol{A}}\boldsymbol{y}\le\tilde{\boldsymbol{b}}\right\} $.
To reduce the computation time during the online phase, the redundant
constraints of the inequality system $\tilde{\boldsymbol{A}}\boldsymbol{y}\le\tilde{\boldsymbol{b}}$
(if there are any) are then deleted. This is done by solving a sequence
of Linear Programs (see, e.g. \citet[Eq. 1.5]{szedlak2017redundancy}),
which obtains the reduced system of inequalities $\boldsymbol{A}\boldsymbol{y}\le\boldsymbol{b}$
such that $\mathcal{Y}_{L}=\left\{ \boldsymbol{y}\in\mathbb{R}^{k}|\boldsymbol{A}\boldsymbol{y}\le\boldsymbol{b}\right\} $.
Then, another sequence of Linear Programs is solved to find the set
\[
\mathcal{E}:=\left\{ i|\boldsymbol{A}_{[i,:]}\boldsymbol{y}=\boldsymbol{b}_{[i]}\;\forall\boldsymbol{y}\in\mathcal{Y}_{L}\right\} \;,
\]
which contains the indexes of the constraints that are always active
for all the points in $\mathcal{Y}_{L}$ (see, e.g. \cite[Alg. 1.3.1]{borrelli2002discrete}
or \cite[Section 5.2]{jones2005polyhedral}). Defining $\mathcal{I}:=\{0,...,a-1\}\backslash\mathcal{E}$
(where $a$ is the number of rows of $\boldsymbol{A}$), let us now
introduce the following notation:\footnote{Note that if $\mathcal{I}=\emptyset$, then we need $\boldsymbol{b}_{\mathcal{I}}=1$
(instead of $\boldsymbol{b}_{\mathcal{I}}=0$) to make sure that there
exists a point in the interior of $\mathcal{Z}_{L}:=\left\{ \boldsymbol{z}\in\mathbb{R}^{n}|\boldsymbol{0}_{1\times n}\boldsymbol{z}\le1\right\} \equiv\mathbb{R}^{n}$
(see Section \ref{subsec:Interior-point-of}).}
\[
\boldsymbol{A}_{\mathcal{E}}:=\begin{cases}
\boldsymbol{0}_{1\times k} & \mathcal{E}=\emptyset\\
\boldsymbol{A}_{[\mathcal{E},:]} & \text{otherwise}
\end{cases},\;\boldsymbol{b}_{\mathcal{E}}:=\begin{cases}
0 & \mathcal{E}=\emptyset\\
\boldsymbol{b}_{[\mathcal{E}]} & \text{otherwise}
\end{cases}
\]

\[
\boldsymbol{A}_{\mathcal{I}}:=\begin{cases}
\boldsymbol{0}_{1\times k} & \mathcal{I}=\emptyset\\
\boldsymbol{A}_{[\mathcal{I},:]} & \text{otherwise}
\end{cases},\;\boldsymbol{b}_{\mathcal{I}}:=\begin{cases}
1 & \mathcal{I}=\emptyset\\
\boldsymbol{b}_{[\mathcal{I}]} & \text{otherwise}
\end{cases}
\]
Then, the affine hull of $\mathcal{Y}$ is given by
\[
\text{aff}\left(\mathcal{Y}\right)=\left\{ \boldsymbol{y}\in\mathbb{R}^{k}|\boldsymbol{A}_{\mathcal{E}}\boldsymbol{y}=\boldsymbol{b}_{\mathcal{E}}\right\} \;,
\]
where we have used the assumption that $\text{aff}\left(\mathcal{Y}\right)=\text{aff}\left(\mathcal{Y}_{L}\right)$
(see Section~\ref{sec:Problem-Setup}). The dimension of this affine
hull is then
\begin{equation}
n:=k-\text{rank}\left(\boldsymbol{A}_{\mathcal{E}}\right)\;.\label{eq:def_of_n}
\end{equation}
Note also that 
\[
\mathcal{Y}_{L}=\underbrace{\left\{ \boldsymbol{y}\in\mathbb{R}^{k}|\boldsymbol{A}_{\mathcal{E}}\boldsymbol{y}=\boldsymbol{b}_{\mathcal{E}}\right\} }_{=\text{aff}\left(\mathcal{Y}_{L}\right)=\text{aff}\left(\mathcal{Y}\right)}\cap\left\{ \boldsymbol{y}\in\mathbb{R}^{k}|\boldsymbol{A}_{\mathcal{I}}\boldsymbol{y}\le\boldsymbol{b}_{\mathcal{I}}\right\} \;.
\]
In general, $\left(\boldsymbol{A}_{\mathcal{E}},\boldsymbol{b}_{\mathcal{E}}\right)$
may be different than $\left(\boldsymbol{A}_{2},\boldsymbol{b}_{2}\right)$,
and $\left(\boldsymbol{A}_{\mathcal{I}},\boldsymbol{b}_{\mathcal{I}}\right)$
may be different than $\left(\boldsymbol{A}_{1},\boldsymbol{b}_{1}\right)$.
This can happen, for example, when $\text{aff}\left(\left\{ \boldsymbol{y}\in\mathbb{R}^{k}|\boldsymbol{A}_{1}\boldsymbol{y}\le\boldsymbol{b}_{1}\right\} \right)\neq\mathbb{R}^{k}$
(intuitively this means that the inequality constraint defined by Eq.~\ref{eq:linear_ineq}
\textit{hides} equality constraints), or when there are redundant constraints
in Eqs.~\ref{eq:linear_ineq} and/or~\ref{eq:linear_eq}.

\subsection{Set $\mathcal{Z}$ \label{subsec:set_z}}

Any point in $\boldsymbol{y}\in\text{aff}\left(\mathcal{Y}\right)$
can be expressed as 
\begin{equation}
\boldsymbol{y}=\boldsymbol{f}(\boldsymbol{z}):=\boldsymbol{N}\boldsymbol{z}+\underbrace{\boldsymbol{A}_{\mathcal{E}}^{\dagger}\boldsymbol{b}_{\mathcal{E}}}_{:=\boldsymbol{y}_{p}}\;,\label{eq:f_definition}
\end{equation}
where $\boldsymbol{z}\in\mathbb{R}^{n}$, $\boldsymbol{N}$ is a $k\times n$
matrix whose columns form an orthonormal basis for the null space
of $\boldsymbol{A}_{\mathcal{E}}$, and $\boldsymbol{A}_{\mathcal{E}}^{\dagger}$
is the pseudoinverse of $\boldsymbol{A}_{\mathcal{E}}$. Using Eq.
\ref{eq:f_definition}, we have that $\boldsymbol{A}_{\mathcal{I}}\boldsymbol{y}\le\boldsymbol{b}_{\mathcal{I}}$
is equivalent to:
\[
\boldsymbol{A}_{\mathcal{I}}\left(\boldsymbol{N}\boldsymbol{z}+\boldsymbol{y}_{p}\right)\le\boldsymbol{b}_{\mathcal{I}}
\]
\begin{equation}
\underbrace{\boldsymbol{A}_{\mathcal{I}}\boldsymbol{N}}_{:=\boldsymbol{A}_{p}}\boldsymbol{z}\le\underbrace{\boldsymbol{b}_{\mathcal{I}}-\boldsymbol{A}_{\mathcal{I}}\boldsymbol{y}_{p}}_{:=\boldsymbol{b}_{p}}\label{eq:Ap_and_bp}
\end{equation}

Defining now these sets (see also Fig. \ref{fig:polyhedron3d_and_projection}):
\begin{align*}
\mathcal{Z}_{L}:= & \left\{ \boldsymbol{z}\in\mathbb{R}^{n}|\boldsymbol{A}_{p}\boldsymbol{z}\le\boldsymbol{b}_{p}\right\} \\
\mathcal{Z}_{Q_{i}}:= & \left\{ \boldsymbol{z}\in\mathbb{R}^{n}|g_{i}\left(\boldsymbol{f}(\boldsymbol{z})\right)\le0\right\}  & \mathcal{Z}_{Q}:=\bigcap_{i=0}^{\eta-1}\mathcal{Z}_{Q_{i}}\\
\mathcal{Z}_{S_{j}}:= & \left\{ \boldsymbol{z}\in\mathbb{R}^{n}|h_{j}\left(\boldsymbol{f}(\boldsymbol{z})\right)\le0\right\}  & \mathcal{Z}_{S}:=\bigcap_{j=0}^{\mu-1}\mathcal{Z}_{S_{j}}\\
\mathcal{Z}_{M}:= & \left\{ \boldsymbol{z}\in\mathbb{R}^{n}|\boldsymbol{W}\left(\boldsymbol{f}(\boldsymbol{z})\right)\succeq\boldsymbol{0}\right\} \\
\mathcal{Z}:= & \mathcal{Z}_{L}\cap\mathcal{Z}_{Q}\cap\mathcal{Z}_{S}\cap\mathcal{Z}_{M}
\end{align*}
Any point $\boldsymbol{z}\in\mathcal{Z}$ can therefore be mapped to
the corresponding point $\boldsymbol{y}\in\mathcal{Y}$ using 
Eq. \ref{eq:f_definition}.

\subsection{Interior point of $\mathcal{Z}$\label{subsec:Interior-point-of}}

\added{
An interior point $\boldsymbol{z}_0$ of the set $\mathcal{Z}$ is the one that satisfies these strict inequality constraints:
\begin{align}
\begin{split} \label{eq:constraints_interior}
& \boldsymbol{A}_{p}\boldsymbol{z}_0 < \boldsymbol{b}_{p}\\
& g_{i}\left(\boldsymbol{f}\left(\boldsymbol{z}_0\right)\right) < 0 \;\;i=0,...,\eta-1\\
& h_{j}\left(\boldsymbol{f}\left(\boldsymbol{z}_0\right)\right) < 0 \;\;j=0,...,\mu-1\\
& \boldsymbol{W}\left(\boldsymbol{f}(\boldsymbol{z}_0)\right)\succ \boldsymbol{0}
\end{split}
\end{align}
}

\added{Hence, a way to find this interior point $\boldsymbol{z}_{0}$ is to solve the convex program:} 
\begin{align*}
\left(\boldsymbol{z}_{0},\epsilon^{*}\right)=\underset{\boldsymbol{z},\epsilon}{\text{argmax}\text{\ensuremath{\quad}}} & \epsilon\\
\text{s.t.:}\text{\ensuremath{\quad}} & \boldsymbol{A}_{p}\boldsymbol{z}-\boldsymbol{b}_{p}\le-\epsilon\boldsymbol{1}\\
 & g_{i}\left(\boldsymbol{f}\left(\boldsymbol{z}\right)\right)\le-\epsilon\;\;i=0,...,\eta-1\\
 & h_{j}\left(\boldsymbol{f}\left(\boldsymbol{z}\right)\right)\le-\epsilon\;\;j=0,...,\mu-1\\
 & \boldsymbol{W}\left(\boldsymbol{f}(\boldsymbol{z})\right)\succeq\epsilon\boldsymbol{I}\\
 & 0 \le \epsilon \added{\le \delta}
\end{align*}

\added{and then checking that $\epsilon^*>0$. In these constraints, $\epsilon \ge 0$ is a decision variable that dictates how much slack we have while satisfying each inequality constraint. By maximizing $\epsilon$, we ensure that the strict inequalities of Eq.~\ref{eq:constraints_interior} are satisfied. $\delta$ is a fixed positive parameter that simply prevents this optimization problem from being unbounded, which happens when the feasible set $\mathcal{Z}$ is unbounded. We choose $\delta=0.5$.   }

\section{RAYEN: Online \label{sec:Online}}
\begin{figure}[h]
	\begin{centering}
		\includegraphics[width=1\columnwidth]{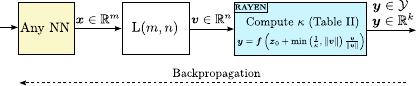}
		\par\end{centering}
	\caption{Neural network equipped with RAYEN. $\kappa$~is computed as shown
		in Table \ref{tab:definition_kappas}, and the map $\boldsymbol{f}\left(\boldsymbol{z}_{0}+\text{min}\left(\frac{1}{\kappa},\left\Vert \boldsymbol{v}\right\Vert \right)\bar{\boldsymbol{v}}\right)$
		is used to guarantee $\boldsymbol{y}\in\mathcal{Y}$. Here, L($m,n$)
		denotes a linear layer with input size $m$ and output size $n$.
		This linear layer is not needed if $m=n$. After the RAYEN module,
		there may be more downstream layers. \added{This same pipeline is used for both training and testing.}  During the training procedure, gradients
		\added{are} backpropagated through RAYEN. \label{fig:pipeline}}
	
	\vskip-2ex
\end{figure}

\selectlanguage{american}%
\begin{figure*}
\begin{centering}
\includegraphics[width=1\textwidth]{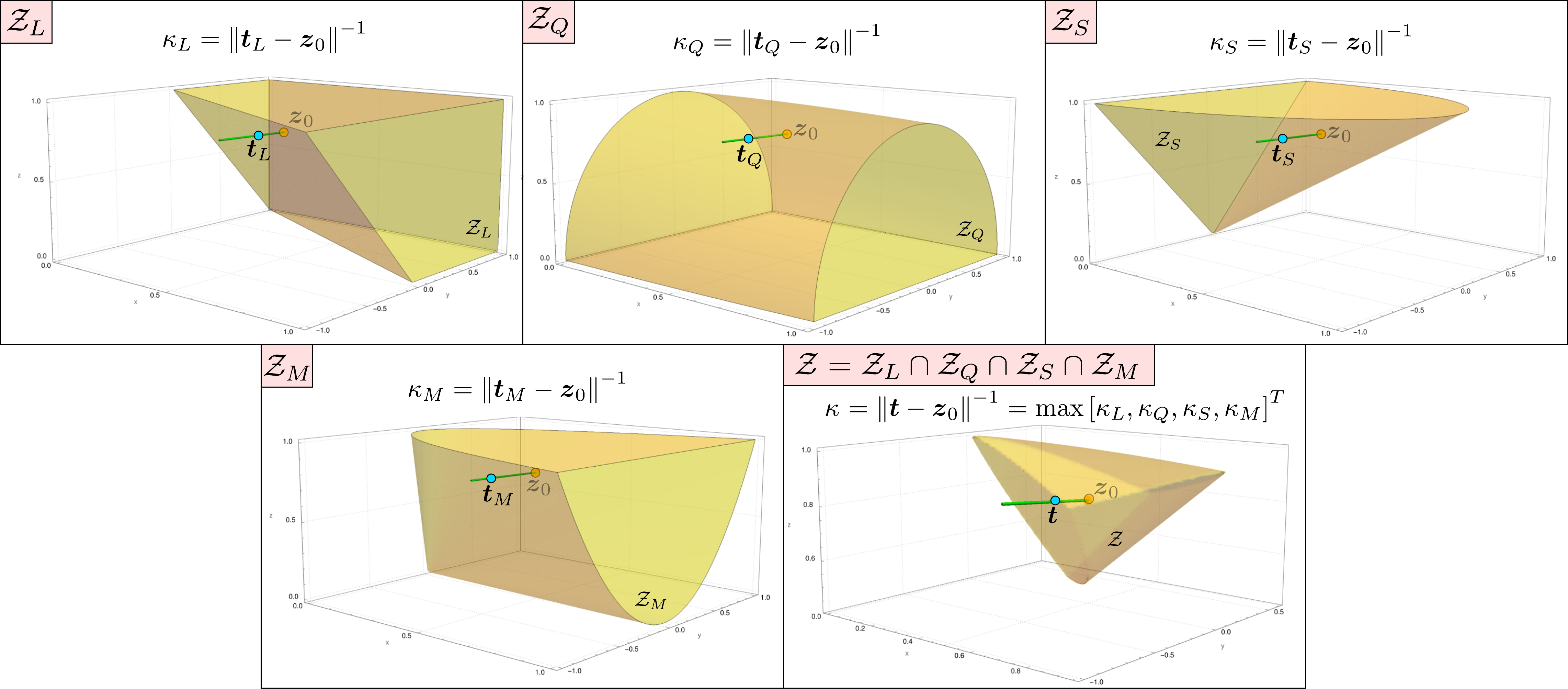}
\par\end{centering}
\caption{Geometric meaning of $\kappa_{L}$, $\kappa_{Q}$, $\kappa_{S}$,
$\kappa_{M}$, and $\kappa$ for a given value of $\bar{\boldsymbol{v}}$.
The green segment indicates the direction of $\bar{\boldsymbol{v}}$.
In this example, $n=3$, $\eta=1$ (i.e., only one quadratic constraint),
and $\mu=1$ (i.e., only one SOC constraint). Here, $\boldsymbol{t}$ ($\boldsymbol{t}_L$, $\boldsymbol{t}_Q$, $\boldsymbol{t}_S$, $\boldsymbol{t}_M$) denotes the intersection between the ray that starts at $\boldsymbol{z}_0$ and follows $\bar{\boldsymbol{v}}$ with $\partial\mathcal{Z}$ ($\partial\mathcal{Z}_L$, $\partial\mathcal{Z}_Q$, $\partial\mathcal{Z}_S$, $\partial\mathcal{Z}_M$ respectively). These rays are the ones that give the name to the algorithm (RAYEN).  \label{fig:examples_all_kappas}}

\vskip-2ex
\end{figure*}
As shown in Fig. \ref{fig:pipeline}, to obtain an $n$-dimensional
latent variable $\boldsymbol{v}$ (where $n$ is the dimension of
$\text{aff}\left(\mathcal{Y}\right)$, see Eq. \ref{eq:def_of_n}),
we first apply a linear layer to the upstream latent variable of
the network $\boldsymbol{x}\in\mathbb{R}^{m}$. This mapping can be
omitted if $m=n$. Assuming now $\boldsymbol{v}\neq\boldsymbol{0}$,\footnote{If $\boldsymbol{v}=\boldsymbol{0}$, \foreignlanguage{english}{we
simply take $\boldsymbol{z}_{1}:=\boldsymbol{z}_{0}$, and therefore
$\kappa$ does not need to be computed.}} let us define the inverse distance to the frontier of $\mathcal{Z}$
along $\bar{\boldsymbol{v}}:=\frac{\boldsymbol{v}}{\left\Vert \boldsymbol{v}\right\Vert }$
as
\[
\kappa:=\text{inf}\left\{ \lambda^{-1}|\boldsymbol{z}_{0}+\lambda\bar{\boldsymbol{v}}\in\mathcal{Z},\;\lambda>0\right\} \;.
\]
Then, it is clear that
\begin{equation}
\boldsymbol{z}_{1}:=\left(\boldsymbol{z}_{0}+\text{min}\left(\frac{1}{\kappa},\left\Vert \boldsymbol{v}\right\Vert \right)\bar{\boldsymbol{v}}\right)\in\mathcal{Z}\;,\label{eq:z1_definition}
\end{equation}
 and that therefore $\boldsymbol{f}\left(\boldsymbol{z}_{1}\right)\in\mathcal{Y}$.
Note that no conservatism is introduced here. In other words, and
given that $\mathcal{Y}$ is a convex set, for any point $\boldsymbol{y}\in\mathcal{Y}$,
there exists at least one $\boldsymbol{v}$ such that $\boldsymbol{f}\left(\boldsymbol{z}_{1}\right)\in\mathcal{Y}$
holds.
To obtain $\kappa$, first note that \foreignlanguage{english}{
\[
\arraycolsep=1.2pt\kappa=\text{max}\left[\begin{array}{cccc}
\kappa_{L} & \kappa_{Q} & \kappa_{S} & \kappa_{M}\end{array}\right]^{T}\;,
\]
where $\kappa_{L}$, }$\kappa_{Q}$, $\kappa_{S}$, and $\kappa_{M}$
are defined in Table \ref{tab:definition_kappas} and are, respectively,
the inverses of the distances to $\mathcal{\partial Z}_{L}$, $\mathcal{\partial Z}_{Q}$,
$\mathcal{\partial Z}_{S}$, and $\mathcal{\partial Z}_{M}$ along
$\bar{\boldsymbol{v}}$ (see also Figs. \ref{fig:polyhedron3d_and_projection}
and \ref{fig:examples_all_kappas}). In the following subsections
we explain how \foreignlanguage{english}{$\kappa_{L}$, }$\kappa_{Q}$,
$\kappa_{S}$, and $\kappa_{M}$ can be obtained.

\selectlanguage{english}%
\begin{table}
\caption{Definition of inverse distances \foreignlanguage{american}{$\kappa_{L}$},
\foreignlanguage{american}{$\kappa_{Q}$}, \foreignlanguage{american}{$\kappa_{S}$},
\foreignlanguage{american}{$\kappa_{M}$}, and~$\kappa$. By definition, all of them are nonnegative. In this Table, $\text{idist}\left(\partial\mathcal{S}\right):=$
\foreignlanguage{american}{$\text{inf}\left\{ \lambda^{-1}|\boldsymbol{z}_{0}+\lambda\bar{\boldsymbol{v}}\in\mathcal{S},\;\lambda>0\right\} $}.
\label{tab:definition_kappas}}

\selectlanguage{american}%
\noindent\resizebox{\columnwidth}{!}{%
\begin{centering}
\begin{tabular}{|c|c|c|c|}
\hline 
 & \textbf{Definition} & \textbf{Computation} & \textbf{Section}\tabularnewline
\hline 
\hline 
$\kappa_{L}$ & $\text{idist}\left(\partial\mathcal{Z}_{L}\right)$ & $\text{relu}\left(\max\left(\boldsymbol{D}\bar{\boldsymbol{v}}\right)\right)$ & \ref{subsec:Computation-of-k_L}\tabularnewline
\hline 
$\kappa_{Q}$ & $\text{idist}\left(\partial\mathcal{Z}_{Q}\right)$ & $\arraycolsep=0.8pt\text{max}\left(\left[\begin{array}{ccc}
\kappa_{Q,0}^{+} & \cdots & \kappa_{Q,\eta-1}^{+}\end{array}\right]^{T}\right)$ & \ref{subsec:Computation-of-k_Q}\tabularnewline
\hline 
$\kappa_{S}$ & $\text{idist}\left(\partial\mathcal{Z}_{S}\right)$ & \selectlanguage{english}%
$\arraycolsep=0.8pt\text{relu}\left(\text{max}\left(\left[\begin{array}{ccccc}
\kappa_{S,0}^{(0)} & \kappa_{S,0}^{(1)} & \cdots & \kappa_{S,\mu-1}^{(0)} & \kappa_{S,\mu-1}^{(1)}\end{array}\right]^{T}\right)\right)$\selectlanguage{american}%
 & \ref{subsec:Computation-of-k_S}\tabularnewline
\hline 
$\kappa_{M}$ & $\text{idist}\left(\partial\mathcal{Z}_{M}\right)$ & \selectlanguage{english}%
$\arraycolsep=1.2pt\text{relu}\left(\text{max}\left(\text{eig}\left(\boldsymbol{L}^{T}\left(-\boldsymbol{S}\right)\boldsymbol{L}\right)\right)\right)$\selectlanguage{american}%
 & \ref{subsec:Computation-of-k_M}\tabularnewline
\hline 
$\kappa$ & $\text{idist}\left(\partial\mathcal{Z}\right)$ & \selectlanguage{english}%
$\arraycolsep=1.2pt\text{max}\left[\begin{array}{cccc}
\kappa_{L} & \kappa_{Q} & \kappa_{S} & \kappa_{M}\end{array}\right]^{T}$\selectlanguage{american}%
 & -\tabularnewline
\hline 
\end{tabular}
\par\end{centering}
}

\selectlanguage{english}%
\end{table}

\subsection{Computation of $\kappa_{L}$ \label{subsec:Computation-of-k_L}}

Given a face $\mathcal{F}_{j}:=\left\{ \boldsymbol{z}\in\mathbb{R}^{n}|\boldsymbol{A}_{p_{[j,:]}}\boldsymbol{z}=\boldsymbol{b}_{p_{[j]}}\right\} $
of the polyhedron $\mathcal{Z}_{L}$, the inverse distance from $\boldsymbol{z}_{0}$
to $\mathcal{F}_{j}$ along the direction $\bar{\boldsymbol{v}}$
can be easily obtained as 
\[
\kappa_{L,j}=\text{relu}\left(\frac{\boldsymbol{A}_{p_{[j,:]}}\bar{\boldsymbol{v}}}{\boldsymbol{b}_{p_{[j]}}-\boldsymbol{A}_{p_{[j,:]}}\boldsymbol{z}_{0}}\right)\;,
\]
where the $\text{relu}(\cdot)$ operator is needed to enforce moving
along $\bar{\boldsymbol{v}}$. Taking into account now all the
faces of $\mathcal{Z}_{L}$, we have that $\kappa_{L}$ is given by
\[
\kappa_{L}:=\text{relu}\left(\max\left(\underbrace{\boldsymbol{A}_{p}\varoslash\left(\left(\boldsymbol{b}_{p}-\boldsymbol{A}_{p}\boldsymbol{z}_{0}\right)\boldsymbol{1}_{1\times n}\right)}_{:=\boldsymbol{D}}\bar{\boldsymbol{v}}\right)\right)\;,
\]
where $\varoslash$ denotes the element-wise division between two
matrices, and where $\boldsymbol{D}$ can be computed offline because
it does not depend on $\bar{\boldsymbol{v}}$. Note that if $\kappa_{L}=0$, then
$\mathcal{Z}_L$ is unbounded along the ray that starts at $\boldsymbol{z}_{0}$
and follows $\bar{\boldsymbol{v}}$.
\selectlanguage{american}%

\subsection{Computation of $\kappa_{Q}$ \label{subsec:Computation-of-k_Q}}

The inverse of the distance from $\boldsymbol{z}_{0}$ to $\partial\mathcal{Z}_{Q_{i}}$
along the direction $\bar{\boldsymbol{v}}$ can be obtained by computing
the nonnegative $\kappa_{Q,i}$ that satisfies:
\[
g_{i}\left(\boldsymbol{f}\left(\boldsymbol{z}_{0}+\frac{1}{\kappa_{Q,i}}\bar{\boldsymbol{v}}\right)\right)=0\;,
\]
where $\boldsymbol{f}(\cdot)$ is defined in Eq. \ref{eq:f_definition}.
Multiplying both sides of this equation by $\kappa_{Q,i}^{2}$ yields
a quadratic equation on $\kappa_{Q,i}$, from which its nonnegative
root $\kappa_{Q,i}^{+}$ can be easily obtained. Taking now into account
all the quadratic constraints, $\kappa_{Q}$ is therefore given by:
\[
\arraycolsep=1.2pt\kappa_{Q}:=\text{max}\left(\left[\begin{array}{ccc}
\kappa_{Q,0}^{+} & \cdots & \kappa_{Q,\eta-1}^{+}\end{array}\right]^{T}\right)
\]

\begin{figure}[t]
\begin{centering}
\includegraphics[width=1\columnwidth]{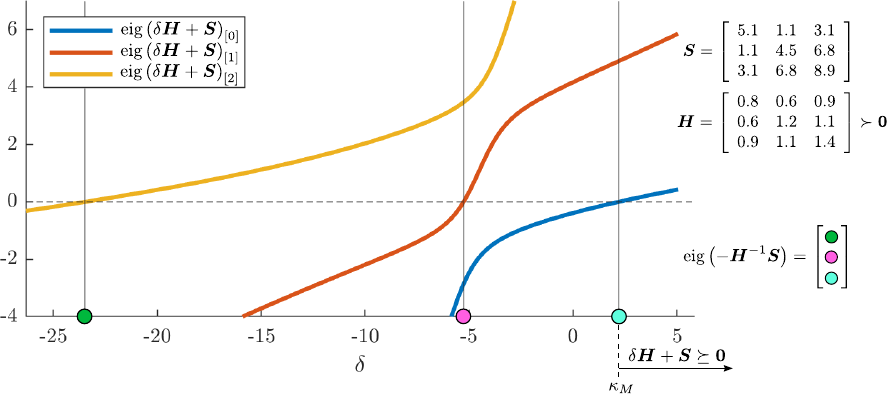}
\par\end{centering}
\caption{Eigenvalues of $\delta\boldsymbol{H}+\boldsymbol{S}$ as a function of $\delta$. In this example, $\boldsymbol{H}\succ\boldsymbol{0}$ and $\boldsymbol{S}$ are $3\times3$ matrices. $\kappa_M$, defined in Eq.~\ref{eq:definitionkM}, is also shown here. \label{fig:eigenvalues}}

\vskip-2ex
\end{figure}

\subsection{Computation of $\kappa_{S}$ \label{subsec:Computation-of-k_S}}

Similar to the previous case, the inverse of the distance from $\boldsymbol{z}_{0}$
to $\partial\mathcal{Z}_{S_{j}}$ along the direction $\bar{\boldsymbol{v}}$
can be obtained by computing the nonnegative $\kappa_{S,j}$ that
satisfies:
\[
h_{j}\left(\boldsymbol{f}\left(\boldsymbol{z}_{0}+\frac{1}{\kappa_{S,j}}\bar{\boldsymbol{v}}\right)\right)=0
\]

\begin{equation}
\left\Vert \boldsymbol{M}_{j}\boldsymbol{f}\left(\boldsymbol{z}_{0}+\frac{1}{\kappa_{S,j}}\bar{\boldsymbol{v}}\right)+\boldsymbol{s}_{j}\right\Vert =\boldsymbol{c}_{j}^{T}\boldsymbol{f}\left(\boldsymbol{z}_{0}+\frac{1}{\kappa_{S,j}}\bar{\boldsymbol{v}}\right)+d_{j}\label{eq:computation_kS}
\end{equation}

Squaring both sides of Eq. \ref{eq:computation_kS}, and then multiplying
them by $\kappa_{S,j}^{2}$ yields a quadratic equation on $\kappa_{S,j}$,
from which its two roots $\kappa_{S,j}^{(0)}$ and $\kappa_{S,j}^{(1)}$
can be easily obtained. The inverse of the distance from $\boldsymbol{z}_{0}$
to $\partial\mathcal{Z}_{S_{j}}$ along $\bar{\boldsymbol{v}}$ will
therefore be given by\footnote{The $\text{relu}(\cdot)$ operator is needed because both
roots can be negative due to the fact that we have squared both sides
of Eq. \ref{eq:computation_kS}. Both roots being negative means that
in that case $\mathcal{Z}_{S_{j}}$ is unbounded in the direction
$\bar{\boldsymbol{v}}$ and therefore we have $\kappa_{S,j}=0$. When
both roots are positive, we need to select the largest one, which is
why the $\text{max}\left(\cdot\right)$ operator is needed. }
 $$\text{relu}\left(\text{max}\left(\left[\begin{array}{cc}
\kappa_{S,j}^{(0)} & \kappa_{S,j}^{(1)}\end{array}\right]^{T}\right)\right)\;.$$
Taking into account all the SOC constraints, $\kappa_{S}$ is therefore
given by
\[
\arraycolsep=1.2pt\kappa_{S}:=\text{relu}\left(\text{max}\left(\left[\begin{array}{ccccc}
\kappa_{S,0}^{(0)} & \kappa_{S,0}^{(1)} & \cdots & \kappa_{S,\mu-1}^{(0)} & \kappa_{S,\mu-1}^{(1)}\end{array}\right]^{T}\right)\right)\;.
\]

\subsection{Computation of $\kappa_{M}$ \label{subsec:Computation-of-k_M}}

The inverse of the distance from $\boldsymbol{z}_{0}$ to $\partial\mathcal{Z}_{M}$
along the direction $\bar{\boldsymbol{v}}$ is defined as:
\[
\kappa_{M}:=\text{inf}\left\{ \lambda^{-1}|\boldsymbol{z}_{0}+\lambda\bar{\boldsymbol{v}}\in\mathcal{Z}_{M},\;\lambda>0\right\} 
\]
Noting that
\[
\boldsymbol{f}\left(\boldsymbol{z}_{0}+\lambda\bar{\boldsymbol{v}}\right)=\lambda\underbrace{\boldsymbol{N}\bar{\boldsymbol{v}}}_{:=\boldsymbol{\rho}}+\underbrace{\boldsymbol{N}\boldsymbol{z}_{0}+\boldsymbol{y}_{p}}_{:=\boldsymbol{y}_{0}}=\left(\lambda\boldsymbol{\rho}+\boldsymbol{y}_{0}\right)\;,
\]
then the condition $\boldsymbol{z}_{0}+\lambda\bar{\boldsymbol{v}}\in\mathcal{Z}_{M}$
is equivalent to
\[
\boldsymbol{W}\left(\boldsymbol{f}\left(\boldsymbol{z}_{0}+\lambda\bar{\boldsymbol{v}}\right)\right)=\boldsymbol{F}_{k}+\sum_{\alpha=0}^{k-1}\left(\lambda\boldsymbol{\rho}_{[\alpha]}+\boldsymbol{y}_{0_{[\alpha]}}\right)\boldsymbol{F}_{\alpha}\succeq\boldsymbol{0}
\]
Dividing everything by $\lambda$ (recall that $\lambda>0$) and rearranging
the terms we have
\[
\lambda^{-1}\underbrace{\left(\boldsymbol{F}_{k}+\sum_{\alpha=0}^{k-1}\left(\boldsymbol{y}_{0_{[\alpha]}}\boldsymbol{F}_{\alpha}\right)\right)}_{:=\boldsymbol{H}\succ\boldsymbol{0}}+\underbrace{\sum_{\alpha=0}^{k-1}\left(\boldsymbol{\rho}_{[\alpha]}\boldsymbol{F}_{\alpha}\right)}_{:=\boldsymbol{S}}\succeq\boldsymbol{0}\;,
\]
where $\boldsymbol{H}\succ\boldsymbol{0}$ derives from the fact that $\boldsymbol{z}_{0}$
is an interior point of $\mathcal{Z}_{M}$. Hence:
\begin{equation}
\kappa_{M}:=\text{inf}\left\{ \delta|\delta\boldsymbol{H}+\boldsymbol{S}\succeq\boldsymbol{0},\;\delta>0\right\} \label{eq:definitionkM}
\end{equation}

Let us now distinguish two cases:
\begin{itemize}
\item If $\boldsymbol{S}\succeq\boldsymbol{0}$, then it is clear that $\kappa_{M}=0$. 
\item Otherwise, we need $\kappa_{M}\boldsymbol{H}+\boldsymbol{S}\in\partial\mathcal{Z}_{M}$. \added{Note that the matrices that belong to $\partial\mathcal{Z}_{M}$
	have at least one zero eigenvalue. Therefore, defining
	$$
	\mathcal{T}:=\left\{\tau_{i}|\left(\tau_{i}\boldsymbol{H}+\boldsymbol{S}\right)\boldsymbol{v}_{i}=0\boldsymbol{v}_{i},\boldsymbol{v}_{i}\neq\boldsymbol{0}\right\}
	$$
	(where $\boldsymbol{v}_{i}$ is the eigenvector associated with the zero eigenvalue of $\tau_{i}\boldsymbol{H}+\boldsymbol{S}$), we know that $\kappa_{M}\in \mathcal{T}$. Also note that the rest of the eigenvalues need to be positive. Given that $\boldsymbol{H} \succ \boldsymbol{0}$, we can conclude that
\[
\kappa_{M}=\text{max}\left( \mathcal{T} \right)\;.
\]}
\added{An example is available in Fig.~\ref{fig:eigenvalues}.} 
Note also that: 
\begin{equation}
\left(\tau_{i}\boldsymbol{H}+\boldsymbol{S}\right)\boldsymbol{v}_{i}=0\boldsymbol{v}_{i}\iff-\boldsymbol{H}^{-1}\boldsymbol{S}\boldsymbol{v}_{i}=\tau_{i}\boldsymbol{v}_{i}\label{eq:definition_eig_mHinvS}
\end{equation}

Letting $\boldsymbol{L}$ denote the lower triangular matrix with
positive diagonal entries such that $\boldsymbol{H}^{-1}=\boldsymbol{L}\boldsymbol{L}^{T}$
(i.e., the Cholesky decomposition of $\boldsymbol{H}^{-1}$), and
using the fact that $\text{eig}\left(-\boldsymbol{H}^{-1}\boldsymbol{S}\right)=\text{eig}\left(\boldsymbol{L}^{T}\left(-\boldsymbol{S}\right)\boldsymbol{L}\right)$
(see proof in Appendix \ref{subsec:Eigenvalues-of-mHinvS}), we have
that\footnote{Eq. \ref{eq:definition_eig_mHinvS} can also
be written as $-\boldsymbol{S}\boldsymbol{v}_{i}=\tau_{i}\boldsymbol{H}\boldsymbol{v}_{i}$, and hence $\kappa_{M}$ could also be found solving the generalized
eigenvalue problem \cite{ghojogh2019eigenvalue,boyd2004convex} of
the pair $\left(-\boldsymbol{S},\boldsymbol{H}\right)$.}
\[
\kappa_{M}=\text{max}\left(\text{eig}\left(\boldsymbol{L}^{T}\left(-\boldsymbol{S}\right)\boldsymbol{L}\right)\right)\;.
\]

\end{itemize}
Putting everything together, we can conclude that:\footnote{Note that Eq. \ref{eq:k_M_definition} gives $\kappa_{M}=0$ for the
case $\boldsymbol{S}\succeq\boldsymbol{0}$. This can be easily proven
as follows: As the eigenvalues of the product of two positive semidefinite
matrices are nonnegative (see, e.g., \cite{zhan2004matrix}), then,
if $\boldsymbol{S}\succeq\boldsymbol{0}$, we have that $\text{eig}\left(-\boldsymbol{H}^{-1}\boldsymbol{S}\right)=\text{eig}\left(\boldsymbol{L}^{T}\left(-\boldsymbol{S}\right)\boldsymbol{L}\right)\leq\boldsymbol{0}$
. And hence, Eq.~\ref{eq:k_M_definition} gives $\kappa_{M}=0$.}
\begin{equation}
\kappa_{M}=\text{relu}\left(\text{max}\left(\text{eig}\left(\boldsymbol{L}^{T}\left(-\boldsymbol{S}\right)\boldsymbol{L}\right)\right)\right)\;.\label{eq:k_M_definition}
\end{equation}
As only the maximum eigenvalue is needed, we can use methods such
as power iteration \cite{muntz1913sur,muntz1913solution,golub2013matrix},
LOBPCG \cite{knyazev2001toward}, or the Lanczos algorithm \cite{lanczos1950iteration,ojalvo1970vibration}
to avoid computing the whole spectrum of the matrix. %
Note also that the matrix $\boldsymbol{L}$ can be computed offline,
since it does not depend on $\boldsymbol{v}$. 
\selectlanguage{american}%
\begin{figure*}
\begin{centering}
\includegraphics[width=1\textwidth]{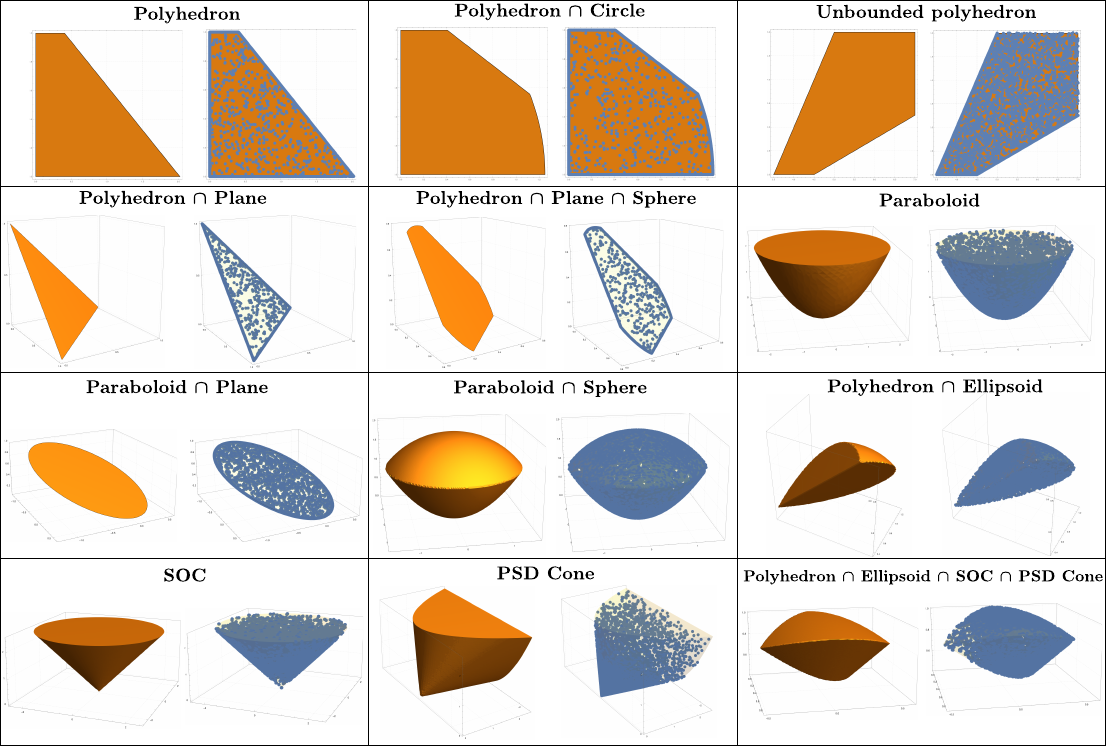}
\par\end{centering}
\caption{RAYEN applied to 12 different constraints. \foreignlanguage{english}{For
each of the constraints, we generate 12K samples $\boldsymbol{v}\in\mathbb{R}^{n}$
sampled uniformly in the box $[-2.5,2.5]^{n}$, compute the corresponding
$\kappa$, and plot the resulting $\boldsymbol{y}=\boldsymbol{f}\left(\boldsymbol{z}_{0}+\text{min}\left(\frac{1}{\kappa},\left\Vert \boldsymbol{v}\right\Vert \right)\bar{\boldsymbol{v}}\right)\in\mathbb{R}^{k}$
for each of those samples (see also Fig. \ref{fig:pipeline}). For each of the subfigures, the left plot shows the feasible set $\mathcal{Y}$, while the right plot shows both $\mathcal{Y}$ and the output samples $\boldsymbol{y}$. Note
how all these output samples satisfy the constraints.
The constraints of the first row have $k=n=2$. The first two constraints
of the second row have $k=3$ and $n=2$. The rest of the constraints
have $k=n=3$. RAYEN works in any dimension, but, for visualization
purposes, only 2D and 3D examples are shown here. }\label{fig:all_constraints}}
\end{figure*}

\subsection{Remarks}

We can have these two cases in Eq. \ref{eq:z1_definition}: 
\begin{itemize}
\item If $\boldsymbol{z}_{0}+\boldsymbol{v}\in\mathcal{Z}$, then $\left\Vert \boldsymbol{v}\right\Vert \le\frac{1}{\kappa}$,
and therefore Eq. \ref{eq:z1_definition} becomes $\boldsymbol{z}_{1}=\boldsymbol{z}_{0}+\boldsymbol{v}$.
\item If $\boldsymbol{z}_{0}+\boldsymbol{v}\notin\mathcal{Z}$, then $\left\Vert \boldsymbol{v}\right\Vert >\frac{1}{\kappa}$
and therefore Eq. \ref{eq:z1_definition} becomes $\boldsymbol{z}_{1}=\boldsymbol{z}_{0}+\frac{1}{\kappa}\bar{\boldsymbol{v}}$.
This operation performs a (nonorthogonal) projection of $\boldsymbol{z}_{0}+\boldsymbol{v}$
onto $\mathcal{Z}$ along the ray that starts at $\boldsymbol{z}_{0}$
and follows $\bar{\boldsymbol{v}}$.
\end{itemize}
Note also that, if $\mathcal{Z}$ \foreignlanguage{english}{($\mathcal{Z}_{L}$,
}$\mathcal{Z}_{Q}$, $\mathcal{Z}_{S}$, $\mathcal{Z}_{M}$\foreignlanguage{english}{
respectively)} is unbounded along the ray that starts at $\boldsymbol{z}_{0}$
and follows $\bar{\boldsymbol{v}}$\foreignlanguage{english}{, then
$\kappa$ ($\kappa_{L}$, }$\kappa_{Q}$, $\kappa_{S}$, $\kappa_{M}$\foreignlanguage{english}{
respectively)} will be zero. If $\kappa=0$, then we have that \foreignlanguage{english}{$\frac{1}{\kappa}=\infty$
and therefore $\boldsymbol{z}_{1}=\boldsymbol{z}_{0}+\boldsymbol{v}$.}

\selectlanguage{english}%
Fig. \ref{fig:all_constraints} shows the results of RAYEN applied
to 12 different sets $\mathcal{Y}$. For each set, we generate 12K
samples $\boldsymbol{v}\in\mathbb{R}^{n}$ sampled uniformly in the
box $[-2.5,2.5]^{n}$ (see Fig. \ref{fig:pipeline}), compute the
corresponding $\kappa$, and plot the resulting point $\boldsymbol{y}=\boldsymbol{f}\left(\boldsymbol{z}_{1}\right)\in\mathbb{R}^{k}$
for each of those samples. All these points $\boldsymbol{y}$ are
guaranteed to lie in the set $\mathcal{Y}$. Note also that \foreignlanguage{american}{the
density of the produced points is higher in the frontiers of the set
due to the $\text{min}\left(\frac{1}{\kappa},\left\Vert \boldsymbol{v}\right\Vert \right)$
operation performed. }

\newcommand{\mysquare}[2]{\fcolorbox{#1}{#2}{\rule{0pt}{5pt}\rule{5pt}{0pt}}}

\definecolor{my_brown}{RGB}{217,194,183}
\definecolor{my_orange}{RGB}{250,225,191}
\definecolor{my_grey}{RGB}{213,208,216}
\definecolor{my_green}{RGB}{206,233,206}

\newcommand\mybox[2][]{%
  \tikz[baseline=(char.base)]\node[%
    fill=#1,%
    inner sep=2pt,%
    anchor=base,%
    rectangle,%
    rounded corners=0pt,%
    draw=black%
  ] (char) {#2};%
}

\newcommand\myboxredborder[2][]{%
  \tikz[baseline=(char.base)]\node[%
    fill=#1,%
    inner sep=2pt,%
    anchor=base,%
    rectangle,%
    rounded corners=0pt,%
    draw=red%
  ] (char) {#2};%
}

\newcommand\myboxblueborder[2][]{%
  \tikz[baseline=(char.base)]\node[%
    fill=#1,%
    inner sep=2pt,%
    anchor=base,%
    rectangle,%
    rounded corners=0pt,%
    draw=blue%
  ] (char) {#2};%
}

\begin{table*}
	\caption{\foreignlanguage{english}{Neural network architecture and details of each algorithm used in
			the comparisons of Section \ref{subsec:results_optimization}. Here,\foreignlanguage{american}{\noun{
					R} denotes a \noun{Relu} layer, L$(a,b)$ denotes a linear layer with
				input size $a$ and output size $b$, and BN denotes a batch normalization
				layer. For \textbf{Bar}, $q$ is the number of vertices
				plus the number of rays obtained by the double description method
				\cite{frerix2020homogeneous}.} \foreignlanguage{american}{\label{tab:all_algorithms}}}}
	
	\centering{}\includegraphics[width=1\textwidth]{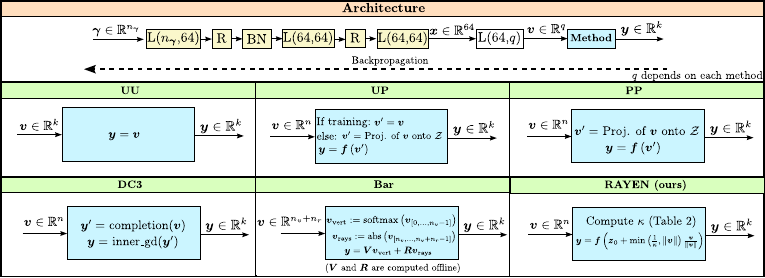}
\end{table*}

\section{Example}\label{sec:example}
\added{Fig.~\ref{fig:polyhedron3d_and_projection}  shows an example with $k=3$, $n=2$, and only linear and quadratic constraints. The steps performed by RAYEN in the offline and online phases are described as follows (see also Alg.~\ref{alg:rayen-algorithm}).
} 
\subsection{Offline}
\added{In the offline phase, RAYEN performs these steps:
\begin{itemize}
    \item The feasible set \mybox[my_orange]{$\mathcal{Y}$}$\;\subset \mathbb{R}^3$  is defined as the intersection between its \mybox[my_brown]{affine hull} (linear equality constraints, Section~\ref{subsec:affine-hull-of}), a \mybox[my_green]{polyhedron} (linear inequality), and an \mybox[my_grey]{ellipsoid $\mathcal{Y}_Q$} (quadratic inequality). The intersection between the \mybox[my_brown]{affine hull} and the \mybox[my_green]{polyhedron} is denoted as \myboxredborder[white]{$\mathcal{Y}_L$}\;$\subset \mathbb{R}^3$.
    \item Then, using the matrix $\boldsymbol{N}$ (Eq. \ref{eq:f_definition}), we compute
    \mybox[my_orange]{$\mathcal{Z}$}$\; \subset \mathbb{R}^2$ as the intersection between    \myboxredborder[white]{$\mathcal{Z}_L$}$\;\subset \mathbb{R}^2$ and \myboxblueborder[white]{$\mathcal{Z}_Q$}$\;\subset \mathbb{R}^2$
    \item The interior point $\boldsymbol{z}_0$ of \mybox[my_orange]{$\mathcal{Z}$} can then be found solving the optimization problem of Section~\ref{subsec:Interior-point-of}. 
\end{itemize}}

\subsection{Online (both training and testing)}
\renewcommand{\labelitemii}{$\circ$}
\added{In the online phase, RAYEN performs these steps:
\begin{itemize}    
    \item After passing the input through the network and the linear layer (Fig.~\ref{fig:pipeline}), we obtain $\boldsymbol{v}\in\mathbb{R}^2$. 
    \item Then, we compute 
    \begin{itemize}
        \item  $\kappa_{L}$, which is the inverse of \added{the} distance from $\boldsymbol{z}_0$ to the border of \myboxredborder[white]{$\mathcal{Z}_L$}, following the direction $\boldsymbol{v}$ (Section \ref{subsec:Computation-of-k_L}).
        \item  $\kappa_{Q}$, which is the inverse of \added{the} distance from $\boldsymbol{z}_0$ to the border of \myboxblueborder[white]{$\mathcal{Z}_Q$}, following the direction $\boldsymbol{v}$ (Section \ref{subsec:Computation-of-k_Q}).
        \item  $\kappa := \arraycolsep=1.2pt\text{max}\left[\begin{array}{cccc}
\kappa_{L} & \kappa_{Q} \end{array}\right]^{T}$.
    \end{itemize} 
     \item From $\boldsymbol{z}_0$, $\boldsymbol{v}$, and $\kappa$, we can compute $\boldsymbol{z}_1$ using Eq.~\ref{eq:z1_definition}, and therefore $\boldsymbol{y}:=\boldsymbol{f}\left(\boldsymbol{z}_1\right)$ (Eq.~\ref{eq:f_definition}) is then guaranteed to lie in \mybox[my_orange]{$\mathcal{Y}$}$\; \subset \mathbb{R}^3$
     \item During training, we use this output $\boldsymbol{y}$ to compute the loss and backpropagate. 
\end{itemize}
}

\section{Results}

\added{In this Section, we first study how RAYEN is able to approximate the optimal solution
of trajectory optimization problems with constraints (Section~\ref{subsec:results_optimization}).
Then, in Section~\ref{subsec:results_legged}, we use RAYEN to constrain the individual and total torques of the joints obtained by a learning-based locomotion policy for a quadruped robot.
Finally, in Section~\ref{subsec:Computation-time},} we
analyze RAYEN's computation time when applied to different constraints
of varying dimensions. 

\subsection{\added{Trajectory} optimization \label{subsec:results_optimization}}

One of the (many) applications of being able to impose constraints
on neural networks is when they are used to approximately solve optimization
problems much faster than standard solvers. In this section, we analyze
how a neural network equipped with RAYEN is able to approximate the
optimal solution of \added{a trajectory} optimization problem while satisfying all the
constraints. \added{Specifically, they are trajectory optimization problems that aim at} obtaining the clamped uniform B-Spline that minimizes a weighted sum of the velocity smoothness,
the acceleration smoothness, and the jerk smoothness, while ensuring
that the whole trajectory remains within a sequence of overlapping
polyhedra that represent the free space. This problem (or small variations
of it) appears extensively in many trajectory planning problems in
Robotics, such as \cite{zucker2013chomp,liu2017planning,tordesillas2020faster,tordesillas2020mader}.
Using the notation shown in Table \ref{tab:Notation-appendix},
the optimization problems are available in Table \ref{tab:opt1_opt2}.
The constraints are these: 

\begin{itemize}
	\item \textbf{Initial and final constraints:} The initial and final states
	are stop conditions. The initial position is fixed, while the final
	position of the trajectory is included in the cost as a soft constraint.
	This final position is taken inside $\mathcal{P}_{n_{\text{polyh}}-1}$
	(the last polyhedron of the corridor).
	\item \textbf{Corridor constraints:} Each interval of the trajectory is
	assigned to one polyhedron, and the convex hull property of the MINVO
	basis \cite{tordesillas2020minvo} is leveraged to ensure that each
	interval remains inside that assigned polyhedron. Compared to approaches
	that only impose constraints on discretization points along the trajectory,
	this approach guarantees safety for the whole trajectory $t\in[0,t_{f}]$.
	\item \textbf{Dynamic limit constraints:} Optimization 1 uses linear constraints
	to impose a maximum velocity $v_{\text{max}}$ and acceleration $a_{\text{max}}$.
	Optimization 2 uses quadratic constraints instead, and also imposes
	a maximum jerk $j_{\text{max}}$. Similar to the corridor constraints,
	we also leverage here the convex hull property of the MINVO basis \cite{tordesillas2020minvo} . 
\end{itemize}

\added{Hence, optimization problem 1 has only linear constraints, while optimization problem 2 has both linear and convex quadratic constraints. Both have a convex quadratic objective function, which depends}
on a parameter $\boldsymbol{\gamma}$. Nonconvex objective functions
could also be used, but we use convex objective functions to enable a direct 
comparison with the globally optimal solution obtained
using a state-of-the-art convex optimization solver, such as Gurobi \cite{gurobi}. 
\vspace{0.3cm}
\begin{table}[]
\renewcommand{\arraystretch}{1.2}
	\begin{centering}
		\caption{Notation used in \added{the formulation of the trajectory optimization problems (Section \ref{subsec:results_optimization})}. \label{tab:Notation-appendix}}
		\par\end{centering}
	\noindent\resizebox{\columnwidth}{!}{%
		\begin{centering}
			\begin{tabular}{|>{\centering}m{0.26\columnwidth}|>{\raggedright}m{0.74\columnwidth}|}
				\hline 
				\textbf{Symbol} & \textbf{\qquad \qquad \qquad \qquad Meaning}\tabularnewline
				\hline 
				\hline 
				$\ensuremath{\mathbf{p},\mathbf{v},\mathbf{a},\mathbf{j}}$ & Position, velocity, acceleration, and jerk\tabularnewline
				\hline 
				$\mathbf{\mathbf{p}}_{0},\mathbf{\mathbf{p}}_{f}$ & Initial and final positions\tabularnewline
				\hline 
				$n_{\text{interv}}$ & Number of intervals of the B-Spline\tabularnewline
				\hline 
				$n_{\text{cp}}$ & Number of position control points of the B-Spline\tabularnewline
				\hline 
				$\boldsymbol{a}_{l}$ & $l$-th acceleration control point of the B-Spline\tabularnewline
				\hline 
				$\boldsymbol{j}_{l}$ & $l$-th jerk control point of the B-Spline\tabularnewline
				\hline 
				$j,J$ & $j$ is the index of the interval of the trajectory, $j\in J:=\{0,1,...,n_{\text{interv}}-1\}$\tabularnewline
				\hline 
				$\ensuremath{\mathcal{Q}_{j}^{\text{MV}}},\mathcal{V}_{j}^{\text{MV}}$ & $\mathcal{Q}_{j}^{\text{MV}}$ is the set of position control points
				of the interval $j$ using the MINVO \added{basis~\cite{tordesillas2020minvo}}. Analogous definition for
				the velocity control points $\mathcal{V}_{j}^{\text{MV}}$\tabularnewline
				\hline 
				$\mathcal{P}_{0},...,\mathcal{P}_{n_{\text{polyh}}-1}$ & Sequence of overlapping polyhedra. They can be obtained using methods
				such as \cite{liu2017planning,deits2015computing,toumieh2022voxel,zhong2020generating,gao2020teach}
				\tabularnewline
				\hline 
				\added{$\alpha_{\text{v}}$, $\alpha_{\text{a}}$, $\alpha_{\text{j}}$} & \added{Weights used in the cost functions (see Table~\ref{tab:opt1_opt2})} 
				\tabularnewline
				\hline 
				\added{$\arraycolsep=2.1pt\boldsymbol{\gamma}$} & \added{$\arraycolsep=2.1pt\boldsymbol{\gamma}:=\left[\begin{array}{cccc}
				\alpha_{\text{v}} & \alpha_{\text{a}} & \alpha_{\text{j}} & \mathbf{\mathbf{p}}_{f}^{T}\end{array}\right]^{T}$}
				\tabularnewline
				\hline
				\vspace{0.05cm}
				$\left\lfloor a\right\rfloor $ & Floor function (i.e., rounding to the nearest integer $\le a$)\tabularnewline
				\hline 
			\end{tabular}
			\par\end{centering}
	}
\end{table}

\selectlanguage{english}%
\begin{table*}
\caption{\added{Trajectory optimization} problems \added{1 and 2 (in both original and standard form)} and losses used in Section \ref{subsec:results_optimization}.
$p_{\text{soft},L}$ and $p_{\text{soft},Q}$ are defined, respectively,
in Eq. \ref{eq:soft_L} and Eq. \ref{eq:soft_q}. The decision variables are the control
points of the spline $\mathbf{p}(t)$.
In these optimizations, the constraint $\mathcal{Q}_{j}^{\text{MV}}\subseteq\mathcal{P}_{\left\lfloor \frac{j}{2}\right\rfloor },\;\forall j\in J$
is a linear constraint, because the MINVO control \added{points~\cite{tordesillas2020minvo}} are linear
functions of the spline control points. \label{tab:opt1_opt2} \label{tab:definitions_problems}}

\selectlanguage{american}%
\resizebox{\textwidth}{!}{
\selectlanguage{english}%
\begin{centering}
\begin{tabular}{|c|c|c||c|c|}
\cline{2-5} 
\multicolumn{1}{c|}{\selectlanguage{american}%
\selectlanguage{english}%
} & \multicolumn{2}{c||}{\selectlanguage{american}%
\textbf{\added{Optimization} 1} \selectlanguage{english}%
} & \multicolumn{2}{c|}{\selectlanguage{american}%
\textbf{\added{Optimization} 2} \selectlanguage{english}%
}\tabularnewline
\cline{2-5} 
\multicolumn{1}{c|}{\selectlanguage{american}%
\selectlanguage{english}%
} & \multicolumn{2}{>{\centering}p{0.4\textwidth}||}{\selectlanguage{american}%
\added{\hspace{1.2cm}\textbf{\uline{Original form:}}} \hspace{0cm}
\newline
\newline
				$\begin{aligned}\underset{\mathbf{p}(t)\in\mathcal{S}}{\text{min}}\quad & \alpha_{\text{v}}\int_{0}^{t_{f}}\left\Vert \mathbf{v}(t)\right\Vert ^{2}dt+\alpha_{\text{a}}\int_{0}^{t_{f}}\left\Vert \mathbf{a}(t)\right\Vert ^{2}dt\\
& +\alpha_{\text{j}}\int_{0}^{t_{f}}\left\Vert \mathbf{j}(t)\right\Vert ^{2}dt+\left\Vert \mathbf{p}(t_{f})-\mathbf{\mathbf{p}}_{f}\right\Vert ^{2}\\
\text{s.t.:\quad} & \mathbf{p}(0)=\mathbf{\mathbf{p}}_{0}\\
& \mathbf{v}(0)=\mathbf{v}(t_{f})=\boldsymbol{0}\:\text{m/s}\\
& \mathcal{Q}_{j}^{\text{MV}}\subseteq\mathcal{P}_{\left\lfloor \frac{j}{2}\right\rfloor },\;\forall j\in J\\
& \text{abs}\left(\boldsymbol{v}\right)\le v_{\text{max}}\boldsymbol{1}\hspace{0.5em}\;\forall\boldsymbol{v}\in\mathcal{V}_{j}^{\text{MV}},\;\forall j\in J\\
& \text{abs}\left(\boldsymbol{a}_{l}\right)\le a_{\text{max}}\boldsymbol{1}\hspace{0.5em}\forall l\in\{0,1,...,n_{\text{cp}}-3\}\\
\\
\\
\\
\end{aligned}
$

\vspace{0.01cm}

\begin{minipage}[t]{0.38\textwidth}%
	Here, $\mathcal{S}$ is the set of clamped uniform B-Splines with
	dimension $2$, degree $2$, and 11 knots (i.e., $n_{\text{cp}}=8$
	and $n_{\text{interv}}=6$). Moreover, $n_{\text{polyh}}=3$,  $t_{f}=35$
	s\added{, $v_{\text{max}}=4\,\text{m/s}$,
		$a_{\text{max}}=8\;$m/s\protect\textsuperscript{2}, and $j_{\text{max}}=50\;$m/s\protect\textsuperscript{3}}. %
\end{minipage}
\selectlanguage{english}%
} & \multicolumn{2}{>{\centering}p{0.5\textwidth}|}{\selectlanguage{american}%
\added{\textbf{\uline{Original form:}}} \hspace{6.5cm}
\newline
\newline
$\begin{aligned}\underset{\mathbf{p}(t)\in\mathcal{S}}{\text{min}}\quad & \alpha_{\text{v}}\int_{0}^{t_{f}}\left\Vert \mathbf{v}(t)\right\Vert ^{2}dt+\alpha_{\text{a}}\int_{0}^{t_{f}}\left\Vert \mathbf{a}(t)\right\Vert ^{2}dt\\
& +\alpha_{\text{j}}\int_{0}^{t_{f}}\left\Vert \mathbf{j}(t)\right\Vert ^{2}dt+\left\Vert \mathbf{p}(t_{f})-\mathbf{\mathbf{p}}_{f}\right\Vert ^{2}\\
\text{s.t.:}\quad & \mathbf{p}(0)=\mathbf{\mathbf{p}}_{0}\\
& \mathbf{v}(0)=\mathbf{v}(t_{f})=\boldsymbol{0}\:\text{m/s}\\
& \mathbf{a}(0)=\mathbf{a}(t_{f})=\boldsymbol{0}\:\text{m/s}^{2}\\
& \mathcal{Q}_{j}^{\text{MV}}\subseteq\mathcal{P}_{\left\lfloor \frac{j}{2}\right\rfloor },\;\forall j\in J\\
& \left\Vert \boldsymbol{v}\right\Vert ^{2}\ensuremath{\le}v_{\text{max}}^{2}\quad\;\forall\boldsymbol{v}\in\mathcal{V}_{j}^{\text{MV}},\;\forall j\in J\\
& \left\Vert \boldsymbol{a}_{l}\right\Vert ^{2}\le a_{\text{max}}^{2}\hspace{0.5em}\forall l\in\{0,1,...,n_{\text{cp}}-3\}\\
& \left\Vert \boldsymbol{j}_{l}\right\Vert ^{2}\le j_{\text{max}}^{2}\hspace{0.5em}\;\forall l\in\{0,1,...,n_{\text{cp}}-4\}
\end{aligned}
$

\vspace{0.3cm}

\begin{minipage}[t]{0.49\textwidth}%
	Here, $\mathcal{S}$ is the set of clamped uniform B-Splines with
	dimension $3$, degree $3$, and 19 knots (i.e., $n_{\text{cp}}=15$
	and $n_{\text{interv}}=12$). Moreover, $n_{\text{polyh}}=6$, 
	$t_{f}=15$ s, \added{$v_{\text{max}}=4\,\text{m/s}$,
		$a_{\text{max}}=8\;$m/s\protect\textsuperscript{2}, and $j_{\text{max}}=50\;$m/s\protect\textsuperscript{3}}. %
\end{minipage}
\selectlanguage{english}%
}\tabularnewline
\cline{2-5} 
\multicolumn{1}{c|}{\selectlanguage{american}%
	\selectlanguage{english}%
} & \multicolumn{2}{>{\centering}p{0.4\textwidth}||}{\selectlanguage{american}%
   \added{\textbf{\uline{Standard form:}}} \hspace{4.9cm}
   \newline
   \newline
   {\small
	$\begin{aligned}\underset{\boldsymbol{y}}{\text{min}} & \quad p_{1}(\boldsymbol{y},\boldsymbol{\gamma})\\
	\text{s.t.:} & \quad\boldsymbol{A}_{1}\boldsymbol{y}\le\boldsymbol{b}_{1}\\
	\text{} & \quad\boldsymbol{A}_{2}\boldsymbol{y}=\boldsymbol{b}_{2}
	\end{aligned}
	$, where $\left\{ \begin{aligned} & \boldsymbol{\gamma}\in\mathbb{R}^{5},\boldsymbol{y}\in\mathbb{R}^{16}\\
	\text{} & \boldsymbol{A}_{1}\in\mathbb{R}^{168\times16}\\
    \text{} & \boldsymbol{b}_{1}\in\mathbb{R}^{168}\\
	\text{} & \boldsymbol{A}_{2}\in\mathbb{R}^{6\times16}\\
    \text{} & \boldsymbol{b}_{2}\in\mathbb{R}^{6}
	\end{aligned}
	\right.$
    }
	\vspace{0.55cm}
	\newline
		\begin{minipage}[t]{0.39\textwidth}%
		\vspace{0.1cm}
	\added{This problem has a quadratic cost with linear constraints. $p_{1}(\boldsymbol{y},\boldsymbol{\gamma})$ is the cost function (see top row of this table), $\arraycolsep=2.1pt\boldsymbol{\gamma}:=\left[\begin{array}{cccc}
		\alpha_{\text{v}} & \alpha_{\text{a}} & \alpha_{\text{j}} & \mathbf{\mathbf{p}}_{f}^{T}\end{array}\right]^{T}$ is a parameter, and $\boldsymbol{y}$ is the decision variable that contains the control points of the spline.}
	\vspace{0.15cm}
\end{minipage}
	\selectlanguage{english}%
} & \multicolumn{2}{>{\centering}p{0.5\textwidth}|}{\selectlanguage{american}%
   \added{\textbf{\uline{Standard form:}}} \hspace{6.5cm}
\newline
\newline
{\small
	$\begin{aligned}\underset{\boldsymbol{y}}{\text{min}} & \quad p_{2}(\boldsymbol{y},\boldsymbol{\gamma})\\
	\text{s.t.:} & \quad\boldsymbol{A}_{1}\boldsymbol{y}\le\boldsymbol{b}_{1}\\
	\text{} & \quad\boldsymbol{A}_{2}\boldsymbol{y}=\boldsymbol{b}_{2}\\
	& \quad g_{i}\left(\boldsymbol{y}\right)\le0\quad i=0,...,\eta-1
	\end{aligned}
	$, where $\left\{ \begin{aligned}\text{} & \boldsymbol{\gamma}\in\mathbb{R}^{6},\boldsymbol{y}\in\mathbb{R}^{45}\\
	\text{} & \boldsymbol{A}_{1}\in\mathbb{R}^{1072\times45}\\
    \text{} & \boldsymbol{b}_{1}\in\mathbb{R}^{1072} \\
	\text{} & \boldsymbol{A}_{2}\in\mathbb{R}^{15\times45}\\
    \text{} & \boldsymbol{b}_{2}\in\mathbb{R}^{15}\\
	\text{} & \eta=72
	\end{aligned}
	\right.$
    }
	\newline
	\begin{minipage}[t]{0.49\textwidth}%
		\vspace{0.22cm}
	\added{This problem has a quadratic cost with linear and quadratic constraints. $p_{2}(\boldsymbol{y},\boldsymbol{\gamma})$ is the cost function (see top row of this table), $\arraycolsep=2.1pt\boldsymbol{\gamma}:=\left[\begin{array}{cccc}
		\alpha_{\text{v}} & \alpha_{\text{a}} & \alpha_{\text{j}} & \mathbf{\mathbf{p}}_{f}^{T}\end{array}\right]^{T}$ is a parameter, and $\boldsymbol{y}$ is the decision variable that contains the control points of the spline.}
	\vspace{0.15cm}
\end{minipage}
	\selectlanguage{english}%
}\tabularnewline
\hline 
\textbf{Method} & \multirow{1}{*}{\selectlanguage{american}%
\textbf{Training loss}\selectlanguage{english}%
} & \multirow{1}{*}{\selectlanguage{american}%
\textbf{Test Loss}\selectlanguage{english}%
} & \multirow{1}{*}{\selectlanguage{american}%
\textbf{Training loss}\selectlanguage{english}%
} & \multirow{1}{*}{\selectlanguage{american}%
\textbf{Test Loss}\selectlanguage{english}%
}\tabularnewline
\hline 
\selectlanguage{american}%
\textbf{Bar}\selectlanguage{english}%
 & \selectlanguage{american}%
$p_{1}(\boldsymbol{y},\boldsymbol{\gamma})$\selectlanguage{english}%
 & \multirow{7}{*}{\selectlanguage{american}%
$p_{1}(\boldsymbol{y},\boldsymbol{\gamma})$\selectlanguage{english}%
} & \multicolumn{2}{c|}{\selectlanguage{american}%
Does not support quadratic constraints \selectlanguage{english}%
}\tabularnewline
\cline{1-2} \cline{4-5} 
\textbf{UU} & \multirow{3}{*}{\selectlanguage{american}%
$p_{1}(\boldsymbol{y},\boldsymbol{\gamma})+\omega p_{\text{soft},L}$\selectlanguage{english}%
} &  & \multirow{3}{*}{\selectlanguage{american}%
$p_{2}(\boldsymbol{y},\boldsymbol{\gamma})+\omega\left(p_{\text{soft},L}+p_{\text{soft},Q}\right)$\selectlanguage{english}%
} & \multirow{6}{*}{\selectlanguage{american}%
$p_{2}(\boldsymbol{y},\boldsymbol{\gamma})$\selectlanguage{english}%
}\tabularnewline
\cline{1-1} 
\textbf{UP} &  &  &  & \tabularnewline
\cline{1-1} 
\selectlanguage{american}%
\textbf{DC3}\selectlanguage{english}%
 &  &  &  & \tabularnewline
\cline{1-2} \cline{4-4} 
\selectlanguage{american}%
\textbf{PP}\selectlanguage{english}%
 & \multirow{2}{*}{\selectlanguage{american}%
$p_{1}(\boldsymbol{y},\boldsymbol{\gamma})$\selectlanguage{english}%
} &  & \multirow{2}{*}{\selectlanguage{american}%
$p_{2}(\boldsymbol{y},\boldsymbol{\gamma})$\selectlanguage{english}%
} & \tabularnewline
\cline{1-1} 
\selectlanguage{american}%
\textbf{RAYEN}\selectlanguage{english}%
 &  &  &  & \tabularnewline
\cline{1-2} \cline{4-4} 
\selectlanguage{american}%
\textbf{Gurobi}\selectlanguage{english}%
 & \selectlanguage{american}%
-\selectlanguage{english}%
 &  & \selectlanguage{american}%
-\selectlanguage{english}%
 & \tabularnewline
\hline 
\end{tabular}
\par\end{centering}
}
\end{table*}

\definecolor{RedLight}{RGB}{255,204,204}
\definecolor{GreenLight}{RGB}{204,255,204}
\newcommand{\vi}{\cellcolor{RedLight}}
\newcommand{\fe}{\cellcolor{GreenLight}}

\begin{figure*}
	\centering
	
	\captionof{table}{Results for the optimization problems. For all the learning-based
		methods, the inference time is per sample in the batch (using a batch
		size of 512 samples). For Gurobi, the average time to solve the optimization
		for each of the 512 samples is shown. In Optimization 2, \foreignlanguage{american}{\textbf{DC3}} with \textbf{$\omega=1000$
		}and\foreignlanguage{american}{ \textbf{DC3}} with \textbf{$\omega=5000$
		}diverged for, respectively,\textbf{ $5.3\%$ }and\textbf{ $7.8\%$
		}of the samples, generating NaNs. The losses and violations shown
		are for the converged samples. In this table, we highlight in bold
		the best value of each column when compared to the learning-based
		methods that generate feasible solutions (i.e., zero violation of
		the constraints). \label{tab:results_optimization}}
\resizebox{\textwidth}{!}{
\begin{centering}
	\renewcommand{\arraystretch}{0.9}
	\begin{tabular}{|c|c|c|c|c|c|c|c|}
		\hline 
		\multirow{3}{*}{\textbf{Method}} & \multicolumn{7}{c|}{\selectlanguage{american}%
			\textbf{\added{Optimization} 1 (Linear Constraints)}\selectlanguage{english}%
		}\tabularnewline
		\cline{2-8} 
		& \multirow{2}{*}{\selectlanguage{american}%
			$\begin{array}{c}
			\text{\textbf{Num. of}}\\
			\text{\textbf{params}}
			\end{array}$\selectlanguage{english}%
		} & \multicolumn{3}{c|}{\selectlanguage{american}%
			\textbf{Same distribution as training:} $\boldsymbol{\gamma}\in\text{TE}_{1,\text{in}}$\selectlanguage{english}%
		} & \multicolumn{3}{c|}{\selectlanguage{american}%
			\textbf{Different distribution as training:} $\boldsymbol{\gamma}\in\text{TE}_{1,\text{out}}$\selectlanguage{english}%
		}\tabularnewline
		\cline{3-8} 
		&  & \selectlanguage{american}%
		\textbf{Normalized Loss}\selectlanguage{english}%
		& \selectlanguage{american}%
		\textbf{Violation}\selectlanguage{english}%
		& \selectlanguage{american}%
		\textbf{Inference Time (}$\ensuremath{\mu}s$\textbf{)}\selectlanguage{english}%
		& \selectlanguage{american}%
		\textbf{Normalized Loss}\selectlanguage{english}%
		& \selectlanguage{american}%
		\textbf{Violation}\selectlanguage{english}%
		& \selectlanguage{american}%
		\textbf{Inference Time (}$\ensuremath{\mu}s$\textbf{)}\selectlanguage{english}%
		\tabularnewline
		\hline 
		\textbf{UU}, \textbf{$\omega=0$} & \multirow{5}{*}{\selectlanguage{american}%
			9872\selectlanguage{english}%
		} & \selectlanguage{american}%
		0.0006\selectlanguage{english}%
		& \selectlanguage{american}%
		\vi{}1943.58\selectlanguage{english}%
		& \selectlanguage{american}%
		0.34\selectlanguage{english}%
		& \selectlanguage{american}%
		0.0013\selectlanguage{english}%
		& \selectlanguage{american}%
		\vi{}1937.43\selectlanguage{english}%
		& \selectlanguage{american}%
		0.43\selectlanguage{english}%
		\tabularnewline
		\textbf{UU}, \textbf{$\omega=10$} &  & \selectlanguage{american}%
		2.4431\selectlanguage{english}%
		& \selectlanguage{american}%
		\vi{}2.68\selectlanguage{english}%
		& \selectlanguage{american}%
		0.36\selectlanguage{english}%
		& \selectlanguage{american}%
		1.8362\selectlanguage{english}%
		& \selectlanguage{american}%
		\vi{}14.02\selectlanguage{english}%
		& \selectlanguage{american}%
		0.43\selectlanguage{english}%
		\tabularnewline
		\textbf{UU}, \textbf{$\omega=100$} &  & \selectlanguage{american}%
		1.0278\selectlanguage{english}%
		& \selectlanguage{american}%
		\vi{}0.00418\selectlanguage{english}%
		& \selectlanguage{american}%
		0.36\selectlanguage{english}%
		& \selectlanguage{american}%
		1.0220\selectlanguage{english}%
		& \selectlanguage{american}%
		\vi{}0.08535\selectlanguage{english}%
		& \selectlanguage{american}%
		0.43\selectlanguage{english}%
		\tabularnewline
		\textbf{UU}, \textbf{$\omega=1000$} &  & \selectlanguage{american}%
		1.4391\selectlanguage{english}%
		& \selectlanguage{american}%
		\vi{}0.00255\selectlanguage{english}%
		& \selectlanguage{american}%
		0.36\selectlanguage{english}%
		& \selectlanguage{american}%
		1.2263\selectlanguage{english}%
		& \selectlanguage{american}%
		\vi{}0.19527\selectlanguage{english}%
		& \selectlanguage{american}%
		0.44\selectlanguage{english}%
		\tabularnewline
		\textbf{UU}, \textbf{$\omega=5000$} &  & \selectlanguage{american}%
		2.1784\selectlanguage{english}%
		& \selectlanguage{american}%
		\vi{}0.00325\selectlanguage{english}%
		& \selectlanguage{american}%
		0.36\selectlanguage{english}%
		& \selectlanguage{american}%
		1.4767\selectlanguage{english}%
		& \selectlanguage{american}%
		\vi{}0.02354\selectlanguage{english}%
		& \selectlanguage{american}%
		0.43\selectlanguage{english}%
		\tabularnewline
		\hline 
		\textbf{UP}, \textbf{$\omega=0$} & \multirow{5}{*}{\selectlanguage{american}%
			\textbf{9482}\selectlanguage{english}%
		} & \selectlanguage{american}%
		1.0533\selectlanguage{english}%
		& \selectlanguage{american}%
		\fe{}0.00000\selectlanguage{english}%
		& \selectlanguage{american}%
		4979.26\selectlanguage{english}%
		& \selectlanguage{american}%
		1.0700\selectlanguage{english}%
		& \selectlanguage{american}%
		\fe{}0.00000\selectlanguage{english}%
		& \selectlanguage{american}%
		4910.93\selectlanguage{english}%
		\tabularnewline
		\textbf{UP}, \textbf{$\omega=10$} &  & \selectlanguage{american}%
		1.0176\selectlanguage{english}%
		& \selectlanguage{american}%
		\fe{}0.00000\selectlanguage{english}%
		& \selectlanguage{american}%
		5153.89\selectlanguage{english}%
		& \selectlanguage{american}%
		1.0384\selectlanguage{english}%
		& \selectlanguage{american}%
		\fe{}0.00000\selectlanguage{english}%
		& \selectlanguage{american}%
		5012.32\selectlanguage{english}%
		\tabularnewline
		\textbf{UP}, \textbf{$\omega=100$} &  & \selectlanguage{american}%
		1.0319\selectlanguage{english}%
		& \selectlanguage{american}%
		\fe{}0.00000\selectlanguage{english}%
		& \selectlanguage{american}%
		4866.78\selectlanguage{english}%
		& \selectlanguage{american}%
		1.4119\selectlanguage{english}%
		& \selectlanguage{american}%
		\fe{}0.00000\selectlanguage{english}%
		& \selectlanguage{american}%
		5013.74\selectlanguage{english}%
		\tabularnewline
		\textbf{UP}, \textbf{$\omega=1000$} &  & \selectlanguage{american}%
		1.0686\selectlanguage{english}%
		& \selectlanguage{american}%
		\fe{}0.00000\selectlanguage{english}%
		& \selectlanguage{american}%
		4673.84\selectlanguage{english}%
		& \selectlanguage{american}%
		1.0882\selectlanguage{english}%
		& \selectlanguage{american}%
		\fe{}0.00000\selectlanguage{english}%
		& \selectlanguage{american}%
		4875.74\selectlanguage{english}%
		\tabularnewline
		\textbf{UP}, \textbf{$\omega=5000$} &  & \selectlanguage{american}%
		1.1456\selectlanguage{english}%
		& \selectlanguage{american}%
		\fe{}0.00000\selectlanguage{english}%
		& \selectlanguage{american}%
		4742.92\selectlanguage{english}%
		& \selectlanguage{american}%
		1.2436\selectlanguage{english}%
		& \selectlanguage{american}%
		\fe{}0.00000\selectlanguage{english}%
		& \selectlanguage{american}%
		4754.38\selectlanguage{english}%
		\tabularnewline
		\hline 
		\selectlanguage{american}%
		\textbf{PP}\selectlanguage{english}%
		& \selectlanguage{american}%
		\textbf{9482}\selectlanguage{english}%
		& \selectlanguage{american}%
		11.8104\selectlanguage{english}%
		& \selectlanguage{american}%
		\fe{}0.00000\selectlanguage{english}%
		& \selectlanguage{american}%
		5054.10\selectlanguage{english}%
		& \selectlanguage{american}%
		4.5885\selectlanguage{english}%
		& \selectlanguage{american}%
		\fe{}0.00000\selectlanguage{english}%
		& \selectlanguage{american}%
		5249.53\selectlanguage{english}%
		\tabularnewline
		\hline 
		\selectlanguage{american}%
		\textbf{DC3}\foreignlanguage{english}{, \textbf{$\omega=0$}}\selectlanguage{english}%
		& \multirow{5}{*}{\selectlanguage{american}%
			\textbf{9482}\selectlanguage{english}%
		} & \selectlanguage{american}%
		0.6789\selectlanguage{english}%
		& \selectlanguage{american}%
		\vi{}85.11\selectlanguage{english}%
		& \selectlanguage{american}%
		143.85\selectlanguage{english}%
		& \selectlanguage{american}%
		61.1547\selectlanguage{english}%
		& \selectlanguage{american}%
		\vi{}9394.15\selectlanguage{english}%
		& \selectlanguage{american}%
		146.01\selectlanguage{english}%
		\tabularnewline
		\selectlanguage{american}%
		\textbf{DC3}\foreignlanguage{english}{, \textbf{$\omega=10$}}\selectlanguage{english}%
		&  & \selectlanguage{american}%
		1.0000\selectlanguage{english}%
		& \selectlanguage{american}%
		\vi{}0.01875\selectlanguage{english}%
		& \selectlanguage{american}%
		142.21\selectlanguage{english}%
		& \selectlanguage{american}%
		0.9968\selectlanguage{english}%
		& \selectlanguage{american}%
		\vi{}0.87179\selectlanguage{english}%
		& \selectlanguage{american}%
		144.96\selectlanguage{english}%
		\tabularnewline
		\selectlanguage{american}%
		\textbf{DC3}\foreignlanguage{english}{, \textbf{$\omega=100$}}\selectlanguage{english}%
		&  & \selectlanguage{american}%
		1.0304\selectlanguage{english}%
		& \selectlanguage{american}%
		\vi{}0.00089\selectlanguage{english}%
		& \selectlanguage{american}%
		145.11\selectlanguage{english}%
		& \selectlanguage{american}%
		1.1979\selectlanguage{english}%
		& \selectlanguage{american}%
		\vi{}21.01\selectlanguage{english}%
		& \selectlanguage{american}%
		145.92\selectlanguage{english}%
		\tabularnewline
		\selectlanguage{american}%
		\textbf{DC3}\foreignlanguage{english}{, \textbf{$\omega=1000$}}\selectlanguage{english}%
		&  & \selectlanguage{american}%
		1.1068\selectlanguage{english}%
		& \selectlanguage{american}%
		\vi{}0.00018\selectlanguage{english}%
		& \selectlanguage{american}%
		145.50\selectlanguage{english}%
		& \selectlanguage{american}%
		1.0780\selectlanguage{english}%
		& \selectlanguage{american}%
		\vi{}0.32573\selectlanguage{english}%
		& \selectlanguage{american}%
		147.21\selectlanguage{english}%
		\tabularnewline
		\selectlanguage{american}%
		\textbf{DC3}\foreignlanguage{english}{, \textbf{$\omega=5000$}}\selectlanguage{english}%
		&  & \selectlanguage{american}%
		1.1413\selectlanguage{english}%
		& \selectlanguage{american}%
		\fe{}0.00000\selectlanguage{english}%
		& \selectlanguage{american}%
		145.22\selectlanguage{english}%
		& \selectlanguage{american}%
		1.1237\selectlanguage{english}%
		& \selectlanguage{american}%
		\vi{}0.01873\selectlanguage{english}%
		& \selectlanguage{american}%
		147.28\selectlanguage{english}%
		\tabularnewline
		\hline 
		\selectlanguage{american}%
		\textbf{Bar}\selectlanguage{english}%
		& \selectlanguage{american}%
		1869847\selectlanguage{english}%
		& \selectlanguage{american}%
		1.0143\selectlanguage{english}%
		& \selectlanguage{american}%
		\fe{}0.00000\selectlanguage{english}%
		& \selectlanguage{american}%
		13.96\selectlanguage{english}%
		& \selectlanguage{american}%
		\textbf{1.0070}\selectlanguage{english}%
		& \selectlanguage{american}%
		\fe{}0.00000\selectlanguage{english}%
		& \selectlanguage{american}%
		12.95\selectlanguage{english}%
		\tabularnewline
		\hline 
		\selectlanguage{american}%
		\textbf{RAYEN}\selectlanguage{english}%
		& \selectlanguage{american}%
		\textbf{9482}\selectlanguage{english}%
		& \selectlanguage{american}%
		\textbf{1.0075}\selectlanguage{english}%
		& \selectlanguage{american}%
		\fe{}0.00000\selectlanguage{english}%
		& \selectlanguage{american}%
		\textbf{0.69}\selectlanguage{english}%
		& \selectlanguage{american}%
		1.0076\selectlanguage{english}%
		& \selectlanguage{american}%
		\fe{}0.00000\selectlanguage{english}%
		& \selectlanguage{american}%
		\textbf{0.83}\selectlanguage{english}%
		\tabularnewline
		\hline 
		\hline 
		\selectlanguage{american}%
		\textbf{Gurobi}\selectlanguage{english}%
		& \selectlanguage{american}%
		-\selectlanguage{english}%
		& \selectlanguage{american}%
		1.0000\selectlanguage{english}%
		& \selectlanguage{american}%
		\fe{}0.00000\selectlanguage{english}%
		& \selectlanguage{american}%
		1515.48\selectlanguage{english}%
		& \selectlanguage{american}%
		1.0000\selectlanguage{english}%
		& \selectlanguage{american}%
		\fe{}0.00000\selectlanguage{english}%
		& \selectlanguage{american}%
		1405.60\selectlanguage{english}%
		\tabularnewline
		\hline 
	\end{tabular}
	\par\end{centering}
}
\vspace{0.3cm}
\newline
\renewcommand{\arraystretch}{0.9}
\resizebox{\textwidth}{!}{
\begin{centering}
	\begin{tabular}{|c|c|c|c|c|c|c|c|}
		\hline 
		\multirow{3}{*}{\textbf{Method}} & \multicolumn{7}{c|}{\selectlanguage{american}%
			\textbf{\added{Optimization} 2 (Linear and Convex Quadratic Constraints)}\selectlanguage{english}%
		}\tabularnewline
		\cline{2-8} 
		& \multirow{2}{*}{\selectlanguage{american}%
			$\begin{array}{c}
			\text{\textbf{Num. of}}\\
			\text{\textbf{params}}
			\end{array}$\selectlanguage{english}%
		} & \multicolumn{3}{c|}{\selectlanguage{american}%
			\textbf{Same distribution as training:} $\boldsymbol{\gamma}\in\text{TE}_{2,\text{in}}$\selectlanguage{english}%
		} & \multicolumn{3}{c|}{\selectlanguage{american}%
			\textbf{Different distribution as training:} $\boldsymbol{\gamma}\in\text{TE}_{2,\text{out}}$\selectlanguage{english}%
		}\tabularnewline
		\cline{3-8} 
		&  & \selectlanguage{american}%
		\textbf{Normalized Loss}\selectlanguage{english}%
		& \selectlanguage{american}%
		\textbf{Violation}\selectlanguage{english}%
		& \selectlanguage{american}%
		\textbf{Inference Time (}$\ensuremath{\mu}s$\textbf{)}\selectlanguage{english}%
		& \selectlanguage{american}%
		\textbf{Normalized Loss}\selectlanguage{english}%
		& \selectlanguage{american}%
		\textbf{Violation}\selectlanguage{english}%
		& \selectlanguage{american}%
		\textbf{Inference Time (}$\ensuremath{\mu}s$\textbf{)}\selectlanguage{english}%
		\tabularnewline
		\hline 
		\textbf{UU}, \textbf{$\omega=0$} & \multirow{5}{*}{\selectlanguage{american}%
			11821\selectlanguage{english}%
		} & \selectlanguage{american}%
		0.0039\selectlanguage{english}%
		& \selectlanguage{american}%
		\vi{}830.86\selectlanguage{english}%
		& \selectlanguage{american}%
		0.40\selectlanguage{english}%
		& \selectlanguage{american}%
		75.0200\selectlanguage{english}%
		& \selectlanguage{american}%
		\vi{}94371.24\selectlanguage{english}%
		& \selectlanguage{american}%
		0.44\selectlanguage{english}%
		\tabularnewline
		\textbf{UU}, \textbf{$\omega=10$} &  & \selectlanguage{american}%
		0.7876\selectlanguage{english}%
		& \selectlanguage{american}%
		\vi{}2.07\selectlanguage{english}%
		& \selectlanguage{american}%
		0.40\selectlanguage{english}%
		& \selectlanguage{american}%
		2.4223\selectlanguage{english}%
		& \selectlanguage{american}%
		\vi{}879.23\selectlanguage{english}%
		& \selectlanguage{american}%
		0.43\selectlanguage{english}%
		\tabularnewline
		\textbf{UU}, \textbf{$\omega=100$} &  & \selectlanguage{american}%
		1.1846\selectlanguage{english}%
		& \selectlanguage{american}%
		\vi{}0.02793\selectlanguage{english}%
		& \selectlanguage{american}%
		0.38\selectlanguage{english}%
		& \selectlanguage{american}%
		15.3420\selectlanguage{english}%
		& \selectlanguage{american}%
		\vi{}6256.17\selectlanguage{english}%
		& \selectlanguage{american}%
		0.44\selectlanguage{english}%
		\tabularnewline
		\textbf{UU}, \textbf{$\omega=1000$} &  & \selectlanguage{american}%
		1.6887\selectlanguage{english}%
		& \selectlanguage{american}%
		\vi{}0.01280\selectlanguage{english}%
		& \selectlanguage{american}%
		0.38\selectlanguage{english}%
		& \selectlanguage{american}%
		1.6474\selectlanguage{english}%
		& \selectlanguage{american}%
		\vi{}32.06\selectlanguage{english}%
		& \selectlanguage{american}%
		0.43\selectlanguage{english}%
		\tabularnewline
		\textbf{UU}, \textbf{$\omega=5000$} &  & \selectlanguage{american}%
		2.7045\selectlanguage{english}%
		& \selectlanguage{american}%
		\vi{}0.00656\selectlanguage{english}%
		& \selectlanguage{american}%
		0.40\selectlanguage{english}%
		& \selectlanguage{american}%
		2.8392\selectlanguage{english}%
		& \selectlanguage{american}%
		\vi{}0.01392\selectlanguage{english}%
		& \selectlanguage{american}%
		0.43\selectlanguage{english}%
		\tabularnewline
		\hline 
		\textbf{UP}, \textbf{$\omega=0$} & \multirow{5}{*}{\selectlanguage{american}%
			\textbf{10846}\selectlanguage{english}%
		} & \selectlanguage{american}%
		7.9809\selectlanguage{english}%
		& \selectlanguage{american}%
		\fe{}0.00000\selectlanguage{english}%
		& \selectlanguage{american}%
		10455.95\selectlanguage{english}%
		& \selectlanguage{american}%
		10.3325\selectlanguage{english}%
		& \selectlanguage{american}%
		\fe{}0.00000\selectlanguage{english}%
		& \selectlanguage{american}%
		10749.64\selectlanguage{english}%
		\tabularnewline
		\textbf{UP}, \textbf{$\omega=10$} &  & \selectlanguage{american}%
		1.1096\selectlanguage{english}%
		& \selectlanguage{american}%
		\fe{}0.00000\selectlanguage{english}%
		& \selectlanguage{american}%
		10595.68\selectlanguage{english}%
		& \selectlanguage{american}%
		1.3900\selectlanguage{english}%
		& \selectlanguage{american}%
		\fe{}0.00000\selectlanguage{english}%
		& \selectlanguage{american}%
		9572.01\selectlanguage{english}%
		\tabularnewline
		\textbf{UP}, \textbf{$\omega=100$} &  & \selectlanguage{american}%
		1.0386\selectlanguage{english}%
		& \selectlanguage{american}%
		\fe{}0.00000\selectlanguage{english}%
		& \selectlanguage{american}%
		11750.17\selectlanguage{english}%
		& \selectlanguage{american}%
		1.6074\selectlanguage{english}%
		& \selectlanguage{american}%
		\fe{}0.00000\selectlanguage{english}%
		& \selectlanguage{american}%
		8974.19\selectlanguage{english}%
		\tabularnewline
		\textbf{UP}, \textbf{$\omega=1000$} &  & \selectlanguage{american}%
		1.1305\selectlanguage{english}%
		& \selectlanguage{american}%
		\fe{}0.00000\selectlanguage{english}%
		& \selectlanguage{american}%
		10999.27\selectlanguage{english}%
		& \selectlanguage{american}%
		1.3692\selectlanguage{english}%
		& \selectlanguage{american}%
		\fe{}0.00000\selectlanguage{english}%
		& \selectlanguage{american}%
		10579.41\selectlanguage{english}%
		\tabularnewline
		\textbf{UP}, \textbf{$\omega=5000$} &  & \selectlanguage{american}%
		1.1455\selectlanguage{english}%
		& \selectlanguage{american}%
		\fe{}0.00000\selectlanguage{english}%
		& \selectlanguage{american}%
		10554.54\selectlanguage{english}%
		& \selectlanguage{american}%
		1.1972\selectlanguage{english}%
		& \selectlanguage{american}%
		\fe{}0.00000\selectlanguage{english}%
		& \selectlanguage{american}%
		9719.98\selectlanguage{english}%
		\tabularnewline
		\hline 
		\selectlanguage{american}%
		\textbf{PP}\selectlanguage{english}%
		& \selectlanguage{american}%
		10846\selectlanguage{english}%
		& \selectlanguage{american}%
		2.3375\selectlanguage{english}%
		& \selectlanguage{american}%
		\fe{}0.00000\selectlanguage{english}%
		& \selectlanguage{american}%
		10177.71\selectlanguage{english}%
		& \selectlanguage{american}%
		2.4672\selectlanguage{english}%
		& \selectlanguage{american}%
		\fe{}0.00000\selectlanguage{english}%
		& \selectlanguage{american}%
		9123.03\selectlanguage{english}%
		\tabularnewline
		\hline 
		\selectlanguage{american}%
		\textbf{DC3}\foreignlanguage{english}{, \textbf{$\omega=0$}}\selectlanguage{english}%
		& \multirow{5}{*}{\selectlanguage{american}%
			\textbf{10846}\selectlanguage{english}%
		} & \selectlanguage{american}%
		0.1978\selectlanguage{english}%
		& \selectlanguage{american}%
		\vi{}217.17\selectlanguage{english}%
		& \selectlanguage{american}%
		6768.66\selectlanguage{english}%
		& \selectlanguage{american}%
		0.3104\selectlanguage{english}%
		& \selectlanguage{american}%
		\vi{}254.82\selectlanguage{english}%
		& \selectlanguage{american}%
		6846.01\selectlanguage{english}%
		\tabularnewline
		\selectlanguage{american}%
		\textbf{DC3}\foreignlanguage{english}{, \textbf{$\omega=10$}}\selectlanguage{english}%
		&  & \selectlanguage{american}%
		0.8339\selectlanguage{english}%
		& \selectlanguage{american}%
		\vi{}1.87\selectlanguage{english}%
		& \selectlanguage{american}%
		6866.14\selectlanguage{english}%
		& \selectlanguage{american}%
		0.7727\selectlanguage{english}%
		& \selectlanguage{american}%
		\vi{}8.49\selectlanguage{english}%
		& \selectlanguage{american}%
		6875.68\selectlanguage{english}%
		\tabularnewline
		\selectlanguage{american}%
		\textbf{DC3}\foreignlanguage{english}{, \textbf{$\omega=100$}}\selectlanguage{english}%
		&  & \selectlanguage{american}%
		1.0064\selectlanguage{english}%
		& \selectlanguage{american}%
		\vi{}0.04475\selectlanguage{english}%
		& \selectlanguage{american}%
		6725.53\selectlanguage{english}%
		& \selectlanguage{american}%
		1.0033\selectlanguage{english}%
		& \selectlanguage{american}%
		\vi{}0.19906\selectlanguage{english}%
		& \selectlanguage{american}%
		6788.33\selectlanguage{english}%
		\tabularnewline
		\selectlanguage{american}%
		\textbf{DC3}\foreignlanguage{english}{, \textbf{$\omega=1000$}}\selectlanguage{english}%
		&  & \selectlanguage{american}%
		1.0912\selectlanguage{english}%
		& \selectlanguage{american}%
		\vi{}0.00146\selectlanguage{english}%
		& \selectlanguage{american}%
		6834.01\selectlanguage{english}%
		& \selectlanguage{american}%
		3.5102\selectlanguage{english}%
		& \selectlanguage{american}%
		\vi{}506.26\selectlanguage{english}%
		& \selectlanguage{american}%
		7183.40\selectlanguage{english}%
		\tabularnewline
		\selectlanguage{american}%
		\textbf{DC3}\foreignlanguage{english}{, \textbf{$\omega=5000$}}\selectlanguage{english}%
		&  & \selectlanguage{american}%
		1.1814\selectlanguage{english}%
		& \selectlanguage{american}%
		\vi{}0.00009\selectlanguage{english}%
		& \selectlanguage{american}%
		6790.15\selectlanguage{english}%
		& \selectlanguage{american}%
		2.7504\selectlanguage{english}%
		& \selectlanguage{american}%
		\vi{}618.55\selectlanguage{english}%
		& \selectlanguage{american}%
		7431.77\selectlanguage{english}%
		\tabularnewline
		\hline 
		\selectlanguage{american}%
		\textbf{Bar}\selectlanguage{english}%
		& \multicolumn{7}{c|}{\selectlanguage{american}%
			\vi{}Does not support quadratic constraints\selectlanguage{english}%
		}\tabularnewline
		\hline 
		\selectlanguage{american}%
		\textbf{RAYEN}\selectlanguage{english}%
		& \selectlanguage{american}%
		\textbf{10846}\selectlanguage{english}%
		& \selectlanguage{american}%
		\textbf{1.0144}\selectlanguage{english}%
		& \selectlanguage{american}%
		\fe{}0.00000\selectlanguage{english}%
		& \selectlanguage{american}%
		\textbf{4.86}\selectlanguage{english}%
		& \selectlanguage{american}%
		\textbf{1.0164}\selectlanguage{english}%
		& \selectlanguage{american}%
		\fe{}0.00000\selectlanguage{english}%
		& \selectlanguage{american}%
		\textbf{5.11}\selectlanguage{english}%
		\tabularnewline
		\hline 
		\hline 
		\selectlanguage{american}%
		\textbf{Gurobi}\selectlanguage{english}%
		& \selectlanguage{american}%
		-\selectlanguage{english}%
		& \selectlanguage{american}%
		1.0000\selectlanguage{english}%
		& \selectlanguage{american}%
		\fe{}0.00000\selectlanguage{english}%
		& \selectlanguage{american}%
		7351.42\selectlanguage{english}%
		& \selectlanguage{american}%
		1.0000\selectlanguage{english}%
		& \selectlanguage{american}%
		\fe{}0.00000\selectlanguage{english}%
		& \selectlanguage{american}%
		7130.43\selectlanguage{english}%
		\tabularnewline
		\hline 
	\end{tabular}
	\par\end{centering}
}
	\vspace{0.4cm}
	\newline
	\begin{centering}
		\includegraphics[width=0.98\columnwidth]{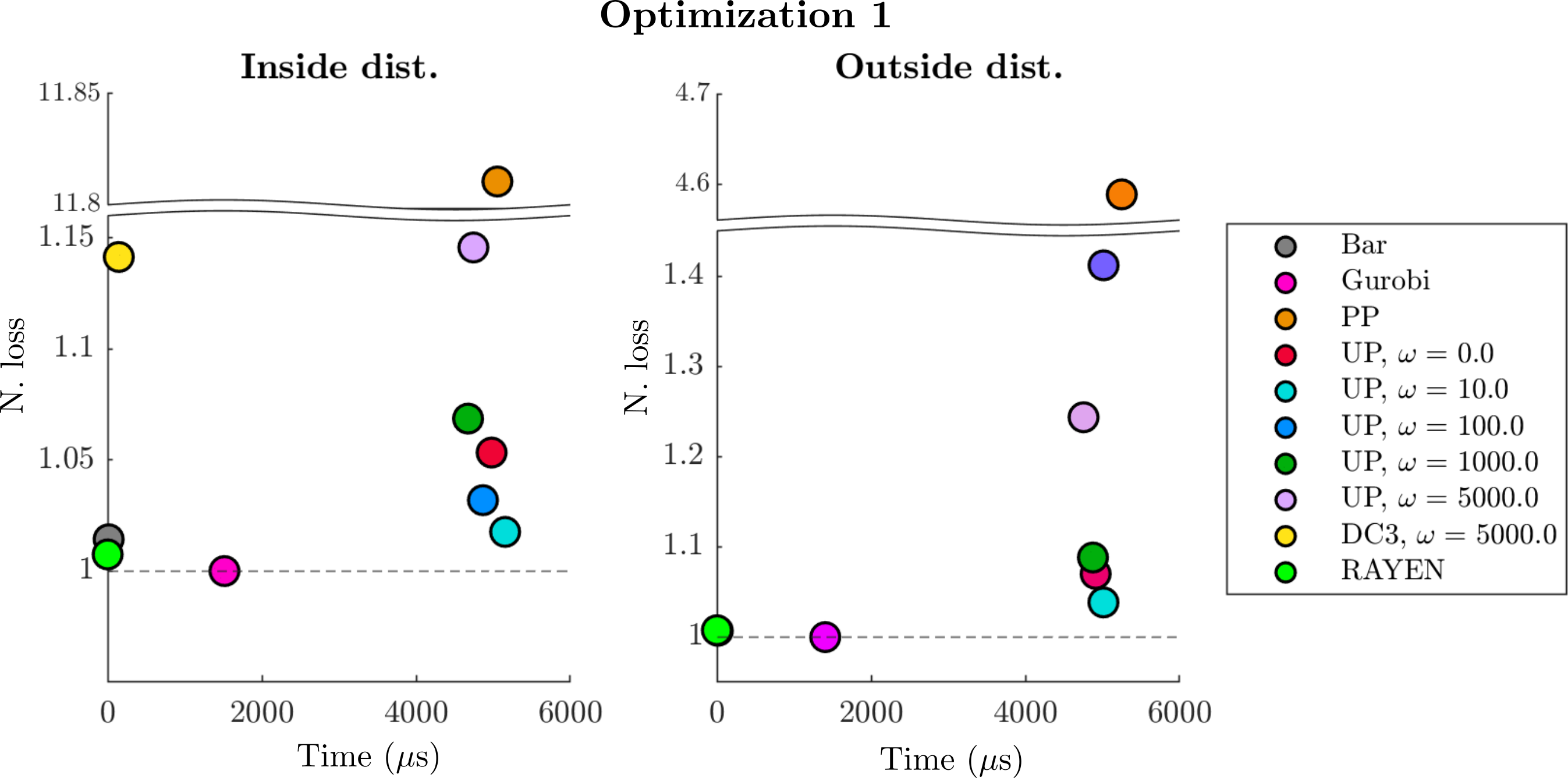}\hspace{0.5cm}\includegraphics[width=0.98\columnwidth]{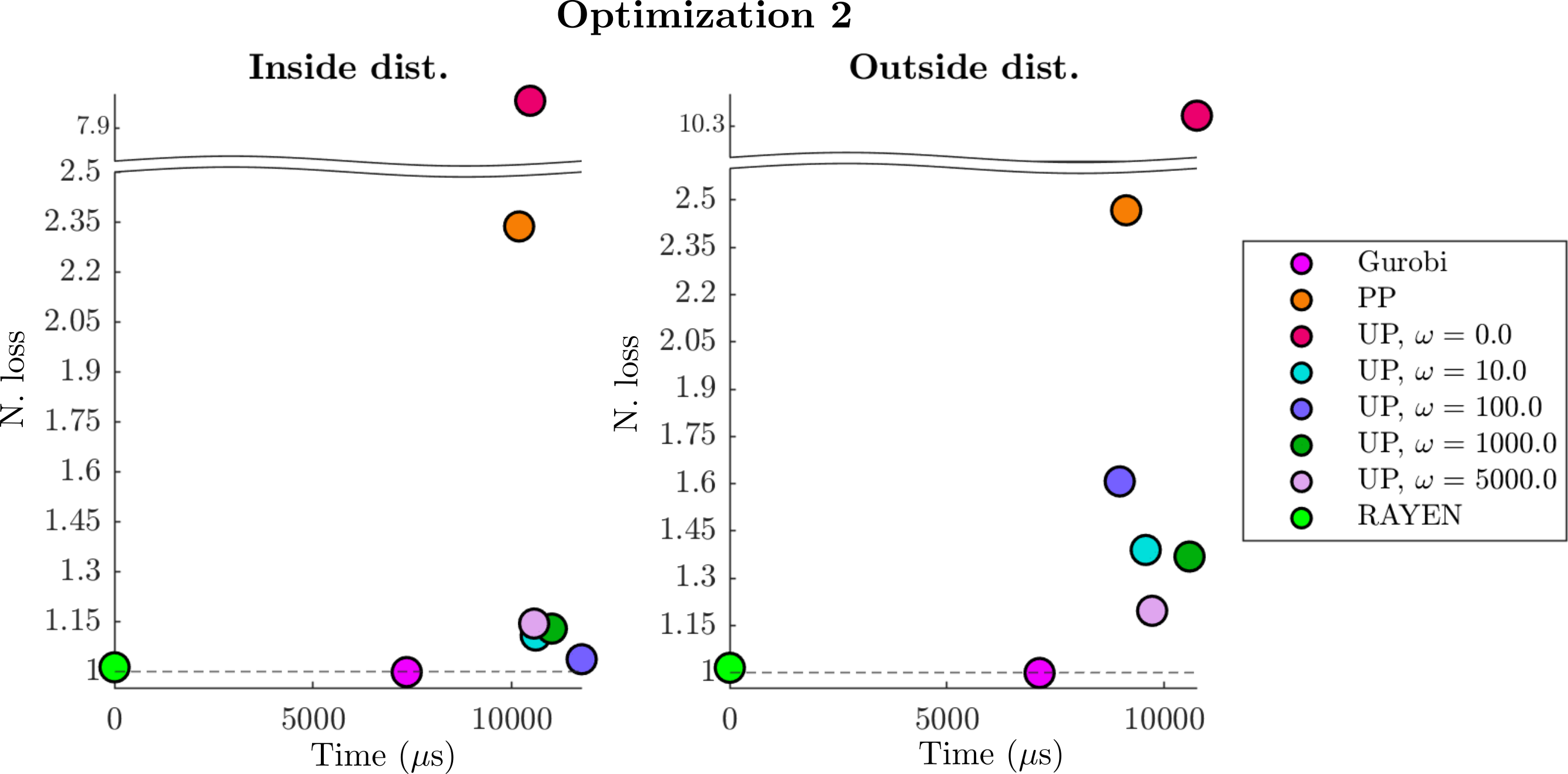}
		\par\end{centering}
	\captionof{figure}{Comparison of the computation time vs. the normalized loss \added{(defined as the loss obtained divided by the globally-optimal loss obtained by Gurobi)} for the
		methods that generate feasible solutions. The dashed line corresponds
		to a normalized cost of 1. \label{fig:loss_vs_time_and_params}}

\end{figure*}

We compare RAYEN with the following approaches (see also Table \ref{tab:all_algorithms}):
\begin{itemize}
\item \textbf{UU: }Both training and testing are \textbf{U}nconstrained.
This method produces directly $\boldsymbol{y}=\boldsymbol{v}$. A
soft cost, described below, is added to the training loss to penalize
the violation of the constraints. 
\item \textbf{UP:} When training, this method simply outputs $\boldsymbol{f}(\boldsymbol{v})$
(see Eq. \ref{eq:f_definition}), meaning that $\boldsymbol{y}$ is
\textbf{U}nconstrained with respect to the inequality constraints.
When testing, this method outputs $\boldsymbol{f}(\boldsymbol{v}')$,
where $\boldsymbol{v}'$ is the orthogonal \textbf{P}rojection of
$\boldsymbol{v}$ onto $\mathcal{Z}$. The projection is computed
using \cite{agrawal2019differentiable}. This method also has a soft
cost in the training loss to penalize the violation of the constraints. 
\item \textbf{PP:} The orthogonal \textbf{P}rojection onto $\mathcal{Z}$
is performed during both training and testing. These projections are
implemented using \cite{agrawal2019differentiable}. 
\item \textbf{DC3: }Algorithm proposed by \cite{donti2021dc3}. Completion
is used to enforce the equality constraints, and then an inner gradient
decent method (which is also performed in the testing phase) enforces
the inequality constraints. This method also includes a soft cost
in the training loss to penalize the violation of the constraints.
\item \textbf{Bar:} Generalization of the barycentric coordinates method
\cite{frerix2020homogeneous} (proposed for constraints $\boldsymbol{A}\boldsymbol{y}\le\boldsymbol{0}$)
for any constraint of the type $\boldsymbol{A}\boldsymbol{y}\le\boldsymbol{b}$.
This method only supports linear constraints. The matrices
of vertices $\boldsymbol{V}$ and rays $\boldsymbol{R}$ (see Table
\ref{tab:all_algorithms}) are computed offline using the double description
method \cite{motzkin1953double,fukuda2005double} on the H-representation
(half-space representation) of~$\mathcal{Y}_{L}$. As detailed in Table~\ref{tab:all_algorithms}, the $\text{softmax}(\cdot)$
operator produces the weights for the convex combination of the vertices,
while the\noun{ }$\text{abs}(\cdot)$ operator produces the weights
for the conical combination of the rays. 
\end{itemize}
Defining the following soft costs:
\begin{eqnarray}
p_{\text{soft},L}&:=&\left\Vert \text{relu}\left(\boldsymbol{A}_{1}\boldsymbol{y}-\boldsymbol{b}_{1}\right)\right\Vert ^{2}+\left\Vert \boldsymbol{A}_{2}\boldsymbol{y}-\boldsymbol{b}_{2}\right\Vert ^{2}\label{eq:soft_L} \\
p_{\text{soft},Q}&:=&\sum_{i=0}^{\eta-1}\left\Vert \text{relu}\left(g_{i}\left(\boldsymbol{y}\right)\right)\right\Vert ^{2}\label{eq:soft_q}
\end{eqnarray}
the training and test losses in each algorithm are defined in Table
\ref{tab:definitions_problems}. In these losses, the soft costs are
weighted with $\omega\ge0$. The methods \textbf{Bar}, \textbf{PP},
and \textbf{RAYEN} do not need a soft cost since all the constraints
are guaranteed to be satisfied both during training and testing. Note
also that in \textbf{DC3}, the term $\left\Vert \boldsymbol{A}_{2}\boldsymbol{y}-\boldsymbol{b}_{2}\right\Vert ^{2}$
of $p_{\text{soft},L}$ will always be zero since this algorithm satisfies
the equality constraints by construction.

All the methods are implemented in \noun{Pytorch} \cite{paszke2019pytorch},
and the Adam optimizer \cite{kingma2015adam} with a learning rate
of $10^{-4}$ is used for training. A total of 2000 epochs are performed
for training, and the policy with the best validation loss is selected
for testing. For each optimization problem, a total of 1216 samples
of $\boldsymbol{\gamma}\in\mathbb{R}^{n_{\boldsymbol{\gamma}}}$ are drawn
from uniform distributions detailed in Appendix \ref{subsec:Training,-validation,-and},
and $\approx70\%$ of them are used in the training set and $\approx30\%$
in the validation set. The batch size for training is 256. \added{All these results were obtained with the double-precision floating-point format and using a desktop equipped with an Intel Core i9-12900 \texttimes{} 24, 64GB of RAM, and an NVIDIA GeForce RTX 4080.}

For each optimization problem $j\in\{1,2\}$, we use two testing sets
(each one with 512 samples $\boldsymbol{\gamma}\in\mathbb{R}^{n_{\boldsymbol{\gamma}}}$):
$\text{TE}_{j,\text{in}}$ and $\text{TE}_{j,\text{out}}$. $\text{TE}_{j,\text{in}}$
uses the same distribution as the one used for training, while $\text{TE}_{j,\text{out}}$
does not. Details on these distributions are available in Appendix
\ref{subsec:Training,-validation,-and}. The goal of using the test
set $\text{TE}_{j,\text{out}}$ is to study how well the trained policy
generalizes to unseen inputs. The results for these test sets are
shown in Table~\ref{tab:results_optimization} and \added{plotted in} Fig.~\ref{fig:loss_vs_time_and_params}. \added{In these results, the violation
is defined as the squared distance to the closest point in the feasible
set. The loss is normalized with the globally-optimal loss obtained
by Gurobi~\cite{gurobi}. Hence, a normalized loss smaller than $1$ implies that
the solution violates the constraints. Both problems 1 and
2 are convex, and therefore the optimal solution found by Gurobi is
globally optimal}.

\begin{figure}
	\centering
	\begin{subfigure}[b]{1.0\columnwidth}
		\centering
		\includegraphics[width=\textwidth]{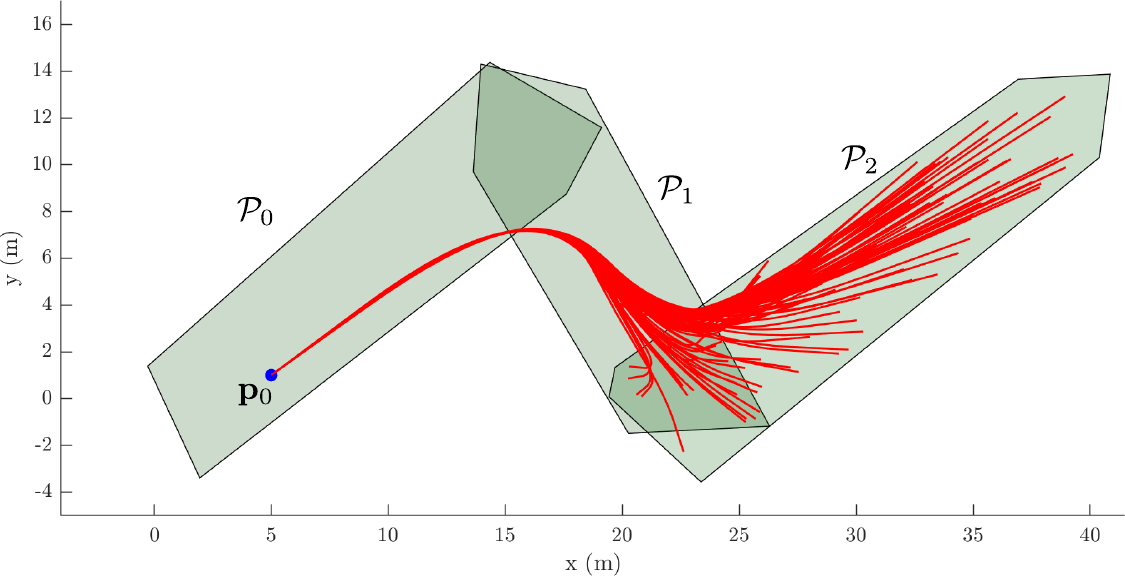}
		\caption{Optimization 1. It has $n_{\text{polyh}}=3$.}
		\label{fig:trajectories2d_processed}
	\end{subfigure}
	\vskip 1cm
	\begin{subfigure}[b]{1.0\columnwidth}
		\centering
		\includegraphics[width=\textwidth]{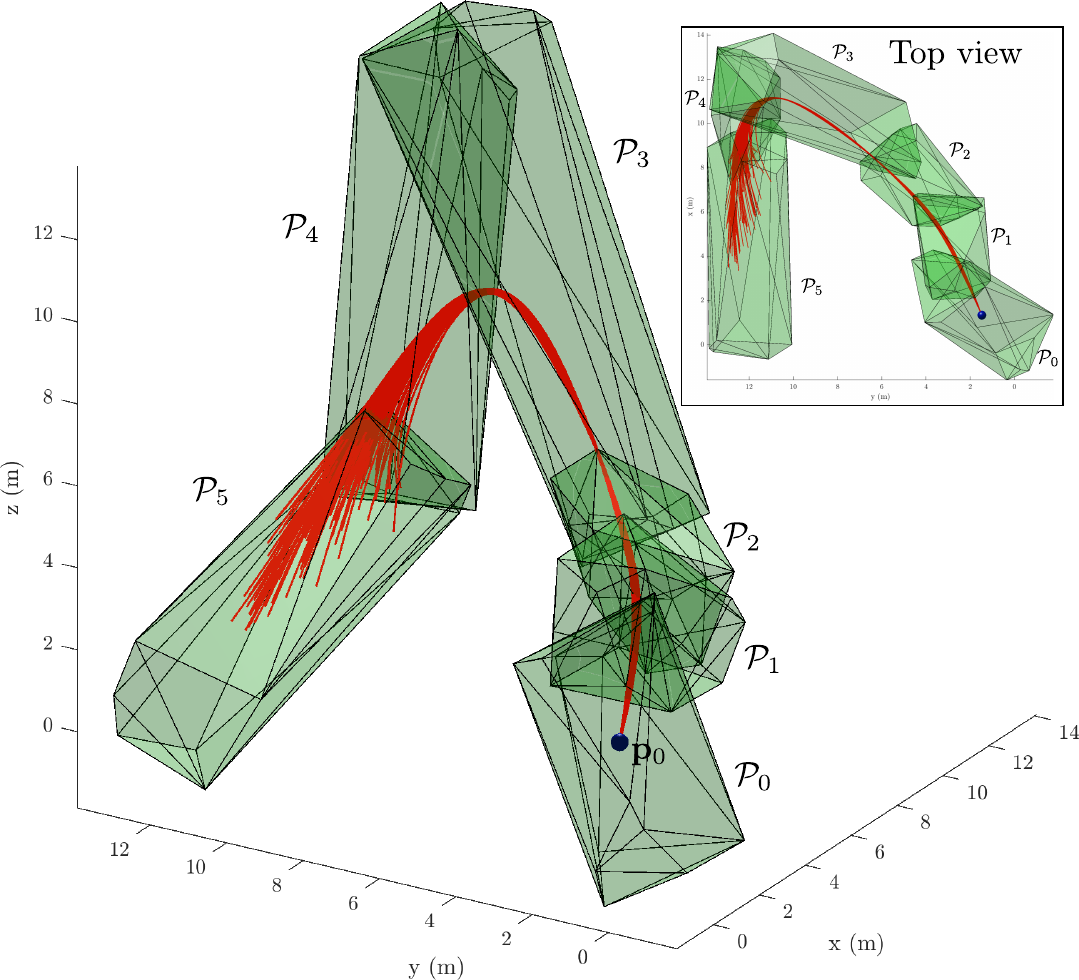}
		\caption{Optimization 2. It has $n_{\text{polyh}}=6$.}
		\label{fig:trajectories3d_processed}
	\end{subfigure}
	\caption{Trajectories generated by the trained network equipped with RAYEN. For each optimization $j\in\{1,2\}$, each plot shows the trajectories obtained using 100 different~$\boldsymbol{\gamma}$ from the testing set $\text{TE}_{j,\text{in}}$. $\mathbf{p}_0$ is the initial condition, and $\mathcal{P}_i$ denotes each polyhedron. The final position $\mathbf{p}_f$ is taken randomly inside the last polyhedron. }
	\label{fig:trajectories_processed}
\end{figure}

The following conclusions
can be extracted from these results:

\textbf{\uline{Same distribution as in training:}}
\begin{itemize}
\item \uline{Optimization 1}: \textbf{RAYEN} is able to obtain the normalized
loss closest to the globally-optimal one, while guaranteeing a zero
violation of the constraints. The methods \textbf{UP}, \textbf{PP}, \textbf{DC3} with $\omega=5000$, \textbf{Bar}, and \textbf{RAYEN}
generate feasible solutions. In terms of computation time, \textbf{RAYEN}
is between 20 and 7468 times faster than these methods. 
\item \uline{Optimization 2:} Again, \textbf{RAYEN} is able to obtain
the normalized loss closest to the globally-optimal one while guaranteeing
a zero violation of the constraints. Compared to the feasible approaches
(\textbf{UP} and \textbf{PP}), \textbf{RAYEN}'s computation time is
between 2094 and 2417 times faster. 
\end{itemize}
\textbf{\uline{Generalization: Different distribution as in training:}}
\begin{itemize}
\item \uline{Optimization 1}: Both\textbf{ RAYEN} and \textbf{Bar} are
able to obtain normalized losses very close to the globally-optimal one,
while guaranteeing a zero violation of the constraints. Although \textbf{Bar}'s
loss is slightly better than \textbf{RAYEN}'s (1.0070 vs. 1.0076),
\textbf{Bar}'s method is 16 times slower, and it requires 197 times
as many parameters as \textbf{RAYEN}. This high number of parameters
required by \textbf{Bar} is due to the high number of vertices of
the polyhedron $\mathcal{Y}$. Note also that \textbf{Bar }only supports
linear constraints.
\item \uline{Optimization 2:} \textbf{RAYEN} is the algorithm that has
the best generalization (smallest normalized loss) while achieving
a zero violation of the constraints. 
\end{itemize}
It is also important to note that, when tested in the same distribution
as the training set, the violations of both \textbf{UU} and \textbf{DC3}
tend to decrease with higher values of $\omega$. These violations
however tend to become quite large when using a different distribution
than the one used in training. 

Using RAYEN, some examples of the trajectories generated by the trained
network for 100 different $\boldsymbol{\gamma}$ taken from the testing
set $\text{TE}_{1,\text{in}}$ (see Appendix~\ref{subsec:Training,-validation,-and})
are shown in Fig.~\ref{fig:trajectories_processed}. \added{The initial and final boundary conditions, kinematic limits (velocity, acceleration, and jerk), and collision-avoidance constraints (defined by the green safety corridor) are guaranteed to be satisfied at all times.}\footnote{As the degree of the spline of the Optimization 1 is 2, then the term $\int_{0}^{t_{f}}\left\Vert \mathbf{j}(t)\right\Vert ^{2}dt$
	is clearly $0$, and therefore $\alpha_{\text{j}}$ does not affect
	the optimal solution. We keep this term simply for consistency purposes
	with Optimization 2, but it could also be removed as well. }

\begin{figure}
	\begin{centering}
		\includegraphics[width=1\columnwidth]{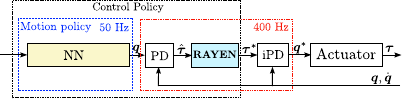}
		\par\end{centering}
	\caption{\added{The control architecture for the quadruped robot employs a two-tiered hierarchical policy: a 50 Hz high-level Neural Network (NN) motion policy that generates desired joint positions $\boldsymbol{q}$, and a 400 Hz low-level safety and tracking loop. Within the faster loop, a PD controller computes an unconstrained preliminary torque $\hat{\boldsymbol{\tau}}$ based on the error between the desired and actual joint states ($\boldsymbol{q}, \dot{\boldsymbol{q}}$). This torque is then passed through RAYEN, outputting a violation-free torque $\boldsymbol{\tau}^*$. 
			Finally, an inverse PD (iPD) controller maps it back into a safe desired joint position command $\boldsymbol{q}^*$ for the physical actuators to execute.
    \label{fig:torque_limits}}}
\end{figure}

\begin{addedstuff}

\subsection{\added{Torque Constraints on Legged Robot} \label{subsec:results_legged}}

In recent years, the use of neural networks for locomotion policies in legged robotics has shown unprecedented performance and robustness~\cite{lee2020learning,miki2022learn,rudin2022advanced,li2024reinforcement}. Typically trained through reinforcement learning, the system's hardware limits are either handled directly through the simulator, or implicitly handled through shaping rewards or termination of the training episode~\cite{rudin2022learning}. Although constrained RL methods exist~\cite{kim2023not,lee2024exploring}, they only enforce constraints in an expected value fashion. Especially for torque-controlled robots, training a policy that guarantees (or at least that is aware) of the actuator limits is vital. 

We present a case study in which RAYEN is used to enforce four constraints commonly encountered in legged robotics literature \cite{lee2024exploring, valsecchi2023barry, ma2023learning}. While the first constraint can be handled via simple clipping, the remaining three introduce coupling among the admissible leg torques, rendering independent clipping infeasible.
\begin{itemize}
   \item Joint torque limits with a box constraint
\begin{equation}
    \label{eq:torque_box_constr}
    \boldsymbol{\tau}_{\mathrm{min}}  \leq  \boldsymbol{\tau} \leq \boldsymbol{\tau}_{\mathrm{max}}.
\end{equation}
Here, the vector $\boldsymbol{\tau}$ contains all the individual joint torques, the inequalities are element-wise, and  $\boldsymbol{\tau}_{\mathrm{min}}$ and $\boldsymbol{\tau}_{\mathrm{max}}$ are vectors that represent, respectively, the lower and upper torque limits. Note that $\boldsymbol{\tau}_{\mathrm{min}}$ and $\boldsymbol{\tau}_{\mathrm{max}}$ are usually symmetrical, but could also depend on other characteristic motor variables, as in~\citet{shin2023actuator}.

    \item Sum of absolute torques over all $n_j$ joints (weighted $\ell_1$-ball)

    \begin{equation}
    \label{eq:tau_sum_global}
    \sum_{i=1}^{n_j} c_i \, \bigl| \tau_i \bigr| \;\le\; I_{\mathrm{abs}},
    \end{equation}
which directly limits the total current draw\footnote{We employ the common assumption that the drawn current per joint is proportional to the effective torque.} $I_{\mathrm{abs}}$, in practice usually limited by a fuse.

    \item Sum of absolute torques per leg (weighted $\ell_1$-ball)
    \begin{equation}
        \label{eq:tau_sum_per_leg}
        \sum_{i\, \in \, \mathrm{leg}} c_i \, \bigl| \tau_i \bigr| \;\le\; I_{\mathrm{leg}}.
    \end{equation}

    This constraint limits the total current draw $I_{\mathrm{leg}}$ of all joints per leg, in practice usually limited by a dedicated fuse or in our case by a power cable connector type.

    \item Limit on torques squared (weighted $\ell_2$-ball)
    \begin{equation}
        \label{eq:power_constr}
        \|\boldsymbol{C} \, \boldsymbol{\tau}\|^2_2 \leq p_{\text{tot}},
    \end{equation}
    Assuming a proportional relation of torque squared and thermal power losses \cite{valsecchi2023barry}, this constraint can be leveraged to regularize the motor temperatures implicitly. Overheating can be mitigated by limiting the total cooling power $p_{\text{tot}}$, whereby $\boldsymbol{C}$ can be adjusted to account for different thermal constants.
\end{itemize}

As a concrete example application, we apply the torque constraints to a locomotion policy which we train to follow base velocity commands, with the same setup as presented in \citet{rudin2022learning}. During training, we enforce all of the above constraints using RAYEN. The feasible region in torque space therefore results in the intersection of weighted $\ell_1$, $\ell_2$, and $\ell_\infty$-balls (i.e., the intersection of ellipsoidal, diamond-like, and box-like constraints). As previous work has shown~\cite{hwangbo2019learning}, training a motion policy using joint impedance position references as action space has several training and robustness benefits over training with torque output directly, e.g., allowing the policy to be evaluated at a lower frequency. To satisfy the constraints on torque level, RAYEN is therefore applied to the higher frequency torque, as shown in Fig.~\ref{fig:torque_limits}.

\end{addedstuff}

\begin{figure*}
	\begin{centering}
		\includegraphics[width=\textwidth]{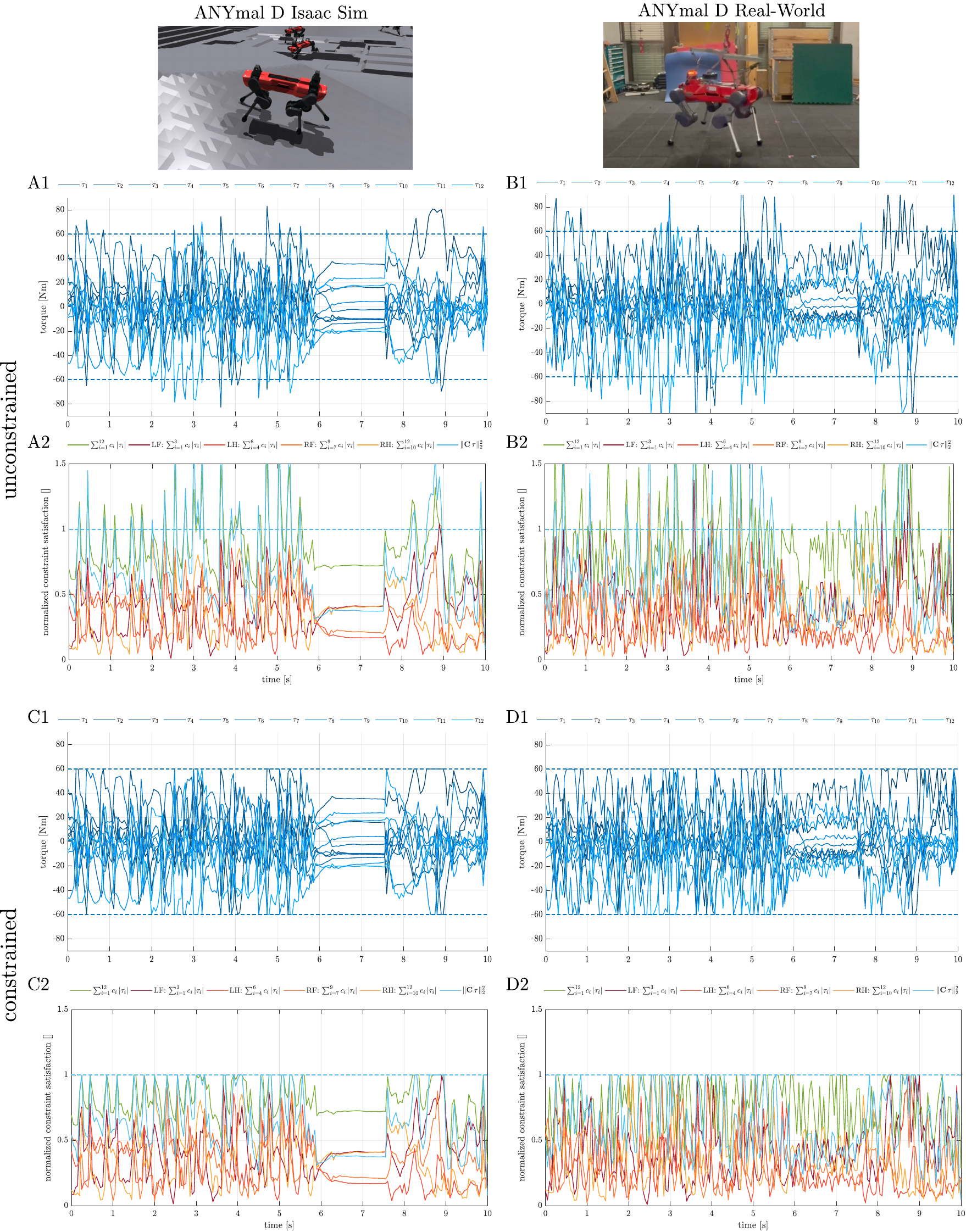}
		\par\end{centering}
	\caption{\added{Robot experiments with and without RAYEN in simulation and on the real-world ANYmal D quadruped for a fixed command sequence. We show the joint torque commands produced by the locomotion policies in subfigures denoted by "1"s, and corresponding constraint functions of the ball constraints in the subfigures denoted by "2"s. Subfigures "A" and "B" show the unconstrained case in simulation and the real-world, respectively, and we can clearly see violation in all four constraint groups, whereby the limits are denoted by dotted lines. "C" and "D" show the constrained case in simulation and the real-world, respectively, and we can clearly see strict constraint satisfaction of all four constraint groups.}}
	\label{fig:exp}
\end{figure*}

\begin{figure*}
	\centering
	\begin{subfigure}{.23\textwidth}
		\centering
		\includegraphics[width=1.0\linewidth]{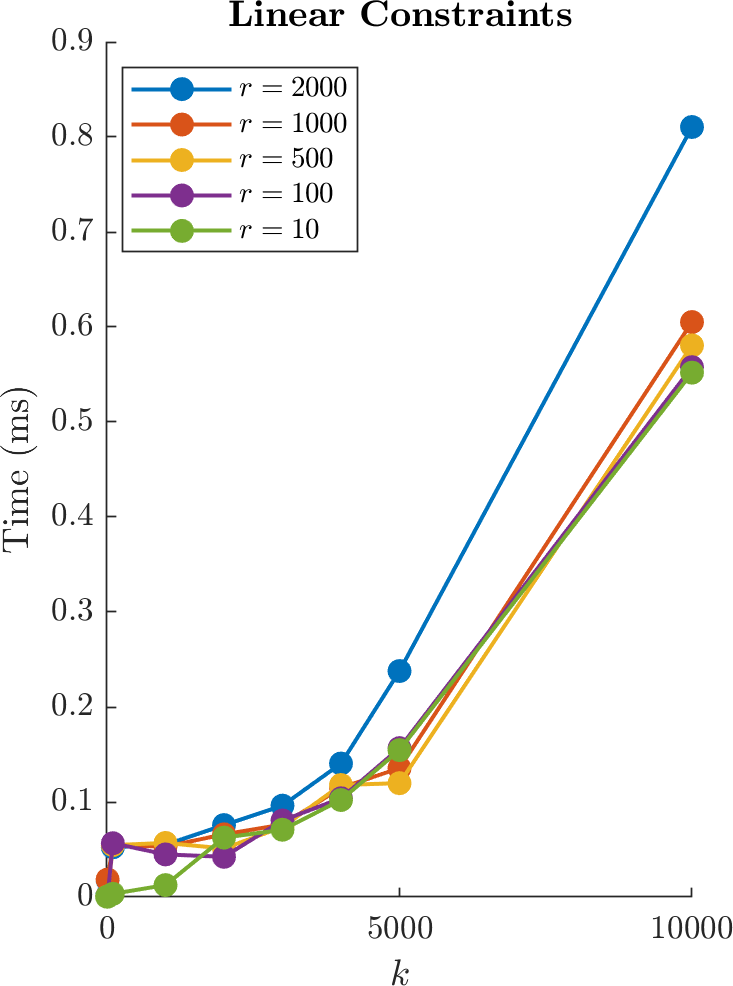}
		\caption{$\boldsymbol{A}_1\in\mathbb{R}^{r\times k}$ }
	\end{subfigure}%
	\begin{subfigure}{.21\textwidth}
		\centering
		\includegraphics[width=1.0\linewidth]{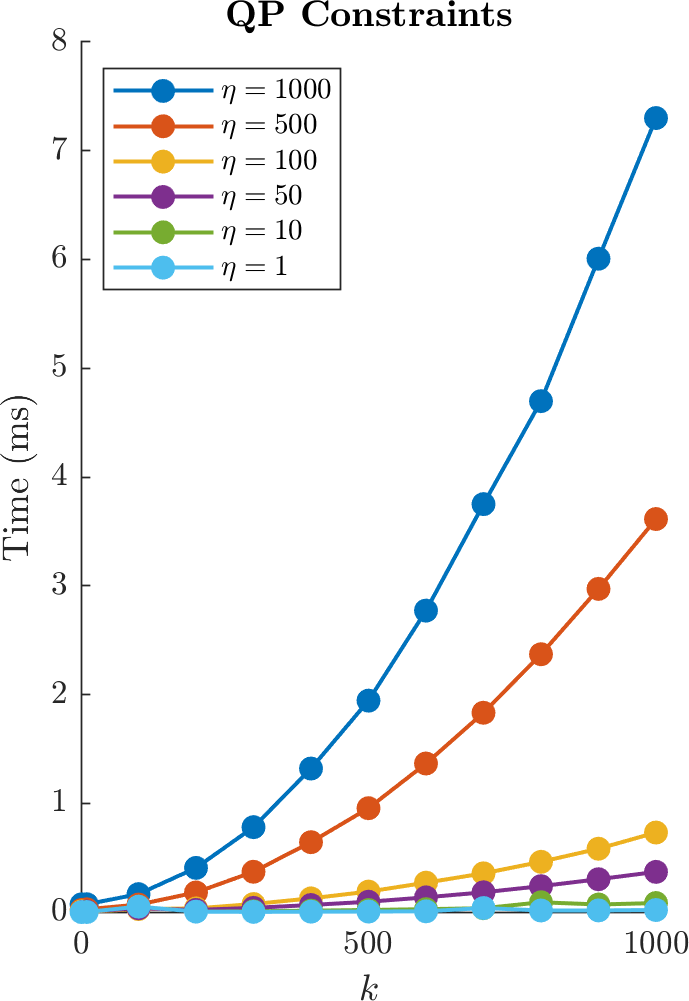}
		\caption{$\boldsymbol{P}_{i}\in\mathbb{R}^{k\times k}$}
	\end{subfigure}%
	\begin{subfigure}{.32\textwidth}
		\centering
		\includegraphics[width=1.0\linewidth]{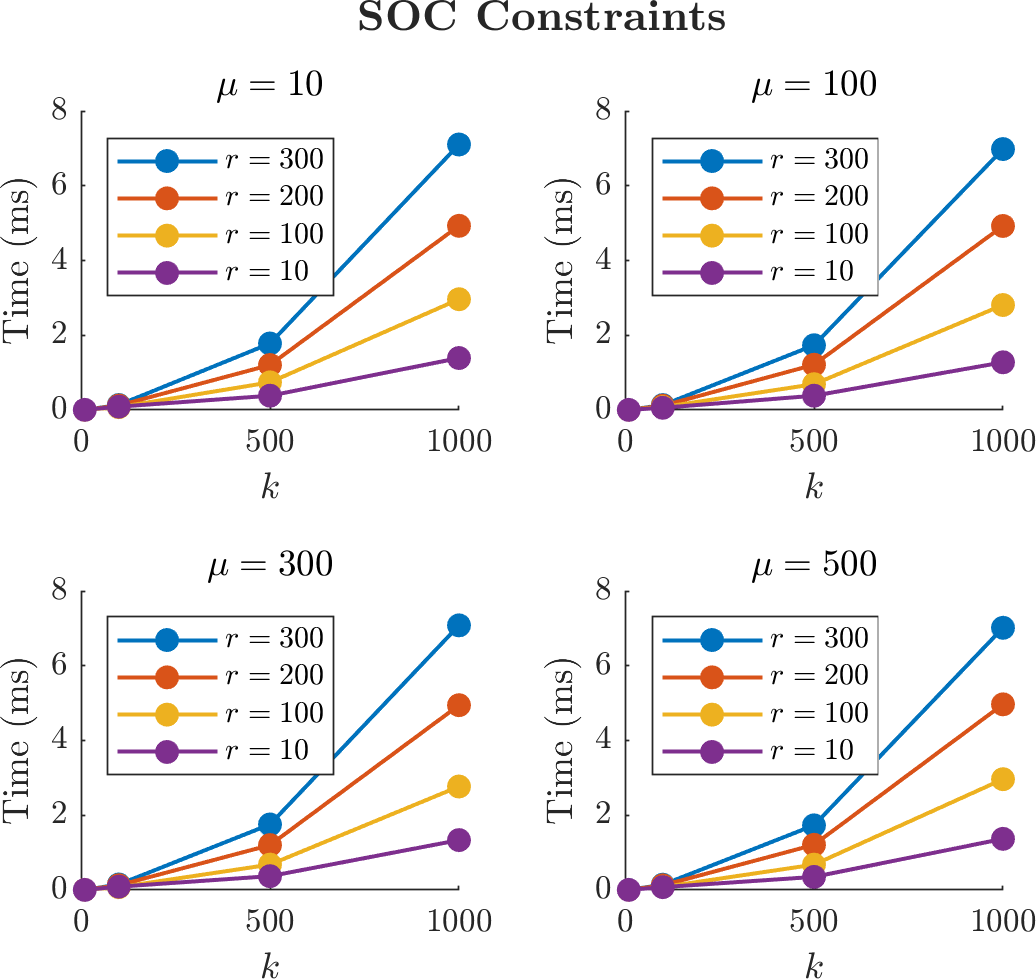}
		\caption{$\boldsymbol{M}_{j}\in\mathbb{R}^{r\times k}$}
	\end{subfigure}%
	\begin{subfigure}{.23\textwidth}
		\centering
		\includegraphics[width=1.0\linewidth]{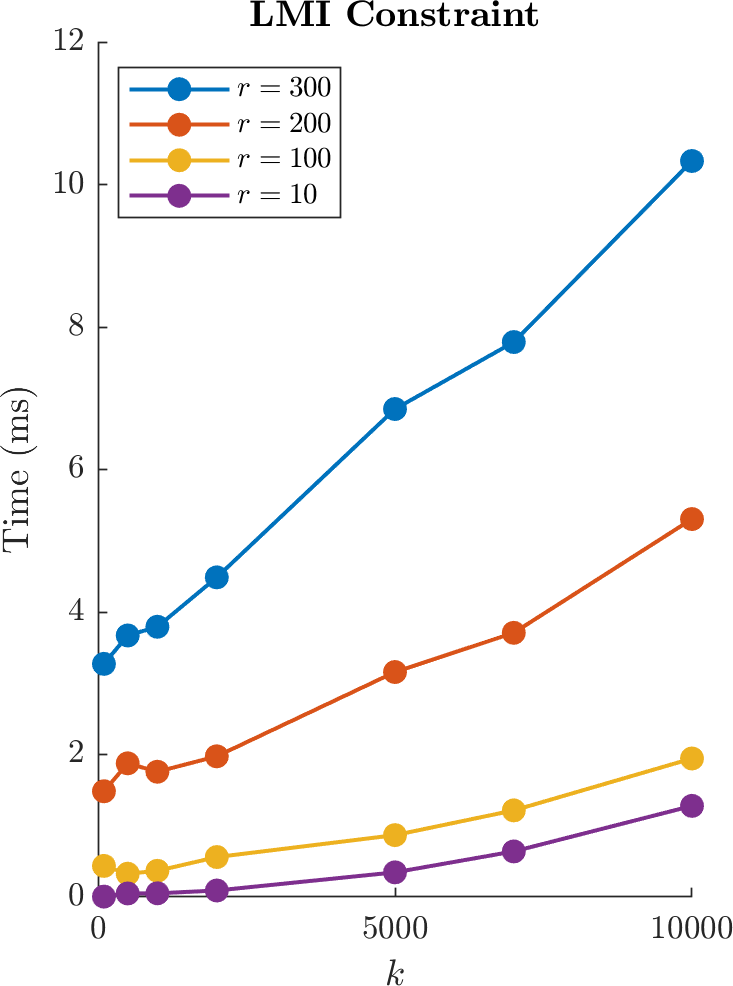}
		\caption{$\boldsymbol{F}_{i}\in\mathbb{R}^{r\times r}$}
	\end{subfigure}
	\caption{Computation times of RAYEN for different dense random constraints of varying size. In these plots, $\boldsymbol{y}\in\mathbb{R}^{k}$, $\eta$ is the number of quadratic constraints (see Eq.~\ref{eq:convex_quad}), and $\mu$ is the number of SOC constraints (see Eq.~\ref{eq:soc}). The plots show the inference time for each sample in a batch of 2000 samples. The double-precision floating-point format \noun{float64} was used.}
	\label{fig:time_analysis}
\end{figure*}

\added{Fig.~\ref{fig:exp} shows the results of the multiple robot experiments conducted with and without RAYEN, both in simulation and on the real-world ANYmal D quadruped. We show the joint torque commands produced by the locomotion policies and the constraint functions of the ball constraints. Subfigures "A" and "B" show the unconstrained case in simulation and the real-world, respectively. In this baseline setting, we can clearly see violation in all four constraint groups, whereby the limits are denoted by dotted lines. In contrast, subfigures "C" and "D" show the constrained case in simulation and the real-world, respectively. By utilizing the proposed method, we can clearly see strict constraint satisfaction of all the constraints. These results demonstrate that the RAYEN framework reliably enforces the prescribed physical limits during operation on the physical ANYmal D quadruped, successfully bridging the gap between simulation and real-world deployment.}

\subsection{Computation time \label{subsec:Computation-time}}

\begin{figure}
	\begin{centering}
		\includegraphics[width=1\columnwidth]{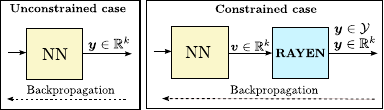}
		\par\end{centering}
	\caption{\added{The difference in computation time between the unconstrained and constrained architectures is determined by the computation time of the \myboxblue[]{RAYEN module}. Here, we take $n=m=k$ (see Table \ref{tab:Notation}}). \label{fig:pipeline_comparison_comp_time}}
	\vskip-2ex
\end{figure}

\added{A natural question to ask is what is the additional computation time of a neural network equipped with RAYEN compared to the same neural network without RAYEN (and therefore, without constraint guarantees). For this analysis, we will take $n=m=k$ and choose two networks with the same architecture (Fig.~\ref{fig:pipeline_comparison_comp_time}). Once trained, the only difference between the \myboxyellow[]{networks} is in the weights, since one of them has been trained with RAYEN and another one without it. Hence, differences in computation time at testing time will be primarily determined by the computation time of the
\myboxblue[]{RAYEN module} (Fig.~\ref{fig:pipeline_comparison_comp_time}). To study the computation time of this module,}
we generate random dense constraints of varying dimensions and measure the computation
time required by RAYEN to impose these constraints on $\boldsymbol{y}\in\mathbb{R}^{k}$
for a random $\boldsymbol{v}\in\mathbb{R}^{n}$ (Fig.~\ref{fig:pipeline_comparison_comp_time}).
All the details of how these random constraints are generated are
available in the code \added{released}. 
\added{We run RAYEN with the double-precision floating-point format and on a desktop  with an Intel Core i9-12900 \texttimes{} 24, 64GB of RAM, and an NVIDIA GeForce RTX 4080.}
The computation times per sample (in
a batch of 2000 samples) achieved by RAYEN are shown in Fig.~\ref{fig:time_analysis}. Note that the \added{computation overhead is negligible,} even for
high-dimensional variables, as shown in the following statistics:
\begin{itemize}
\item 2K linear constraints on a 10K-dimensional variable: $<0.9$ ms. 
\item 1K dense convex quadratic constraints on a 1K-dimensional variable:
$<8$ ms. 
\item 500 dense SOC constraints on a 1K-dimensional variable with matrices
$\boldsymbol{M}_{j}$ of 300 rows: $<8$ ms. 
\item A dense $300\times300$ LMI constraint on a 10K-dimensional variable:
$<11$ ms.
\end{itemize}

\subsection{\addednew{Ablation: Differentiating through RAYEN}\label{subsec:ablation_ste}}

\begin{addedstuffnew}
To isolate the benefit of differentiating through RAYEN, we compare it with a stop-gradient variant, denoted as RAYEN-STE (see Fig.~\ref{fig:RAYEN_STE}). This variant uses the straight-through estimator: its forward pass is identical to the one of RAYEN, but its backward pass treats RAYEN as if it were doing $\boldsymbol{y}=\boldsymbol{f}(\boldsymbol{z}_{0}+\boldsymbol{v})$. Both methods therefore produce exactly the same outputs for the same weights, and guarantee constraint satisfaction during both training and testing. The only difference is the gradient that reaches the upstream layers: RAYEN uses the exact Jacobian of Eq.~\ref{eq:z1_definition}, while RAYEN-STE backpropagates as if the network were unconstrained. In PyTorch this can be easily implemented using $$\boldsymbol{f}\left(\boldsymbol{z}_0+\boldsymbol{v}+(\boldsymbol{z}_1-\boldsymbol{z}_0-\boldsymbol{v}).\text{detach}()\right)$$

\begin{figure}
	\begin{centering}
		\includegraphics[width=0.7\columnwidth]{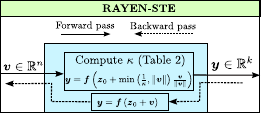}
		\par\end{centering}
	\caption{\addednew{In \textbf{RAYEN-STE}, the forward pass (solid arrows) is identical to the one of \textbf{RAYEN}, but in the backward pass (dashed arrows) the gradients are computed as if the output were $\boldsymbol{y}=\boldsymbol{f}\left(\boldsymbol{z}_{0}+\boldsymbol{v}\right)$.} \label{fig:RAYEN_STE}}
	\vskip-2ex
\end{figure}

We use Optimization 2 of Section~\ref{subsec:results_optimization}, with the same network architecture, training procedure, hyperparameters, and training/validation/test sets. The results are shown in Table~\ref{tab:ablation_ste}.

\begin{table}[h]
\caption{\addednew{Comparison between RAYEN and RAYEN-STE in Optimization 2. The loss is normalized with the globally-optimal loss obtained by Gurobi.} \label{tab:ablation_ste}}
\centering
\small
\setlength{\tabcolsep}{3pt}
\begin{addedstuffnew}
\begin{tabular}{lcccc}
\hline
 & \multicolumn{2}{c}{$\text{TE}_{2,\text{in}}$} & \multicolumn{2}{c}{$\text{TE}_{2,\text{out}}$}\tabularnewline
\textbf{Method} & \textbf{Norm. loss} & \textbf{Violation} & \textbf{Norm. loss} & \textbf{Violation}\tabularnewline
\hline
\textbf{RAYEN} & \textbf{1.0139} & 0.00000 & \textbf{1.0452} & 0.00000\tabularnewline
\textbf{RAYEN-STE} & 2.4711 & 0.00000 & 2.6061 & 0.00000\tabularnewline
\hline
\end{tabular}
\end{addedstuffnew}
\end{table}

As expected, both methods achieve zero violation, since they share the same forward pass. However, differentiating through RAYEN has a large impact on the solution quality: RAYEN obtains a normalized loss within $4.6\%$ of the globally-optimal one in both test sets,\footnote{\addednew{The difference with respect to the values of \textbf{RAYEN} reported in Table~\ref{tab:results_optimization} is due to the random initialization of the network.}} while RAYEN-STE obtains a loss that is approximately $147\%$ ($\text{TE}_{2,\text{in}}$) and $161\%$ ($\text{TE}_{2,\text{out}}$) higher than the optimal one. The reason is that, whenever $\boldsymbol{z}_{0}+\boldsymbol{v}\notin\mathcal{Z}$, 
RAYEN-STE backpropagates the loss gradient as if $\boldsymbol{z}_{1}$ were $\boldsymbol{z}_{0}+\boldsymbol{v}$. This mismatch between the forward and the backward passes results in update directions that do not necessarily decrease the loss of the constrained output. In contrast, the exact gradient of RAYEN captures how $\boldsymbol{z}_{1}$ moves along the boundary of $\mathcal{Z}$ when $\boldsymbol{v}$ changes, which allows the network to converge to near-optimal solutions. These results were obtained using a laptop equipped with an Intel Core Ultra 9 275HX, 32GB of RAM, and an NVIDIA GeForce RTX 5080 Laptop GPU. 
\end{addedstuffnew}

\subsection{\addednew{Preprocessing time (offline)}\label{subsec:preprocessing_time}}

\begin{addedstuffnew}
The time of each step of the offline phase (Section~\ref{sec:Offline:-Computation-of}) for Optimizations 1 and 2 of Section~\ref{subsec:results_optimization} is reported in Table~\ref{tab:preprocessing_time}. 
All the optimization problems of the offline phase are solved with Gurobi~\cite{gurobi} through its CVXPY interface~\cite{diamond2016cvxpy}. These results were obtained using the CPU of the laptop specified in Section~\ref{subsec:ablation_ste}.

\begin{table}[h]
\caption{\addednew{Preprocessing (offline phase) times for Optimizations 1 and 2. The time is in seconds, and the number and type of optimization problems solved are shown in parentheses.
} \label{tab:preprocessing_time}}
\centering
\footnotesize
\setlength{\tabcolsep}{3pt}
\begin{addedstuffnew}
\begin{tabular}{lcc}
\hline
\textbf{Step} & \textbf{Opt. 1 (s)} & \textbf{Opt. 2 (s)}\tabularnewline
\hline
Feasibility check & 0.003 (1 LP) & 0.07 (1 QCQP)\tabularnewline
Removal of redundant constraints & 0.39 (180 LPs) & 8.79 (1102 LPs)\tabularnewline
Affine hull (Section~\ref{subsec:affine-hull-of}) & 0.11 (60 LPs) & 3.75 (619 LPs)\tabularnewline
Interior point (Section~\ref{subsec:Interior-point-of}) & 0.002 (1 LP) & 0.21 (1 QCQP)\tabularnewline
Rest (e.g., set $\mathcal{Z}$, Section~\ref{subsec:set_z}) & 0.06 & 0.61\tabularnewline
\hline
\textbf{Total offline phase} & \textbf{0.57} & \textbf{13.42}\tabularnewline
\hline
\end{tabular}
\end{addedstuffnew}
\end{table}

Note that these offline preprocessing times could be heavily improved by parallelizing and by using Gurobi directly instead of using its CVXPY interface~\cite{diamond2016cvxpy}. 
Moreover, the removal of redundant constraints is only performed to reduce the computation time of the online phase. 
\end{addedstuffnew}

\section{\addednew{Extensions and Variants of RAYEN}}

This paper has focused on convex constraints in which the parameters that define
these constraints (i.e., $\boldsymbol{A}_{1},\boldsymbol{b}_{1},\boldsymbol{A}_{2},\boldsymbol{b}_{2},\boldsymbol{P}_{i},...)$
are fixed and known beforehand. \added{However, in some problems in Robotics, such as in model predictive control (MPC) or trajectory planning, the parameters that define the constraints change in each iteration, or the constraints may be nonconvex. Although we leave the handling of these generic constraints for future work\addednew{, in Sections \ref{subsec:non_fixed_constraints} and \ref{subsec:nonconvex_constraints}} we detail several ways in which RAYEN could be applied to these more generic cases. \addednew{In Section \ref{subsec:alternative_definition} we explain a variant of RAYEN that uses an alternative definition of $\boldsymbol{z}_1$}.}

\subsection{Non-fixed \added{Convex} Constraints} \label{subsec:non_fixed_constraints}

\added{In order to be able to impose convex constraints whose parameters are not fixed and/or depend on the previous layers (or input) of the network, we need to take some extra care:}
\begin{itemize}
\item The constraints must define a nonempty set at all times, and $\mathcal{Z}$
must have interior points. There are many different ways to ensure
this. For instance, if there are no equality constraints, one way
would be to ensure these conditions: 
\begin{align}
 & \boldsymbol{b}_{1}>\boldsymbol{0}\label{eq:nonfixed_b1}\\
 & r_{i}<0\;\;i=0,...,\eta-1\label{eq:nonfixed_ri}\\
 & d_{j}>\left\Vert \boldsymbol{s}_{j}\right\Vert \;\;j=0,...,\mu-1\label{eq:nonfixed_dj}\\
 & \boldsymbol{F}_{k}\succ\boldsymbol{0}\label{eq:nonfixed_Fk}
\end{align}

\begin{figure*}
	\begin{centering}
		\includegraphics[width=1\textwidth]{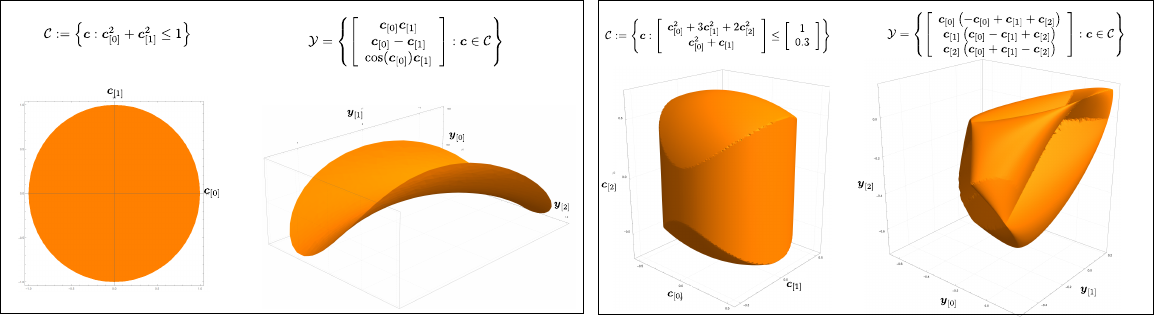}
		\par\end{centering}
	\caption{\added{Two examples of nonconvex sets $\mathcal{Y}$ for which RAYEN could still guarantee zero violation. In both cases, the nonconvex set is defined as a nonlinear transformation of the convex set $\mathcal{C}$.} \label{fig:nonconvex_examples}}
\end{figure*}

Then, we can take $\boldsymbol{z}_{0}=\boldsymbol{0}$, as it is guaranteed
to be in the interior of $\mathcal{Z}$. Eqs.~\ref{eq:nonfixed_b1}, \ref{eq:nonfixed_ri}, and~\ref{eq:nonfixed_dj} can be easily enforced by, e.g., using \noun{sigmoid}
functions. Eq. \ref{eq:nonfixed_Fk}
can be enforced by setting 
$
\boldsymbol{F}_{k}:=\boldsymbol{\Gamma}^{T}\boldsymbol{\Gamma}+\epsilon\boldsymbol{I}\;,
$, 
where $\epsilon>0$ is a predefined scalar, and where the matrix $\boldsymbol{\Gamma}$
depends on the previous layers (or input) of the network. When there
are only linear constraints, other ways to guarantee a nonempty feasible
set are explained in \cite[Section 4.1]{donti2021dc3} and \cite[Appendix A]{amos2017optnet}. 
\item $\boldsymbol{P}_{i}\succeq\boldsymbol{0}$ ($i=0,...,\eta-1$) is
required. Without any loss of generality, this can be enforced by
setting $\boldsymbol{P}_{i}:=\boldsymbol{V}_{i}^{T}\boldsymbol{V}_{i}$,
where $\boldsymbol{V}_{i}$ is a matrix that depends on the previous
layers (or input) of the network.
\item The matrices $\boldsymbol{F}_{\alpha}$ ($\alpha=0,...,k$) must be
symmetric. One way to guarantee this is by setting $\boldsymbol{F}_{\alpha}:=(\boldsymbol{\varDelta}_{\alpha}+\boldsymbol{\varDelta}_{\alpha}^{T})/2$,
where $\boldsymbol{\varDelta}_{\alpha}$ is a matrix that depends
on the previous layers (or input) of the network. 
\item $\text{aff}\left(\left\{ \boldsymbol{y}\in\mathbb{R}^{k}|\boldsymbol{A}_{1}\boldsymbol{y}\le\boldsymbol{b}_{1}\right\} \cap\mathcal{Y}_{Q}\cap\mathcal{Y}_{S}\cap\mathcal{Y}_{M}\right)=\mathbb{R}^{k}$.
Intuitively, this means that the inequality constraints must not \textit{hide}
equality constraints. We can then take $\boldsymbol{A}_{\mathcal{I}}\equiv\boldsymbol{A}_{1}$,
$\boldsymbol{b}_{\mathcal{I}}\equiv\boldsymbol{b}_{1}$, $\boldsymbol{A}_{\mathcal{E}}\equiv\boldsymbol{A}_{2}$,
and $\boldsymbol{b}_{\mathcal{E}}\equiv\boldsymbol{b}_{2}$. 
\end{itemize}
If these conditions are satisfied, then the offline phase (Section~\ref{sec:Offline:-Computation-of}) is not needed, since $\boldsymbol{z}_{0}$
is already known, and the matrices $\boldsymbol{A}_{p}$ and $\boldsymbol{b}_{p}$
can be easily found online by simply computing the nullspace of $\boldsymbol{A}_{\mathcal{E}}$
(Eq. \ref{eq:Ap_and_bp}). \addednew{To evaluate this approach, we perform an experiment in which we train a network to approximately solve the projection of $\boldsymbol{y}_{\text{target}}$ onto the (changing) feasible set while guaranteeing these changing constraints:
\begin{equation}
\begin{aligned}\underset{\boldsymbol{y}\in\mathbb{R}^{3}}{\text{min}}\quad & \left\Vert \boldsymbol{y}-\boldsymbol{y}_{\text{target}}(s)\right\Vert ^{2}\\
\text{s.t.}\quad & \boldsymbol{A}_{1}(s)\boldsymbol{y}\le\boldsymbol{b}_{1}(s)\\
 & \frac{1}{2}\boldsymbol{y}^{T}\boldsymbol{P}_{i}(s)\boldsymbol{y}+\boldsymbol{q}_{i}(s)^{T}\boldsymbol{y}+r_{i}(s)\le0\;\;i=0,1
\end{aligned}
\label{eq:nonfixed_problem}
\end{equation}
}

\begin{addedstuffnew} 
where $s$ denotes the dependency with respect to the specific sample. Hence, constraints will change in each forward pass. The parameters are randomly generated for every sample and every forward pass as follows: the entries of $\boldsymbol{A}_1\in\mathbb{R}^{6\times3}$ are sampled uniformly in $[-1,1]$, the entries of $\boldsymbol{b}_1\in\mathbb{R}^{6}$ in $[0.1,1]$ (which satisfies Eq.~\ref{eq:nonfixed_b1}), $\boldsymbol{P}_i:=\boldsymbol{V}_i^T\boldsymbol{V}_i\in\mathbb{R}^{3\times3}$ with the entries of $\boldsymbol{V}_i$ in $[-1,1]$, the entries of $\boldsymbol{q}_i\in\mathbb{R}^{3}$ in $[-1,1]$, and $r_i\in\mathbb{R}$ in $[-1,-0.1]$ (which satisfies Eq.~\ref{eq:nonfixed_ri}). Hence, $\boldsymbol{z}_0=\boldsymbol{0}$ is a known interior point that is always valid (and hence finding it in the offline phase is not needed). %
The input of the network is therefore the concatenation of all the parameters of the constraints and of the target point $\boldsymbol{y}_{\text{target}}\in\mathbb{R}^{3}$, whose entries are sampled uniformly in $[-2,2]$. We use a multilayer perceptron with four hidden layers of 512 neurons (the first three followed by ReLU activations) followed by RAYEN. It is trained using Adam with an initial learning rate of $10^{-3}$ (with cosine annealing) for $10^{5}$ iterations with a batch size of 256. New constraints are generated in each iteration. For testing, we generate 1000 new samples, and we compare the obtained loss with the globally-optimal one obtained by Gurobi~\cite{gurobi}. These results were obtained with the double-precision floating-point format, and using the laptop used in Section \ref{subsec:ablation_ste}.

\begin{table}[h]
\caption{\addednew{Results for the experiment with non-fixed constraints. The loss is normalized with the globally-optimal loss obtained by Gurobi, and the violation is computed as in Table~\ref{tab:results_optimization} (squared distance to the closest point in the feasible set, averaged over all the samples). All the times are per sample.} \label{tab:nonfixed}}
\centering
\small
\setlength{\tabcolsep}{3pt}
\begin{addedstuffnew}
\begin{tabular}{lc}
\hline
\textbf{Metric} & \textbf{Value}\tabularnewline
\hline
Normalized loss (RAYEN) & 1.0611\tabularnewline
Violation (RAYEN) & 0.00000\tabularnewline
Time to update the constraints & 0.20 $\mu$s\tabularnewline
Time of the forward pass & 3.7 $\mu$s\tabularnewline
Solve time of Gurobi & 442 $\mu$s\tabularnewline
\hline
\end{tabular}
\end{addedstuffnew}
\end{table}

The results are shown in Table~\ref{tab:nonfixed}. RAYEN guarantees the satisfaction of all these changing constraints, obtaining a loss within $6.1\%$ of the globally-optimal one. Updating the constraints only requires $0.20$ $\mu$s per sample, which is negligible compared to the forward pass of the network ($3.7$ $\mu$s per sample), and the whole forward pass is about two orders of magnitude faster than Gurobi.
\end{addedstuffnew}

\added{Another option \addednew{(needed in the case the interior point $\boldsymbol{z}_0$ is not known beforehand)} that allows non-fixed constraints is to use an implicit layer to solve the optimization problem that finds the interior point $\boldsymbol{z}_0$. This however can be very computationally expensive.}

\vspace{0.8cm}

\subsection{Nonconvex Constraints} \label{subsec:nonconvex_constraints}
\begin{addedstuff}
RAYEN could also be applied to any nonconvex set $\mathcal{Y}$ that can be expressed as a (differentiable) function of a convex set $\mathcal{C}$. In other words: 
\added{$$\mathcal{Y}=\left\{ \boldsymbol{\eta}(\boldsymbol{c}):\boldsymbol{c}\in\mathcal{C}\right\} 
\;,$$}
where $\mathcal{C}$ is a convex set (defined by Eqs. \ref{eq:linear_ineq}-\ref{eq:lmi}), and where $\boldsymbol{\eta}(\boldsymbol{c})$ is differentiable. Fig.~\ref{fig:nonconvex_examples} shows two examples of such sets. In these cases, RAYEN is applied first to obtain a point $\boldsymbol{c}\in\mathcal{C}$, and then $\boldsymbol{y}=\boldsymbol{\eta}(\boldsymbol{c})$ is applied to obtain a point $\boldsymbol{y}\in\mathcal{Y}$. Backpropagation happens through both $\boldsymbol{\eta}(\cdot)$ and RAYEN. Note also that the function $\boldsymbol{\eta}(\cdot)$ could have some parameters that depend on the input or latent variable of the network.

Moreover, and because of the way RAYEN works, it could also be applied to the cases where $\mathcal{Z}$ is a (potentially nonconvex) star-shaped set, as long as $\boldsymbol{z}_{0}$ is taken as the star point of $\mathcal{Z}$ (see, e.g., \citet[Definition 5.15]{krantz2014convex} or \citet[Definition II.2.6]{freitag2009complex}).

\end{addedstuff}

Compared to methods such as DC3 \cite{donti2021dc3}, one limitation
of the proposed framework is that it currently does not support generic nonconvex constraints. However, as shown by the results in Section
\ref{subsec:results_optimization}, when the constraints are convex
the performance of our method is much better than~\cite{donti2021dc3}. 

\subsection{\addednew{Alternative Definition of $\boldsymbol{z}_1$}\label{subsec:alternative_definition}}

\begin{addedstuffnew}
When $\boldsymbol{z}_{0}+\boldsymbol{v}\notin\mathcal{Z}$ (i.e., when $\left\Vert \boldsymbol{v}\right\Vert >\frac{1}{\kappa}$), Eq.~\ref{eq:z1_definition} becomes $\boldsymbol{z}_{1}=\boldsymbol{z}_{0}+\frac{1}{\kappa}\bar{\boldsymbol{v}}$, which only depends on $\bar{\boldsymbol{v}}$ (note that $\kappa$ depends only on $\bar{\boldsymbol{v}}$), and therefore $\boldsymbol{z}_{1}$ is independent of $\left\Vert \boldsymbol{v}\right\Vert$. Hence, the loss gradient propagated through $\boldsymbol{z}_{1}$ changes the direction of $\boldsymbol{v}$, but has no radial component that directly encourages a reduction in $\left\Vert \boldsymbol{v}\right\Vert$. Changes in direction move the output along the boundary of the feasible set $\mathcal{Z}$, and may bring the output to a region where a different constraint is active. If the optimal output lies in the interior of $\mathcal{Z}$, reaching it requires leaving the boundary. Note, however, that a change in direction may itself make the constraints inactive, since the boundary distance $1/\kappa$ depends on $\bar{\boldsymbol{v}}$.

There are several ways to mitigate this. One option \added{would be to use 
$\boldsymbol{z}_{1}:=\left(\boldsymbol{z}_{0}+
\frac{1}{e^\beta + \kappa}\bar{\boldsymbol{v}}\right)\in\mathcal{Z}\;
$
where $\arraycolsep=1.6pt\left[\begin{array}{cc}
\boldsymbol{v} & \beta\end{array}\right]^{T}$ is the output of the layer before RAYEN. Empirically, however, this approach generated worse results than the ones obtained using Eq.~\ref{eq:z1_definition}. 
Still, we left this second approach also available in the code since the performance may depend on the specific problem and training process}. Another option is to keep Eq.~\ref{eq:z1_definition} and add to the training loss the auxiliary term
$\omega\,\text{relu}\left(\kappa\left\Vert \boldsymbol{v}\right\Vert -1\right)^{2}$, 
where $\omega>0$ is a weight. This term is zero whenever $\boldsymbol{z}_{0}+\boldsymbol{v}\in\mathcal{Z}$, it does not modify the output of RAYEN (nor, therefore, its constraint satisfaction guarantees), and, when $\boldsymbol{z}_{0}+\boldsymbol{v}\notin\mathcal{Z}$,  it encourages $\boldsymbol z_0+\boldsymbol v$ to return to $\mathcal{Z}$.
\end{addedstuffnew}

\section{Conclusion and Future work}\label{sec:future_work}

This work presented RAYEN, a framework to impose hard convex constraints
on the output or latent variable of a neural network. RAYEN is able
to guarantee by construction the satisfaction of any combination of
linear, convex quadratic, SOC, and LMI constraints. When applied to
approximate the solution of constrained optimization problems, RAYEN
showcases computation times between 20 and 7468 times faster than
state-of-the-art algorithms, while guaranteeing the satisfaction of
the constraints at all times and obtaining a loss very close to the
optimal one. 

Future work includes the \added{application of RAYEN to non-fixed convex constraints (Section~\ref{subsec:non_fixed_constraints}), and the incorporation of more types of convex constraints (such as, for example, exponential cone constraints)}. RAYEN could
still be used for any convex constraint as long as the corresponding
$\kappa_{\square}$ for that constraint can be found. Even if an analytic solution for $\kappa_{\square}$ does not exist, one could always
leverage implicit layers that find the roots of nonlinear equations~\cite{poli2021torchdyn} to numerically find it. We also plan to study
how RAYEN could be used to impose generic nonconvex constraints~\cite{beucler2021enforcing,donti2021dc3}
(via, e.g., sequential convexification), to impose Lipschitz constraints~\cite{pauli2022neural}, to impose generic matrix manifolds constraints~\cite{lezcano2019trivializations}, or leveraged for safe robot control~\cite{xiao2023barriernet}. \addednew{Moreover, and to test RAYEN's ability to impose SOC and LMI constraints on a physical robot, we would also like to perform real-robot experiments involving friction-cone or stability constraints.}

\appendix

\section{Appendix}

\subsection{Proof of $\text{eig}\left(-\boldsymbol{H}^{-1}\boldsymbol{S}\right)=\text{eig}\left(\boldsymbol{L}^{T}\left(-\boldsymbol{S}\right)\boldsymbol{L}\right)$\label{subsec:Eigenvalues-of-mHinvS}}

Given that $\boldsymbol{H}^{-1}\succ\boldsymbol{0}$ (because $\boldsymbol{H}\succ\boldsymbol{0}$),
we can write its Cholesky decomposition as $\boldsymbol{H}^{-1}=\boldsymbol{L}\boldsymbol{L}^{T}$,
where $\boldsymbol{L}$ is a lower triangular matrix with positive
diagonal entries. Hence:
\[
\text{eig}\left(-\boldsymbol{H}^{-1}\boldsymbol{S}\right)=\text{eig}\left(\boldsymbol{L}\boldsymbol{L}^{T}\left(-\boldsymbol{S}\right)\right)=\text{eig}\left(\boldsymbol{L}^{T}\left(-\boldsymbol{S}\right)\boldsymbol{L}\right)\;,
\]
where in the last step we have \added{used} the fact that $\text{eig}\left(\boldsymbol{A}\boldsymbol{B}\right)=\text{eig}\left(\boldsymbol{B}\boldsymbol{A}\right)$ for square matrices $\boldsymbol{A}$ and $\boldsymbol{B}$
(\cite{bhatia2002eigenvalues}). As $\boldsymbol{L}^{T}\left(-\boldsymbol{S}\right)\boldsymbol{L}$
is clearly a real symmetric matrix, this also proves that all the
eigenvalues of $-\boldsymbol{H}^{-1}\boldsymbol{S}$ are real, even
though $-\boldsymbol{H}^{-1}\boldsymbol{S}$ may not be a symmetric
matrix \cite{eigenvaluesProductMartin2022}.

\subsection{Training, validation, and test sets \label{subsec:Training,-validation,-and}}

The $\boldsymbol{\gamma}$ used in Table \ref{tab:all_algorithms}
(the input of the network) is $\arraycolsep=2.1pt\boldsymbol{\gamma}:=\left[\begin{array}{cccc}
\alpha_{\text{v}} & \alpha_{\text{a}} & \alpha_{\text{j}} & \mathbf{\mathbf{p}}_{f}^{T}\end{array}\right]^{T}$, which contains the nonnegative weights $\left\{ \alpha_{\text{v}},\alpha_{\text{a}},\alpha_{\text{j}}\right\} $
and the final position $\mathbf{\mathbf{p}}_{f}$. Letting $j\in\{1,2\}$
denote the optimization problem, we use the following training, validation,
and test sets:
\begin{itemize}
\item \textbf{Training and Validation sets:} $\arraycolsep=1.6pt\left[\begin{array}{ccc}
\alpha_{\text{v}} & \alpha_{\text{a}} & \alpha_{\text{j}}\end{array}\right]^{T}$ is taken uniformly within the cube $[0.0,1.0]^{3}$, while $\mathbf{\mathbf{p}}_{f}$
is taken uniformly within $\mathcal{P}_{n_{\text{polyh}}-1}$.
\item \textbf{Testing sets:}
\begin{itemize}
\item $\text{TE}_{j,\text{in}}$: Same distributions as above.
\item $\text{TE}_{j,\text{out}}$: $\arraycolsep=1.6pt\left[\begin{array}{ccc}
\alpha_{\text{v}} & \alpha_{\text{a}} & \alpha_{\text{j}}\end{array}\right]^{T}$ is taken uniformly within the cube $[1.0,2.0]^{3}$, while $\mathbf{\mathbf{p}}_{f}$
is taken uniformly within $\mathcal{P}_{n_{\text{polyh}}-1}$.
\end{itemize}
\end{itemize}
\section*{Acknowledgment}
The authors would like to thank Dr.~Kasra Khosoussi for helpful insights and discussions. \addednew{The authors would also like to thank the reviewers for their highly useful feedback.} Research supported in part by the Air Force Office of Scientific Research MURI FA9550-19-1-0386 and the NCCR automation.

\selectlanguage{english}%
\bibliographystyle{SageH}
\bibliography{my_bib}
\selectlanguage{american}%

\end{document}